\documentclass{article} % For LaTeX2e
\usepackage{iclr2023_conference,times}
\usepackage{balance}
\usepackage{booktabs}
\usepackage{enumitem}

\usepackage{amsmath,amsfonts,bm}

\def\eqref#1{equation~\ref{#1}}
\def\1{\bm{1}}

\DeclareMathAlphabet{\mathsfit}{\encodingdefault}{\sfdefault}{m}{sl}
\SetMathAlphabet{\mathsfit}{bold}{\encodingdefault}{\sfdefault}{bx}{n}

\newcommand{\method}[1]{SMART}

\usepackage{amsmath}
\usepackage{amssymb}
\usepackage{mathtools}
\usepackage{amsthm}

\usepackage[font=small,labelfont=bf]{caption, subcaption}
\usepackage{wrapfig}
\usepackage{subfloat}
\usepackage{multirow} 
\usepackage{wrapfig,lipsum,booktabs}

\usepackage{colortbl}
\usepackage{tablefootnote}

\usepackage{tikz}
\usetikzlibrary{fit}
\usetikzlibrary{calc,shapes}
\usetikzlibrary{decorations.pathmorphing} % noisy shapes
\usetikzlibrary{fit}					% fitting shapes to coordinates
\usetikzlibrary{backgrounds}	% drawing the background after the foreground
\usetikzlibrary{pgfplots.groupplots}

\usepackage[utf8]{inputenc}
\usepackage{pgfplots}
\DeclareUnicodeCharacter{2212}{−}
\usepgfplotslibrary{groupplots,dateplot}
\usetikzlibrary{patterns,shapes.arrows}
\pgfplotsset{compat=newest}

\usepackage{array}
\usepackage{hyperref}
\usepackage{url}
\usepackage[capitalize,noabbrev]{cleveref}
\usepackage{xspace}
\usepackage{soul}

\newcommand{\ourmod}{{CT}\xspace}
\newcommand{\ourmodfull}{{Control Transformer}\xspace}
\newcommand{\ours}{{SMART}\xspace}
\newcommand{\oursfull}{{Self-supervised Multi-task pretrAining with contRol Transformer}\xspace}
\newcommand{\randinvcap}{{Random Masked Hindsight Control}\xspace}
\newcommand{\randinv}{{random masked hindsight control}\xspace}
\newcommand{\randinvshort}{{Mask-Ctl}\xspace}
\newcommand{\ourobj}{{control-centric objective}\xspace}

\newcommand{\tasks}{\mathcal{T}}
\newcommand{\mdp}{\mathcal{M}}
\newcommand{\states}{\mathcal{S}}
\newcommand{\actions}{\mathcal{A}}
\newcommand{\obss}{\mathcal{O}}
\newcommand{\transition}{P}
\newcommand{\reward}{R}
\newcommand{\emission}{E}

\newcommand{\history}{h}

\newcommand{\rep}{\phi}

\newcommand{\fwdmath}{\text{fwd}}
\newcommand{\invmath}{\text{inv}}
\newcommand{\randinvmath}{\text{mask-inv}}

\definecolor{darkgray176}{RGB}{176,176,176}
\definecolor{darkgreen}{RGB}{0,100,0}
\definecolor{darkorange}{RGB}{255,140,0}
\definecolor{lightgray204}{RGB}{204,204,204}
\definecolor{silver}{RGB}{192,192,192}
\definecolor{steelblue}{RGB}{70,130,180}

\title{SMART: Self-supervised Multi-task pretrAining with contRol Transformers}

\author{
Yanchao Sun$^{\star}$, Shuang Ma$^{\dagger}$, Ratnesh Madaan$^{\dagger}$, Rogerio Bonatti$^{\dagger}$, \\
\hspace{0.5em}\textbf{Furong Huang$^{\star}$ and Ashish Kapoor$^{\dagger}$}\\
\thanks{University of Maryland, College Park, MD. This work was done when the first author was an intern at Microsoft.}
{\tt\small \{ycs, furongh\}@umd.edu} \\
\thanks{Microsoft Redmond, WA}
{\tt\small \{shuama, ratnesh.madaan, rbonatti ,akapoor\}@microsoft.com} 
}

\iclrfinalcopy % Uncomment for camera-ready version, but NOT for submission.
\begin{document}

\maketitle

\begin{abstract}
Self-supervised pretraining has been extensively studied in language and vision domains, where a unified model can be easily adapted to various downstream tasks by pretraining representations without explicit labels.
When it comes to sequential decision-making tasks, however, it is difficult to properly design such a pretraining approach that can cope with both high-dimensional perceptual information and the complexity of sequential control over long interaction horizons.
The challenge becomes combinatorially more complex if we want to pretrain representations amenable to a large variety of tasks. 
To tackle this problem, in this work, we formulate a general pretraining-finetuning pipeline for sequential decision making, under which we propose a generic pretraining framework \textit{Self-supervised Multi-task pretrAining with contRol Transformer (SMART)}. 
By systematically investigating pretraining regimes, we carefully design a Control Transformer (CT) coupled with a novel control-centric pretraining objective in a self-supervised manner. 
SMART encourages the representation to capture the common essential information relevant to short-term control and long-term control, which is transferrable across tasks.
We show by extensive experiments in DeepMind Control Suite that SMART significantly improves the learning efficiency among seen and unseen downstream tasks and domains under different learning scenarios including Imitation Learning (IL) and Reinforcement Learning (RL).
Benefiting from the proposed control-centric objective, SMART is resilient to distribution shift between pretraining and finetuning, and even works well with low-quality pretraining datasets that are randomly collected.

\end{abstract}

% Intro
\section{Introduction}

Self-supervised pretraining has been successful in a wide range of language and vision problems. Examples include BERT~\citep{devlin2019bert}, GPT~\citep{gpt:brown2020language}, MoCo~\citep{he2020momentum}, and CLIP~\citep{radford2021learning}. 
These works demonstrate that one single pretrained model can be easily finetuned to perform many downstream tasks, resulting in a simple, effective, and data-efficient paradigm. 
When it comes to sequential decision making, however, it is not clear yet whether the successes of pretraining approaches can be easily replicated.
% However, pretraining for sequential decision making is still under-explored.

There are research efforts that investigate application of pretrained vision models to facilitate control tasks~\citep{parisi2022unsurprising, radosavovicreal}. However, there are challenges unique to sequential decision making and beyond the considerations of existing vision and language pretraining. We highlight these challenges below:
% which makes it an open problem of seeking the ideal representation for decision making tasks.
(1) \textit{Data distribution shift:}
Training data for decision making tasks is usually composed of trajectories generated under some specific behavior policies.
% different from i.i.d. samples generally used in supervised learning. 
As a result, data distributions during pretraining, downstream task finetuning and even during deployment can be drastically different, resulting in a suboptimal performance~\citep{lee2021offlinetoonline}.
(2) \textit{Large discrepancy between tasks:}
In contrast to language and vision where the underlying semantic information is often shared across tasks, decision making tasks span a large variety of task-specific configurations, transition functions, rewards, as well as action and state spaces. Consequently, it is hard to obtain a generic representation for multiple decision making tasks. 
(3) \textit{Long-term reward maximization:}
% A priori it is often hard to estimate the time-horizon over which the rewards should be maximized. 
The general goal of sequential decision making is to learn a policy that maximizes long-term reward. 
Thus, a good representation for downstream policy learning should capture information relevant for both immediate and long-term planning, which is usually hard in tasks with long horizons, partial observability and continuous control. 
% capturing temporal correlations between observations and actions. \\
(4) \textit{Lack of supervision and high-quality data:} 
% To add on to the complexity, continuous actions demand a higher granularity of learned representations.
Success in representation learning often depends on the availability of high quality expert demonstrations and ground-truth rewards~\citep{lee2022multi, stooke2021decoupling}. However, for most real-world sequential decision making tasks, high-quality data and/or supervisory signals are either non-existent or prohibitively expensive to obtain. %Additionally, uncertain dynamics models and partial observability common to many realistic settings further aggravate the data issues.

Under these challenges, we strive for pretrained representations for control tasks that are \\
(1) {\bf Versatile} so as to handle a wide variety of downstream control tasks and variable downstream learning methods such as imitation and reinforcement learning (IL, RL) etc, \\
%time-granularities 
(2) {\bf Generalizable} to unseen tasks and domains spanning multiple rewards and agent dynamics, and \\
% , reward structures, transition functions \\
(3) {\bf Resilient} and robust to varying-quality pretraining data without supervision. %\\
% and adaptable to large state-action distribution shifts and robust to data of variable quality. 
% Finally, we also want the representation to be useful under different learning regimes such as imitation and reinforcement learning (IL, RL) etc.

% ======== technical
We propose a general pretraining framework named \textit{\oursfull (\ours)}, which aims to satisfy the above listed properties.
We introduce \textit{\ourmodfull(\ourmod)}  which models state-action interactions from high-dimensional observations through causal attention mechanism. 
Different from the recent transformer-based models for sequential decision making~\cite{chen2021decisiontransformer} which directly learn reward-based policies, \ourmod is designed to learn reward-agnostic representations, which enables it as a unified model to fit different learning methods (e.g. IL and RL) and various tasks. 
% By offline pretraining on multiple control tasks, \ours aims to learn generalizable perceptual and control information as representation. 
Built upon \ourmod, we propose a control-centric pretraining objective that consists of three terms: forward dynamics prediction, inverse dynamics prediction and \randinv. 
These terms focus on policy-independent transition probabilities, and encourage \ourmod to capture dynamics information of both short-term and long-term temporal granularities.
In contrast with prior pretrained vision models~\citep{CPC:oord2018representation,parisi2022unsurprising} that primarily focus on learning object-centric semantics, \ours captures the essential control-relevant information which is empirically shown to be more suitable for interactive decision making.  
\ours produces superior performance than training from scratch and state-of-the-art (SOTA) pretraining approaches on a large variety of tasks under both IL and RL. Our main contributions are summarized as follows:

\vspace{-0.5em}
\begin{enumerate}[noitemsep,leftmargin=*]
\setlist{nolistsep}
    \item We propose \ours, a generic pretraining framework for multi-task sequential decision making. 
    \item We introduce the \ourmodfull model and a control-centric pretraining objective to learn representation from offline interaction data, capturing both perceptual and dynamics information with multiple temporal granularities.
    % \item We propose \ours, a generic pretraining framework for sequential decision making. \ours is \textit{versatile}, and can be used for multiple downstream tasks under both IL and RL regimes.
    % \item We conduct extensive experiments showing that \ours is \textit{generalizable}, and can significantly improve the performance of downstream tasks in multiple domains in the Deepmind Control Suite benchmark, even when in previously unseen domains.
    % \item We show that \ours is \textit{resilient} to changes in data distribution, working well even with randomly collected pretraining datasets. While other pretraining method baselines suffer downgraded performance under such conditions, our method shows greater flexibility and can be applied towards more realistic applications where expert data  collection is a challenge. 
    % \ours can even produce high-quality representations when pretrained on randomly-collected trajectories, which demonstrate its resilience to distribution shifts.
    \item We conduct extensive experiments on DeepMind Control Suite~\citep{tassa2018deepmind}. By evaluating \ours on a large variety of tasks under both IL and RL regimes, \ours demonstrates its \textit{versatile} usages for downstream applications. When adapting to unseen tasks and unseen domains, \ours shows superior \textit{generalizability}. \ours can even produce compelling results when pretrained on low-quality data that is randomly collected, validating its \textit{resilience} property.
\end{enumerate}
\vspace{-0.5em}
 \vspace{-2mm}
\section{Related works}
\label{sec:relate}
 \vspace{-3mm}

\textbf{Offline Pretraining of Representation for Control.} 
% Deep Reinforcement Learning (DRL) is known to be sample-consuming, especially with high-dimensional observations such as images. 
Many recent works investigate pretraining representations and finetuning policies for the same task. 
% \citet{yang2021representation} investigate several pretraining objectives for multiple downstream tasks on MuJoCo environments with vector state inputs. 
% They find that many existing representation learning objectives fail to improve the downstream task, while contrastive self-prediction obtains the best results among all tested methods.
\citet{yang2021representation} investigate several pretraining objectives on MuJoCo with vector state inputs. 
They find that many existing representation learning objectives fail to improve the downstream task, while contrastive self-prediction obtains the best results among all tested methods.
\citet{schwarzer2021pretraining} pretrain a convolutional encoder with a combination of several self-supervised objectives, achieving superior performance on the Atari 100K. 
However, these works just demonstrated the single-task pretraining scenario, it is not clear yet whether the methods can be extended to multi-task control.
\citet{stooke2021decoupling} propose ATC, a contrastive learning method with temporal augmentation. By pretraining an encoder on expert demonstrations from one or multiple tasks, ATC outperforms prior unsupervised representation learning methods in downstream online RL tasks, even in tasks unseen during pretraining.

\textbf{Pretrained Visual Representations for Control Tasks.}
Recent studies reveals that visual representations pretrained on control-free datasets can be transferred to control tasks.
\citet{shah2021rrl} show that that a ResNet encoder pretrained on ImageNet is effective for learning manipulation tasks. Some recent papers also show that encoders pretrained with control-free datasets can generalize well to RL settings~\citep{nair2022r3m,seo2022reinforcement,parisi2022unsurprising}. However, the generalizability of the visual encoder can be task-dependent. \citet{kadavath2021pretraining} point out that ResNet pretrained on ImageNet does not help in DMC~\citep{tunyasuvunakool2020} environments.

% %=============== decoupled visual state representation learning=================
% unsurprising~\cite{parisi2022unsurprising}
% ATC~\cite{stooke2021decoupling}
% CPC~\cite{CPC:oord2018representation}

% %============== RL pretraining for single task ======

\textbf{Unsupervised RL.}
Unsupervised RL (URL) focuses on learning exploration policies~\citep{pathak2017curiosity,liu2021behavior}, goal-conditioned policies~\citep{andrychowicz2017hindsight,mendonca2021discovering} or diverse skills~\citep{eysenbach2018diversity} in a task without external rewards, and finetuning the policy later when reward is specified. The unsupervised learning phase of URL is usually interactive and prolonged. Our goal, in contrast, is to train representations of states and actions from fixed offline datasets, with a focus on capturing essential and important information from raw inputs. 

\textbf{Sequential Decision Making with Transformers.}
There is a growing body of work that uses Transformer~\citep{vaswani2017attention} architectures to model and learn sequential decision making problems.
\cite{chen2021decisiontransformer} propose Decision Transformer (DT) for offline RL, which takes a sequence of returns, observations and actions, and outputs action predictions.
Trajectory Transformer~\citep{janner2021offline} also models the trajectory as a sequence of states, actions and rewards, while discretizing each dimension of state/actions.
\cite{bonatti2022pact} propose a pretraining scheme for state-action representations in navigation scenarios using a causal transformer, which can then be finetuned with imitation learning towards different tasks for the same robot.
\citet{furuta2022generalized} propose Generalized DT that unifies a family of algorithms for future information matching with transformers. 
\citet{zheng2022online} extend DT to online learning by blending offline pretraining and online finetuning.
Transformers can also be used as world models for model-based RL~\citep{chen2022transdreamer,micheli2022transformers}.
Recent studies show that transformer-based models can be scaled up with diverse multi-task datasets to produce generalist agents~\citep{reed2022generalist,lee2022multi}.
Our proposed method has a similar structure that regards RL trajectories as sequential inputs. However, differently from most existing transformer models that learn policy from returns, our \ours focuses on learning control-relevant representations with self-supervised pretraining.
% More discussion and comparison with prior models are in \cref{sec:algo}. 

% % ======= causal T =========
% DT~\cite{}
% multi-game DT
% online DT
% Gato

% Preliminary
\vspace{-3mm}
\section{Preliminaries}
\label{sec:prelim}
\vspace{-2mm}

\begin{figure}[t]
    % \vspace{-1em}
    \centering
    \includegraphics[width=\textwidth]{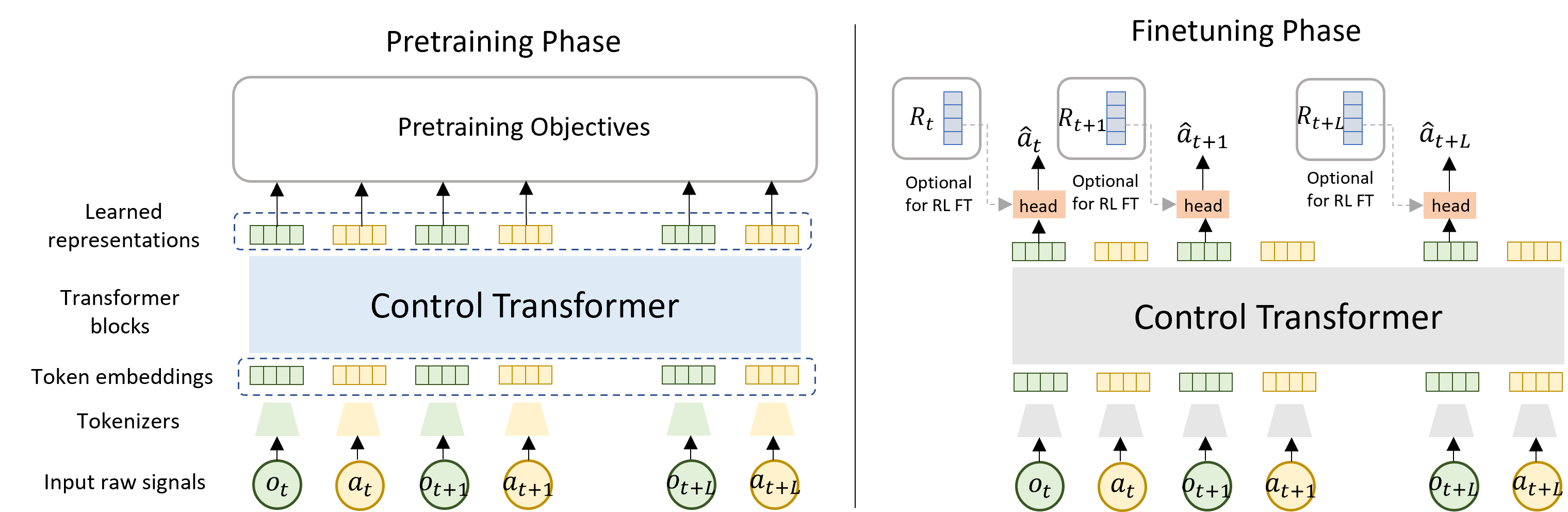}
    % \vspace{-1em}
    \caption{Architecture of \ourmodfull. In the pretraining phase, we use the \ourobj introduced in \cref{ssec:obj} to train representation over multiple tasks; in the finetuning phase where a specific task is given, we learn a policy based on the pretrained representation (pretrained weights are shown in grey blocks). The construction of the policy head can vary for different downstream datasets or learning methods. 
    % For example, RTG embeddings can be fed into the policy head when learning RTG-conditioned policies. 
    }
    \label{fig:model}
    \vspace{-1.5em}
\end{figure}
\textbf{Partially Observable Markov Decision Process.} 
We model control tasks and environments as a partially observable Markov decision process (POMDP) $\mdp = \langle \states, \actions, \obss, \transition, \reward, \emission \rangle$, which is a generalization of Markov decision process (MDP).
Here, $\states$ is the underlying state space, $\actions$ is the action space, $\obss$ is the observation space, $\transition$ is the transition kernal, $\reward$ is the reward function, and $\emission$ is the observation emission function with $\emission(o|s)$ being the probability of observing $o$ given state $s$.
In practice, the observation space can be high dimensional. 
For example, for a mobile robot navigating with camera sensors, the images are observations and its odometry (location, orientation, and associated velocities) and the ground-truth obstacle locations form the underlying state.

\textbf{Learning Agent and Controlling Policy.} 
At every step $t$, the agent receives an observation $o_t$ based on the underlying state $s_t$ (hidden from the agent), takes action $a_t$, and obtains a reward $r_t$ and the environment proceeds to the next state $s_{t+1}$. 
Given a history of observation-action pairs of length $L$ and the current observation,  $\history_t=(o_{t-L}, a_{t-L}, o_{t-L+1}, a_{t-L+1},\cdots, o_{t})$, the agent executes action $a_t$ according to policy $\pi$: $a_t=\pi(\history_t)$.
The agent's goal is to learn an optimal policy $\pi^*$ that maximizes the agent's cumulative reward $\mathbb{E}_{P} [\sum_{t=1}^\infty \gamma^t r_t]$. 

\textbf{Reinforcement Learning (RL) and Imitation Learning (IL).}
RL~\citep{sutton2018reinforcement} is the process where an agent seeks to maximize its policy returns. 
RL agents can learn in an online manner by interacting with the environment, or learn from offline data with pre-collected interactions.
Differently from supervised learning, the training data for RL stems from  policy-dependent interactions with the environment, rendering a non-i.i.d. training regime.
Another effective method for learning control policies is IL~\citep{hussein2017imitation,osa2018algorithmic}, where an agent obtains supervision from expert demonstrations.
For both RL and IL, there could be a discrepancy between training and deployment data distributions due to the different distribution of states induced by the imperfect policy and environment uncertainty.
% Therefore we usually require a large amount of high-quality training data to achieve high performance, sometimes coupled with iterative training algorithms~\citep{ross2011reduction}.
% in environments with unknown dynamics

% \textbf{Transformer for Sequential Decision Making.} 
% Transformer and DT

% Our Method
% \section{Our Pretraining Model and Approach}
% \label{sec:algo}
\vspace{-3mm}
\section{Problem Setup: Pretraining and Finetuning Pipeline}
\label{ssec:setup}
\vspace{-2mm}
\textbf{Multi-Task Control with Shared Representation.} 
% We consider a set of multiple tasks $\tasks$, and assume that observations from all tasks have the same dimensionality. 
We consider a set of multiple tasks $\tasks$ with the same dimensionality in observation space.
In this work we select $\tasks$ from different environment in DeepMind Control Suite (DMC)~\citep{tassa2018deepmind}, in which the agent observes an RGB image of the current state.
Tasks in $\tasks$ can have entirely different state spaces $\states$, different action spaces $\actions$ and different environment dynamics ($\transition, \reward, \emission$). 
We also define the concept of \textit{domain} to differentiate tasks that have different state/action spaces. 
For example, in DMC, ``hopper'' and ``walker'' belong to different domains because they posses distinct action spaces, while ``walker-walk'' and ``walker-run'' are different tasks within the same domain. 
In this paper we use the term \textit{multi-task} to refer tasks spanning potentially multiple domains.
% A graphical relations of tasks involved is shown in Fig.\ref{fig:tasks}, and more details are shown in Appendix \ref{app:imple_env}. \ys{Appendix and the figure only show the tasks we use in experiments, but here we are just formulating the problem and using walker as an example. So I don't think we need to refer to appendix now (we do so in the experiment)}
% Our goal is to learn generic pretrained representations that capture essential and common information, so as to improve policies for different downstream tasks.

\begin{wraptable}{r}{8cm}
\vspace{-2.4em}
\caption{A comparison between pretraining and finetuning.}
\label{tab:pt_ft}
\vspace{-1.1em}
\resizebox{8cm}{!}{
\begin{tabular}{m{4cm}|m{4cm}}
\toprule  
\textbf{Pretraining phase} & \textbf{Finetuning phase} \\ \hline
 Learn generic representation & Learn policy \\ \hline
 Offline & Offline or online \\ \hline
 Multiple tasks & One task, seen or unseen \\ \hline
 Reward or expert demonstration may be absent & Has reward supervision or expert demonstration \\ \hline 
 More samples & Fewer samples \\  \bottomrule
\end{tabular}}
\vspace{-4mm}
\end{wraptable}

\textbf{Pretraining-Finetuning Pipeline.} 
Although pretraining methods are widely applied in many areas, it is not yet clear what role pretraining should play in sequential decision making tasks,
% or RL, 
especially when considering the multi-task setup.
In this work, we follow ideas established in vision and language community to explicitly define our pretraining and finetuning pipeline, which we summarize in Table~\ref{tab:pt_ft}.
% characterizes and compares different properties of pretraining phase and finetuning phase. 
Specifically, during the pretraining phase we train representations with a possibly large offline dataset collected from a set of training tasks $\tasks_{\text{pre}} = \{ \mdp_i \}_{i=1}^n$. 
Then, given a specific downstream task $\mdp$ which may or may not be contained in $\tasks_{\text{pre}}$, we attach a simple policy head on top of the pretrained representation\footnote{The pretrained encoder can be either frozen or finetuned with the policy, depending on the task.} and train it with IL or with RL.
% , either in an online or offline regime. 
% The downstream learning algorithm can be either offline (e.g. behavior cloning) or online (e.g., DQN~\citep{mnih2015human}). 
The central tenet of pretraining is to learn generic representations which allow downstream task finetuning to be simple, effective and efficient, even under low-data regimes.

This pretraining-finetuning pipeline is a general extension of many prior settings of pretraining for decision making.
For example, \citet{stooke2021decoupling} pretrain an encoder on one or multiple tasks, then learn an RL policy in downstream tasks. 
Their pretraining dataset is composed of expert demonstration, and the finetuning process focuses on online RL.
In addition, \citet{schwarzer2021pretraining} learn representations with offline datasets, but perform pretraining and finetuning within the same task.

\vspace{-2mm}
\section{Our Pretraining Model and Approach}
\label{sec:algo}
\vspace{-2mm}
We propose \oursfull (\ours), a general pretraining approach for multi-task sequential decision making. 
We first give an overview of the proposed Control Transformer (CT) and illustrate how it fits in our pretraining-finetuning pipeline in \cref{ssec:overview}. 
Then, we introduce our control-centric pretraining objective in Section~\ref{ssec:obj}. 

\vspace{-1mm}
\subsection{Approach Overview and Model Architecture}
\label{ssec:overview}
\vspace{-2mm}
\textbf{Model Architecture of \ourmodfull.} 
% Learning POMDP tasks with image observations is challenging, as we need to capture both visual (spatial) information and transition (temporal) information. Inspired by the recent success of transformer models in reinforcement learning, we use a causal transformer with CNN tokenizers as the model backbone, such that the spatial features can be captured by the CNN and temporal features can be learned by self-attention. \\
Inspired by the recent success of transformer models in sequential modeling~\citep{chen2021decisiontransformer,janner2021offline}, we propose a \ourmodfull (\ourmod). 
The input to the model is a control sequence of length $2L$ composed of observations and actions: $(o_t, a_t, o_{t+1}, a_{t+1}, \cdots, o_{t+L}, a_{t+L})$.
Different from Decision Transformer (DT)~\citep{chen2021decisiontransformer}, we purposefully do not include the reward signal in the control sequence to keep our representations reward-agnostic, as explained in the end of \cref{sec:downstream-arch}.
Each element of the sequence is embedded into a $d$-dimensional token, with a modality-specific tokenizer jointly trained with Transformer blocks.
% Each element of the sequence is embedded into a $d$-dimensional token, with a learnable tokenizer for each modality.
% (e.g., a CNN for image observations, and an MLP for actions).
% We use a convolutional network to tokenize image observations, and a linear layer to tokenize actions. 
% We also learn an additional time positional embedding and sum it with each token.
We also learn an additional positional embedding and sum it with each token.
The outputs of CT correspond to token embeddings representing each observation and action, and are represented by $\rep(o_t)$ and $\rep(a_t)$, respectively.\footnote{The implementation details are provided in \cref{app:imple_model}.}
% These representations are trained with different prediction/policy heads during pretraining and downstream finetuning, as detailed below.
% For both pretraining and downstream finetuning, we append prediction and policy heads to these representations. 
% During the pretraining phase, we use prediction heads to optimize the control-centric pretraining objective introduced in \cref{ssec:obj}, while policy heads are used in the task-specific finetuning phase.
% Prediction heads are used during the pretraining phase with autoregressive losses, while policy heads are used in the task-specific finetuning phase.
Figure~\ref{fig:model} depicts the \ourmod architecture.
% , where the same transformer backbone is connected with different heads for pretraining and finetuning, as detailed below.  

% \begin{figure}[t]
%     % \vspace{-1em}
%     \centering
%     \includegraphics[width=\textwidth]{figures/model-arch.png}
%     % \vspace{-1em}
%     \caption{Architecture of \ourmodfull. In the pretraining phase, we use the \ourobj introduced in \cref{ssec:obj} to train representation over multiple tasks; in the finetuning phase where a specific task is given, we learn a policy based on the pretrained representation (pretrained weights are shown in grey blocks). The construction of the policy head can vary for different downstream datasets or learning methods. 
%     % For example, RTG embeddings can be fed into the policy head when learning RTG-conditioned policies. 
%     }
%     \label{fig:model}
%     \vspace{-1.5em}
% \end{figure}

\textbf{Pretraining of \ours.} 
We generate an offline dataset for pretraining which contains control trajectories generated by some behavior policies for a set of diverse tasks $\tasks_{\text{pre}}$ spanning multiple domains. 
% The behavior policies are not necessarily experts, and in fact we use exploratory policies and even random action generators, as detailed in \cref{ssec:exp_setup}. 
% (we use random policies or exploratory policies in experiments ()) on a set of \textit{source tasks}. 
% At every training epoch, we sample a batch of control sequences and optimize the model to minimize the proposed control-centric pretraining objective, as detailed in \cref{ssec:obj}. 
During pretraining, we append several prediction heads to the transformer output, and train the entire model by minimizing the control-centric pretraining objective introduced in \cref{ssec:obj}. These prediction heads are used to learn desired representations, will be dropped in finetuning.

\textbf{Downstream Finetuning of \ours.} \label{sec:downstream-arch}
As discussed in \cref{ssec:setup}, the pretrained representation can be used to learn policies for different tasks. 
To do so, we append a policy head $\pi$ to the observation representation, such that $\pi(\rep(o_t))$ predicts the proper action for observation $o_t$.
% When executing the policy, \ourmod is first given an initial observation $o_1$, and generates an action $a_1$, then it observes next observations and generates actions in an autoregressive way. 
We can train the policy head using both IL and RL.
% Training of the downstream policy can be done by IL, offline RL, or online RL. 
% In this work, we follow the paradigm of most transformer-based RL works~\citep{chen2021decisiontransformer,lee2022multi} and evaluate our model with IL and offline RL.
% In this work we present our main findings for the first two settings, given
For our IL experiments we use behavior cloning with expert demonstrations, where we feed $\rep(o_t)$ into a policy head to get action predictions.
For RL we use a return-to-go (RTG)-conditioned policy with trajectories that contain reward values. We feed $\rep(o_t)$ along with an RTG embedding to get the policy head's action predictions.
Online RL with transformer-based models is still a novel field and is not the focus of this work. But we show advantages of our pretraining method for online RL finetuning, following the same settings as online DT presented by \citet{zheng2022online}. See \cref{app:exp_online} for more discussion and results.

\textbf{Comparison with Prior Decision-making Transformers.}
Recent works leverage transformer architectures for modeling sequential decision making problems~\citep{chen2021decisiontransformer,janner2021offline,lee2022multi} as summarized in \cref{sec:relate}. 
Most of these models use reward information in the input sequence, as their goal is to directly learn a policy for a specific task.
In contrast, our goal is to pretrain representations for various downstream tasks, and thus our \ourmod uses reward-free control sequence as the model input; rewards are only used for downstream task when needed. 
There are several \textbf{benefits} of making representation agnostic to reward during pretraining. 
\textbf{(1)} A pretrained model requiring reward input does not flexibly fit some downstream learning scenarios such as behavior cloning, while \ourmod can be a unified model for various learning methods.
% A user can easily finetune our pretrained representation with an additional policy head.
A user can easily learn a policy under different learning methods (IL or RL) without modifying transformer blocks.
\textbf{(2)} Reward distribution can be significantly different when the task or policy changes, making a reward-dependent representation less resilient to distribution shift.
% learning a reward-dependent representation on pretraining tasks may hurt the downstream performance in other tasks. 
We show in \cref{ssec:exp_abl} that \textit{utilizing reward during pretraining may hurt the overall downstream performance}.
\vspace{-2mm}
\subsection{Control-centric Self-supervised Pretraining Objectives}
\label{ssec:obj}
\vspace{-2mm}

\begin{figure}[t]
    \centering
    \includegraphics[width=0.9\textwidth]{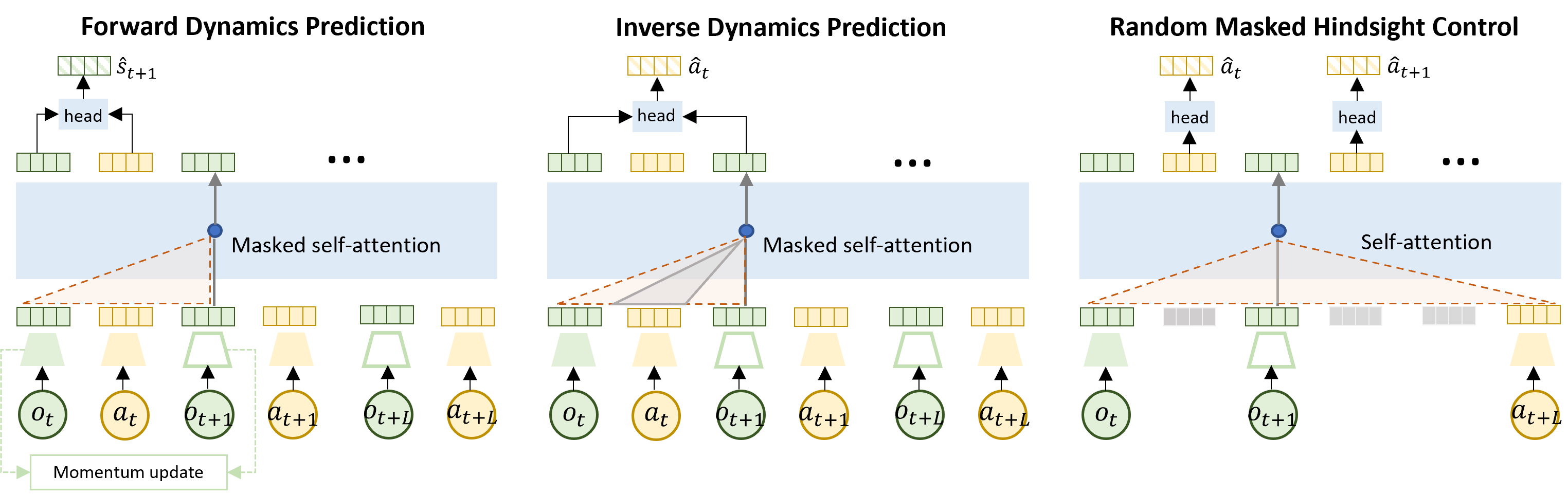}
    \vspace{-2mm}
    \caption{The three terms of our proposed pretraining objective.
    % These terms use different attention masks, while the weights of transformer and encoders are shared and jointly optimized. 
    The red shaded areas denote the attention span, while the grey regions are masked.}
    \label{fig:pretrain-obj}
    \vspace{-1.5em}
\end{figure}

Our pretraining objective employs three terms: forward dynamics prediction, inverse dynamics prediction, and \randinv. The first two terms focus on local and short-term dynamics, while the third term is designed to capture more global and long-term temporal dependence.
As motivated in \cref{ssec:overview}, these terms are based on control sequences and are reward-free, such that they can be used for multiple tasks. 
% The central idea of our pretraining framework is to learn a control-relevant representation that captures the information of both short-term dynamics and long-term dynamics. 
% We now introduce the three terms in our proposed objective, which are also visualized in Figure~\ref{fig:pretrain-obj}.
Figure~\ref{fig:pretrain-obj} illustrates each objective.

\textbf{I. Forward Dynamics Prediction.} 
For each observation-action pair $(o_t, a_t)$ in a control sequence, we aim to predict the next immediate latent state. 
Let $g$ be the observation tokenizer being trained. 
We maintain a momentum encoder $\bar{g}$ as the exponential moving average of $g$, to generate the target latent state from the next observation $o_{t+1}$, i.e.,  $\hat{s}_{t+1}:=\bar{g}(o_{t+1})$.
% whose parameters are updated by $\bar{\theta} = \tau (\bar{\theta}) + (1-\tau) \theta $. 
% The momentum encoder is used to generate the target latent state from the next observation $o_{t+1}$, resulting in  $\hat{s}_{t+1}:=\mathsf{SG}(\bar{E}_{\bar{\theta}}(o_{t+1}))$, where $\mathsf{SG}$ refers to stop gradient. 
The idea of momentum encoder is widely used when the target value is not fixed~\citep{he2020momentum,mnih2015human}, in order to make training more stable. Then, we train a linear head to predict $\hat{s}_{t+1}$ based on $(\rep(o_t),\rep(a_t))$.
% The forward prediction loss is defined as: 
% \begin{equation}
%     \label{loss:forward}
%     L_{\fwdmath} := \mathrm{MSE}\left( f_{\fwdmath}(\rep(o_t),\rep(a_t)), \hat{s}_{t+1} \right),
% \end{equation}
% where $f_{\fwdmath}$ is a linear prediction head. 
This forward prediction captures the local transition information in the embedding space. %\ys{letter E here conflicts with Emission function presented earlier. Can you change this letter? nice catch}

\textbf{II. Inverse Dynamics Prediction.} For each consecutive observation pair $(o_t, o_{t+1})$, we learn to recover the action that leads $o_t$ to $o_{t+1}$
% , which gives the loss function
% \begin{equation}
%     \label{loss:inverse}
%     L_{\invmath} := \mathrm{MSE}\left( f_{\invmath}(\rep(o_t),\rep(o_{t+1})), a_{t} \right),
% \end{equation}
% where $f_{\invmath}$ is a linear prediction head. 
Note that in a causal transformer, $a_t$ is visible to the model when generating representation of $o_{t+1}$ which can lead to trivial solutions, so we modify the original causal mask and mask out $a_t$ from the attention of $o_{t+1}$.
Therefore, the observation representation is forced to contain information for relevant actions and transitions. 

Both forward and inverse predictions focus on local dynamics induced by the transition kernel $P$. However, fitting local dynamics only may result in collapsed representation, i.e., the model learns the same representation for two semantically different observations~\citep{rakelly2021which}. To perform well in downstream tasks, long-term temporal dependence should also be captured in the representation. We achieve this by a novel \randinv term in pretraining. 

\textbf{III. \randinvcap.} 
% We propose to predict multiple randomly masked actions to learn both short-term and long-term temporal dependencies. 
Given a control sequence $\history=(o_t, a_t, \cdots, o_{t+L}, a_{t+L})$,
% (the last action is not used here and thus dropped), 
we randomly mask $k$ actions and $k^\prime$ observations,
% except for $o_t$ and $o_{t+L}$, 
and recover the masked actions based on the remaining incomplete sequence. 
% In our implementation, we use a predefined mask-token $\mathbf{m}$ to replace the original tokens. 
This idea of masked token prediction is related to BERT~\citep{devlin2019bert} for language modeling, but note that we only predict the masked actions for the purpose of control.
% The reason for dropping out $k^\prime$ observations is to force the model to learn global temporal relations. 
More rationale behind the selection of $k$ and $k^\prime$ is explained in \cref{app:imple_obj}.
This objective is akin to asking the question ``what actions should I take to generate such a trajectory?'' 
% Therefore, the global and long-term temporal dependence should be captured by the attentive representation. 
Therefore, we replace the causal attention mask with a non-causal one, to temporarily allow the model ``see the future'', as shown in Figure~\ref{fig:pretrain-obj}(right).
% Let $\tilde{h}$ be the masked control sequence, and $0 \leq s_1,s_2,\cdots,s_{k} \leq L$ be the selected indices for masked actions, then the loss function can be defined as:
% \begin{equation}
%     \label{loss:inverse}
%     L_{\randinvmath} := \sum\nolimits_{i=1}^{k} \mathrm{MSE}\left( f_{\randinvmath}(\rep_M(\mathbf{m}_{t+s_i});\tilde{h}), a_{t+s_i} \right),
% \end{equation}
% where $f_{\randinvmath}$ is a linear prediction head, and $\rep_M$ is the transformer model without a causal mask. 
As a result, we encourage the model to learn controllable representations and global temporal relations, and to attend to the most essential representations for multi-step control.
% learn embeddings and attentions that are useful for understanding the correlations among observations and actions. 
% In experiments, we gradually increase the value of $k$ and $k^\prime$ from 1 to $L$ and $L/2$, respectively. 
The idea of our \randinv is also related to the multi-step inverse prediction proposed by a concurrent work~\citep{lamb2022guaranteed}, which predicts $a_t$ given $s_t$ and $s_{t+l}$ for a random interger $l$ and theoretically shows the effectiveness of this method in discovering controllable states. 
Our \randinv is different as it predicts multiple actions altogether from randomly masked sequences with a transformer model, which can efficiently learn the control information in large-scale tasks, and avoid ambiguity caused by different paths between states. 
An empirical comparison between our formulation and the multi-step inverse prediction~\citep{lamb2022guaranteed} is provided in \cref{app:exp_ablation_variant}.

Finally, our pretraining objective is the summation of the above three terms with equal weights,
% i.e.,
% \begin{equation}
%     \label{loss:all}
%     \min_{\rep, f_{\fwdmath}, f_{\invmath}, f_{\randinvmath}} L_{\fwdmath} + L_{\invmath} + L_{\randinvmath}.
% \end{equation}
% We use equal weights for these objectives, 
which in experiments renders good performance. 
% Calibrating or learning the weights of these objectives may lead to better results in practice, but it is out of the scope of this paper. 
The mathematical formulations and implementation details of the objective are explained in \cref{app:imple_obj}.

% \textbf{Pretraining objectives}
% \begin{itemize}
%     \item Reward-conditioned
%     \item Forward (*) 
%     \item Inverse
%     \item Reward
% \end{itemize}

% \subsection{Downstream adapting manner}
% \begin{itemize}
%     \item BC
%     \item online RL
% \end{itemize}

% \subsection{Apply pretrained model to downstream tasks}
% \begin{itemize}
%     \item BC
%     \item offline RL
%     \item online RL
% \end{itemize}

% Experiments
\vspace{-3mm}
\section{Experiments}
\label{sec:exp}
\vspace{-2mm}
We provide empirical results to demonstrate the effectiveness of our proposed pretraining method, 
% To be specific, we
while aiming to answer the following questions: 
\textbf{(1)} Can \ours effectively improve learning efficiency and performance in a variety of downstream tasks under different learning methods?
\textbf{(2)} How well can \ours generalize to out-of-distribution tasks?
\textbf{(3)} Is \ours resistant to low-quality pretraining data?
\textbf{(4)} How does \ours compare to state-of-the-art pretraining techniques?
\textbf{(5)} How do different pretraining objectives affect the downstream performance?

\vspace{-2mm}
\subsection{Experimental Setup}
\label{ssec:exp_setup}
\vspace{-2mm}

% \textbf{Environments.}
We evaluate \ours on the DeepMind Control (DMC) suite~\citep{tassa2018deepmind}, which contains a series of continuous control tasks with RGB image observations.
There are multiple domains (physical models with different state and action spaces) and multiple tasks (associated with a particular MDP) within each domain, which creates diverse scenarios for evaluating pretrained representations. 
Our experiments use 10 different tasks spanning over 6 domains. 
In pretraining, we use an offline dataset collected over 5 tasks, while the other 5 tasks (with 2 unseen domains) are held out to test the generalizability of \ours.
% The model is pretrained with 5 different tasks from 4 domains, and tested on 5 unseen tasks, including 2 unseen domains. 
A full list of tested domains and tasks is in \cref{app:imple_env}.

\textbf{Pretraining Tasks and Datasets.}
We pretrain \ours on 5 tasks: 
cartpole-swingup, hopper-hop, cheetah-run, walker-stand and walker-run. For each task, we adopt/train behavior policies to collect the following two types of offline datasets. (Details of dataset collection are in \cref{app:imple_env}.)
\vspace{-3mm}
\begin{itemize}[noitemsep,leftmargin=*]
    \setlist{nolistsep}
    \item \texttt{Random}: Trajectories with random environment interactions, with 400K timesteps per task.
    \item \texttt{Exploratory}: Trajectories generated in the exploratory stage of multiple RL agents with different random seeds, with 400K timesteps per task. 
\end{itemize}
\vspace{-3mm}
% Note that the \texttt{Random} dataset is of low quality because it does not include optimal behaviors.
% \cref{ssec:exp_res} shows that most baseline methods pretrained from \texttt{Random} fail to adapt to downstream tasks well, while our \ours is resilient to \texttt{Random} pretraining data and renders good performance.

\textbf{Downstream Tasks, Learning Methods and Datasets.}
We evaluate the pretrained models in the 5 seen tasks and another 5 unseen tasks: cartpole-balance, hopper-stand, walker-walk, pendulum-swingup and finger-spin (the last two tasks are from unseen domains with different state-action spaces). 
% Note that last two tasks are from unseen domains with state-action spaces different from tasks in the pretraining dataset. 
We consider two learning methods: return-to-go-conditioned policy learning (RTG) and behavior cloning (BC).
For RTG, we use the \texttt{Sampled Replay} dataset containing randomly sampled trajectories from the full replay buffer collected by learning agent.
For BC, we use the \texttt{Expert} trajectories with the highest returns from the full replay buffer of learning agents. 
The downstream dataset for every task only has 100K timesteps, making it challenging to learn from scratch.

\textbf{Implementation Details.} 
Our implementation of \ourmod is based on a GPT model~\citep{radford2018improving} with 8 layers and 8 attention heads. We use context length $L$ = 30 and embedding size $d$ = 256. As explained in \cref{app:imple_obj}, $k$ and $k^\prime$ are linearly increased from 1 to $L$ and $L/2$, respectively.
% \sm{attention head?} \sm{we use random mask of k=？}
The observation tokenizer is a standard 3-layer CNN. Action tokenizer, return tokenizer, and all single-layer linear prediction heads.  
We found that freezing the pretrained weights in downstream tasks works well in relatively simple environments, but fails in harder ones. Therefore, we finetune the entire model including transformer blocks for all downstream tasks.
Since actions in different domains have different dimensions and physical meanings, we project the raw actions into a larger common action space to train the action tokenizer. 
When there is a novel downstream task with a different action space, we simply re-initialize the action tokenizer and finetune it. 
Please see \cref{app:imple_model} for more implementation and hyperparameter details. 

\textbf{Baselines.}
% In every downstream task, 
We compare \ours with the following transformer-based pretraining baselines:
\vspace{-3mm}
\begin{itemize}[noitemsep,leftmargin=*]
    \setlist{nolistsep}
    \item \texttt{Scratch} trains a policy with randomly initialized \ourmod representation weights.
    \item \texttt{ACL}~\citep{yang2021representation} is a modified BERT~\citep{devlin2019bert} that randomly masks and predicts tokens with a contrastive loss, pretrained on the same dataset as \ours.
    \item \texttt{DT}~\citep{chen2021decisiontransformer} pretrained on the same dataset as ours but uses extra reward supervision. %pretrains on the same dataset as ours by using reward as supervisory signals.
    \item \texttt{\ourmod-single} is a variant of \ours, which pretrains \ourmod with a single-task dataset containing trajectories from the downstream environment.
\end{itemize}
\vspace{-3mm}
For fair comparisons, we use the same network architecture for the baseline models (except for DT where we keep their original network structure with RTG as transformer inputs) and train them with the same configurations. 
% For DT, we keep their proposed network architecture, i.e., inserting extra RTG as inputs to the transformer).
% \textbf{(1)} Training a policy from scratch with randomly initialized \ourmod representation. 
% \textbf{(2)} Pretraining \ourmod with a single-task dataset containing trajectories from the downstream environment.
% \textbf{(3)} Multi-task pretraining with ACL proposed by~\citet{yang2021representation}, which is modified from Bert~\citep{devlin2019bert} using the same pretraining dataset as ours.
% \textbf{(4)} (Supervised) Multi-task pretraining with DT~\citep{chen2021decisiontransformer} model using our pretraining dataset with extra reward signals.
% We ensure that all baseline models have the same model architectures (except for DT which takes extra RTG inputs to the transformer) and are trained with the same configurations, for a fair comparison. 
We also compare \ours with other state-of-the-art pretraining works, such as \texttt{CPC}~\citep{CPC:oord2018representation} and \texttt{ATC}~\citep{stooke2021decoupling}, using the same pretraining-finetuning pipeline. 
However, as these approaches are built upon ResNet backbones, a direct comparison of a Transformer against a ResNet could be not straightforward. Hence we refer readers to \cref{app:exp_resnet} for more discussions and results.
% ResNet-based pretraining methods CPC~\citep{CPC:oord2018representation} and ATC~\citep{stooke2021decoupling} using the same pretraining-finetuning pipeline. But due to the difference of transformer models and ResNet models, it is not straightforward to fairly compare these pretraining approaches. We refer readers to \cref{app:exp_resnet} for more discussion and results.

\textbf{Evaluation Metrics.}
To evaluate the quality of pretrained representations we report the average cumulative reward obtained in downstream tasks after finetuning. 
We deploy the trained policies in each environment for 50 episodes and report average returns. 
For evaluation of the RTG-conditioned policies, we use the expert score of each task as the initial RTG, as done in DT~\citep{chen2021decisiontransformer}. 
Detailed evaluation settings as described in \cref{app:imple_metric}. 

\vspace{-2mm}
\subsection{Experimental Results}
\label{ssec:exp_res}
\vspace{-2mm}

\begin{figure}[t]
% \vspace{-1em}
 \centering

\rotatebox{90}{\scriptsize{\hspace{1cm}\textbf{RTG}}}
 \begin{subfigure}[t]{0.19\columnwidth}
  \resizebox{\textwidth}{!}{\input{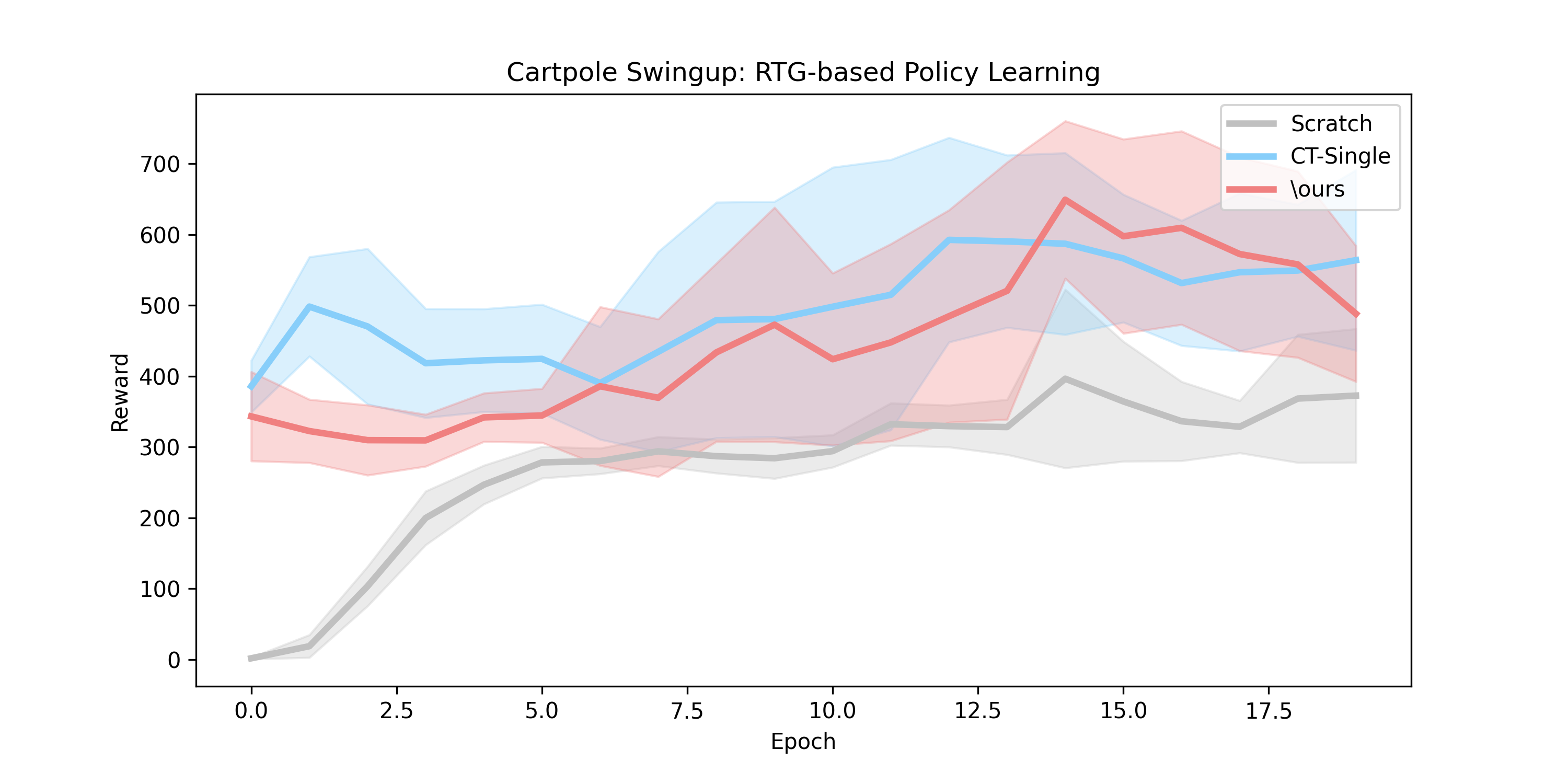}}
%   \vspace{-1.5em}
%   \caption{FoodCollector}
 \end{subfigure}
 \hfill
 \begin{subfigure}[t]{0.19\columnwidth}
  \resizebox{\textwidth}{!}{\input{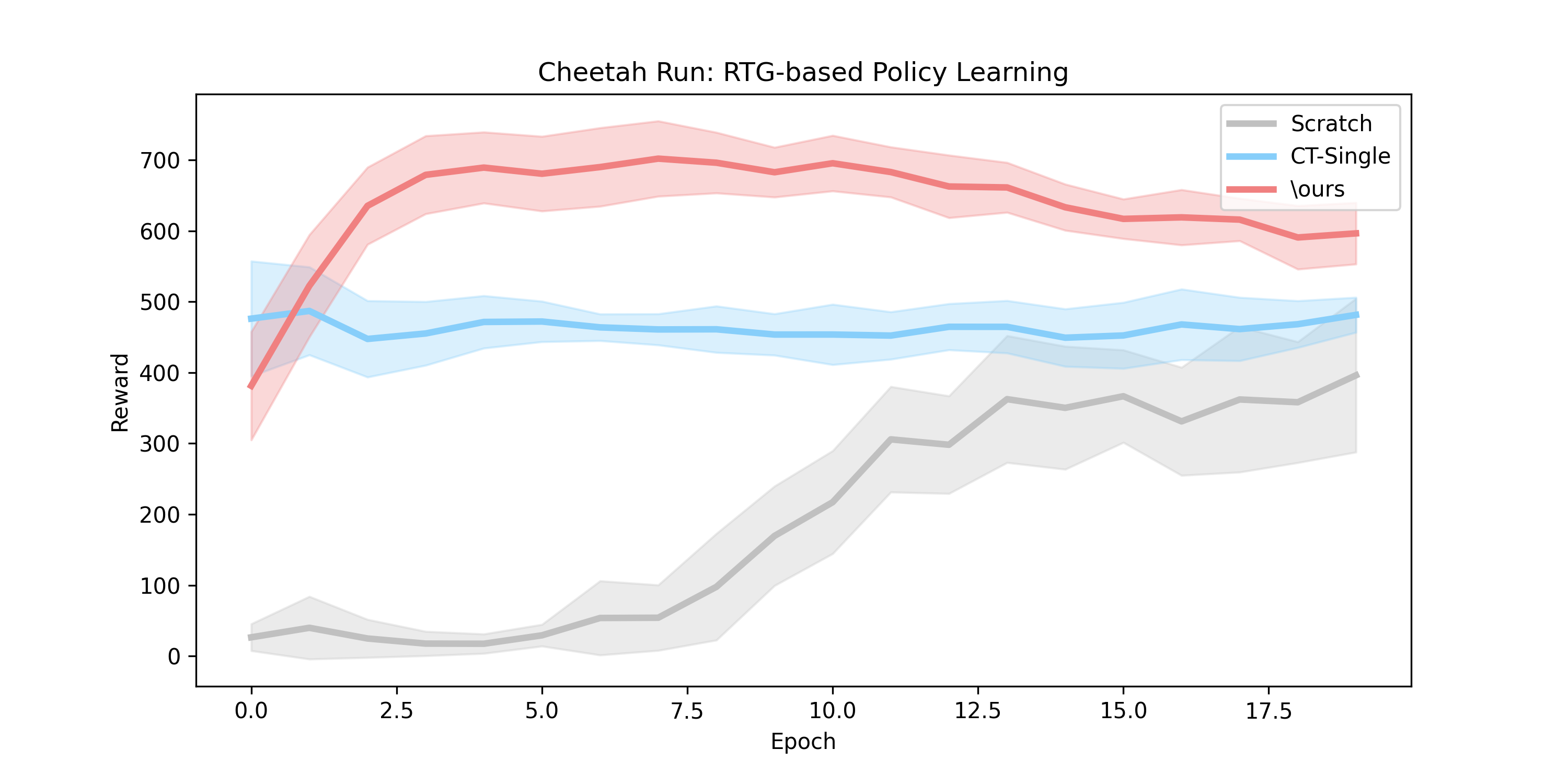}}
%   \vspace{-1.5em}
%   \caption{Disc. \& Adaptive}
 \end{subfigure}
 \hfill
 \begin{subfigure}[t]{0.19\columnwidth}
  \resizebox{\textwidth}{!}{\input{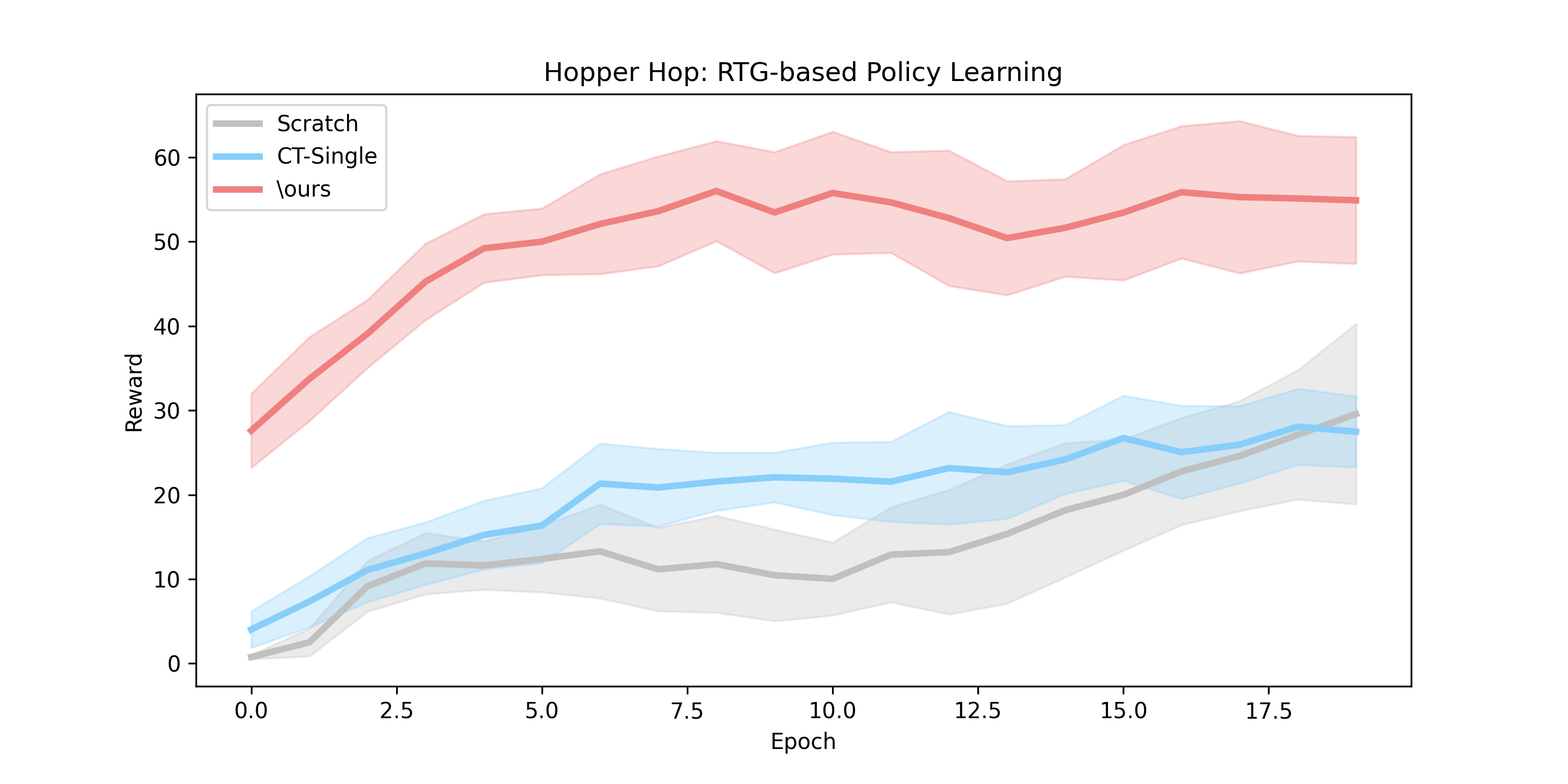}}
%   \vspace{-1.5em}
%   \caption{Cont. \& Non-adaptive}
 \end{subfigure}
 \hfill
 \begin{subfigure}[t]{0.19\columnwidth}
  \resizebox{\textwidth}{!}{\input{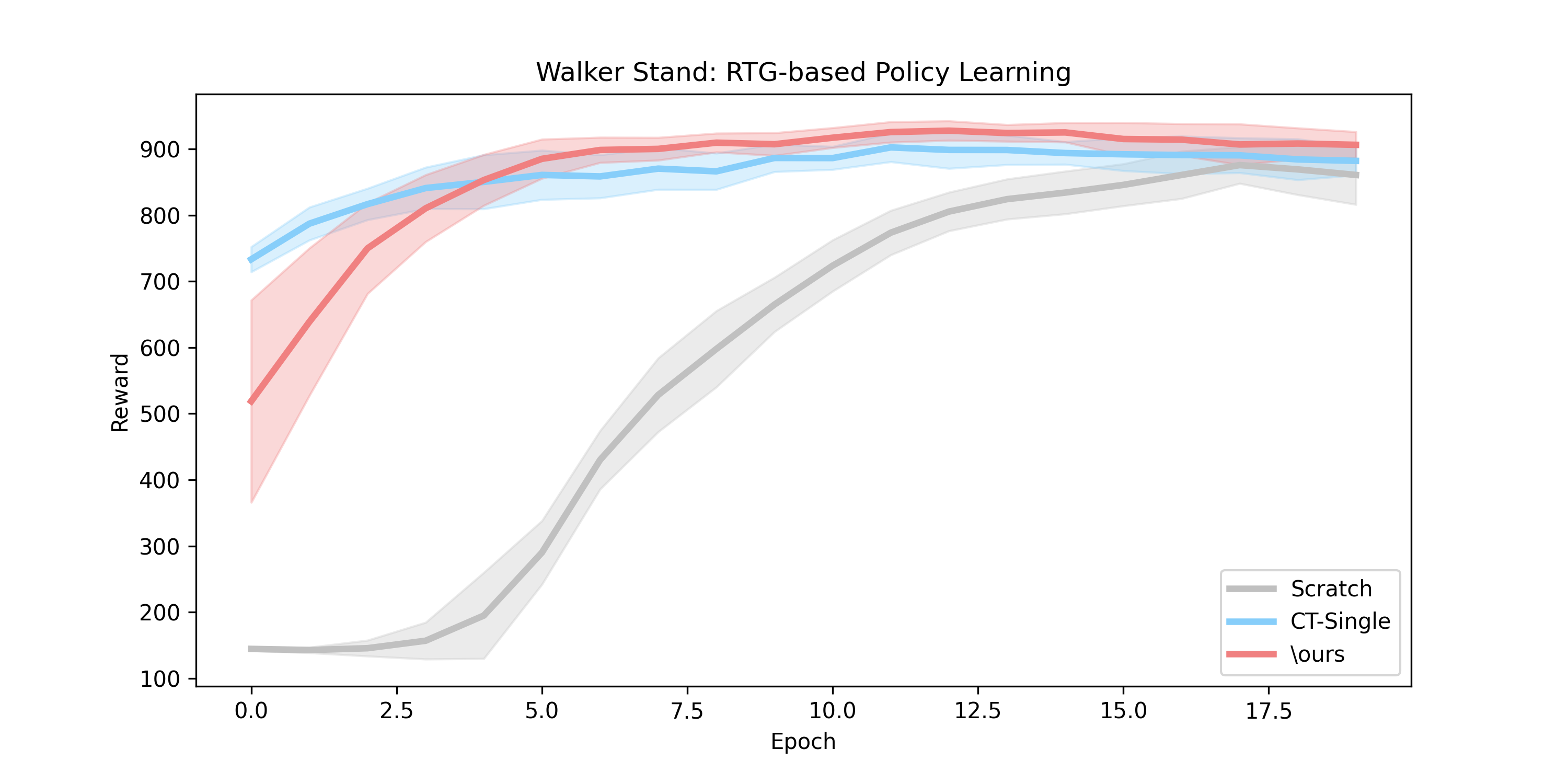}}
%   \vspace{-1.5em}
%   \caption{Cont. \& Adaptive}
 \end{subfigure} 
 \hfill
 \begin{subfigure}[t]{0.19\columnwidth}
  \resizebox{\textwidth}{!}{\input{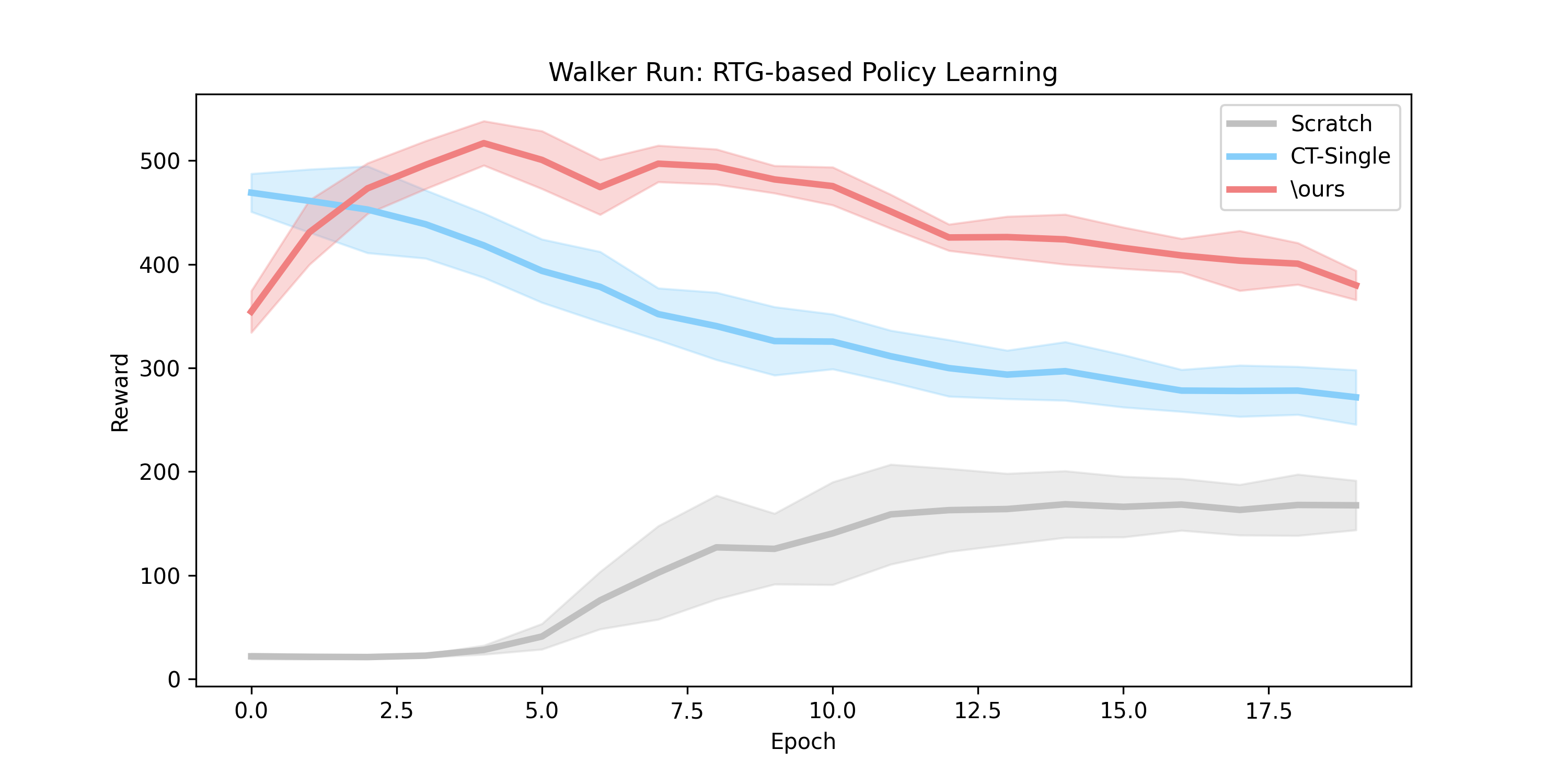}}
%   \vspace{-1.5em}
%   \caption{Cont. \& Adaptive}
 \end{subfigure} 
%  \vspace{0.5em}

\rotatebox{90}{\scriptsize{\hspace{1cm}\textbf{BC}}}
 \begin{subfigure}[t]{0.19\columnwidth}
  \resizebox{\textwidth}{!}{\input{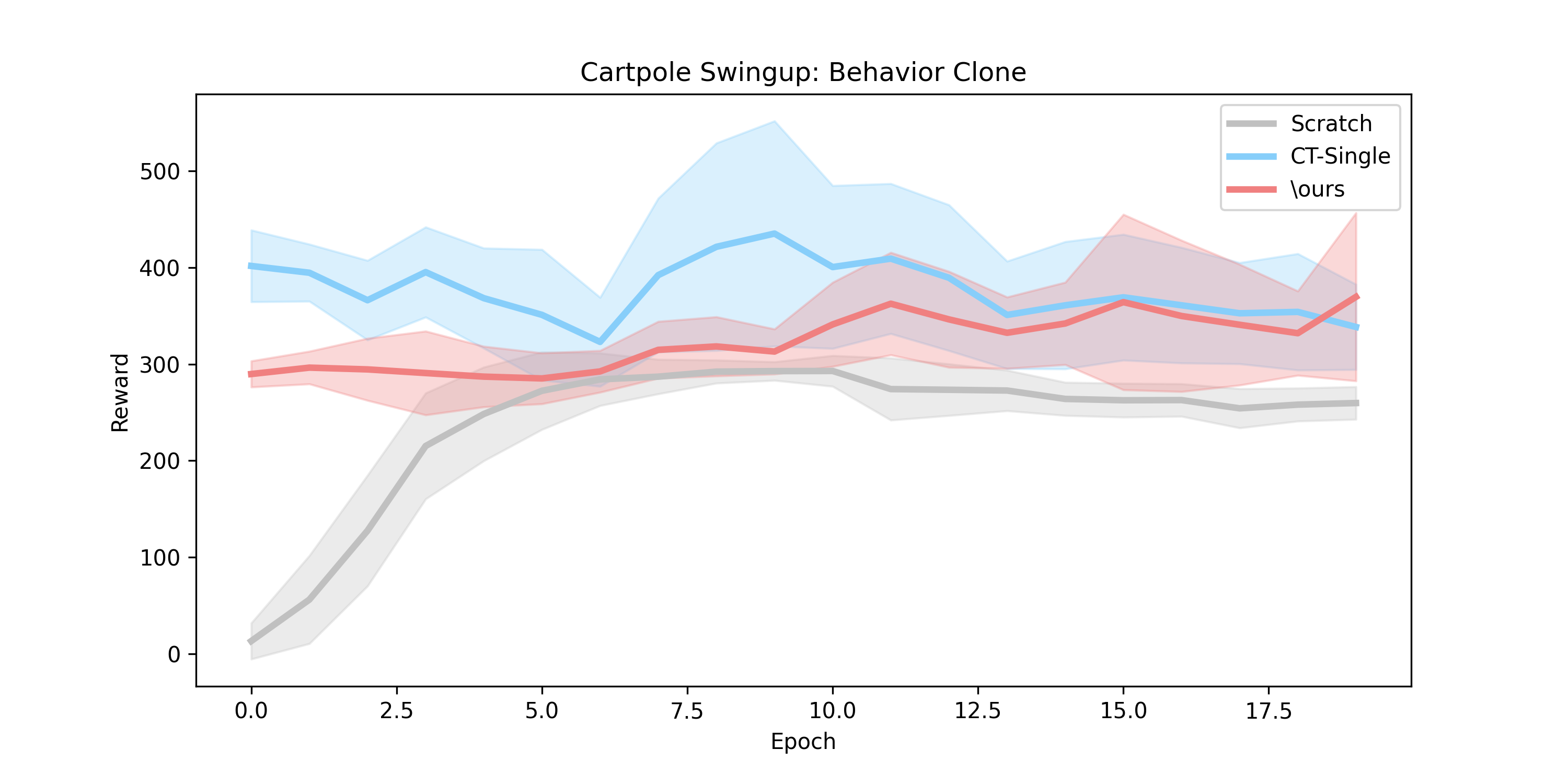}}
%   \vspace{-1.5em}
%   \caption{FoodCollector}
 \end{subfigure}
 \hfill
 \begin{subfigure}[t]{0.19\columnwidth}
  \resizebox{\textwidth}{!}{\input{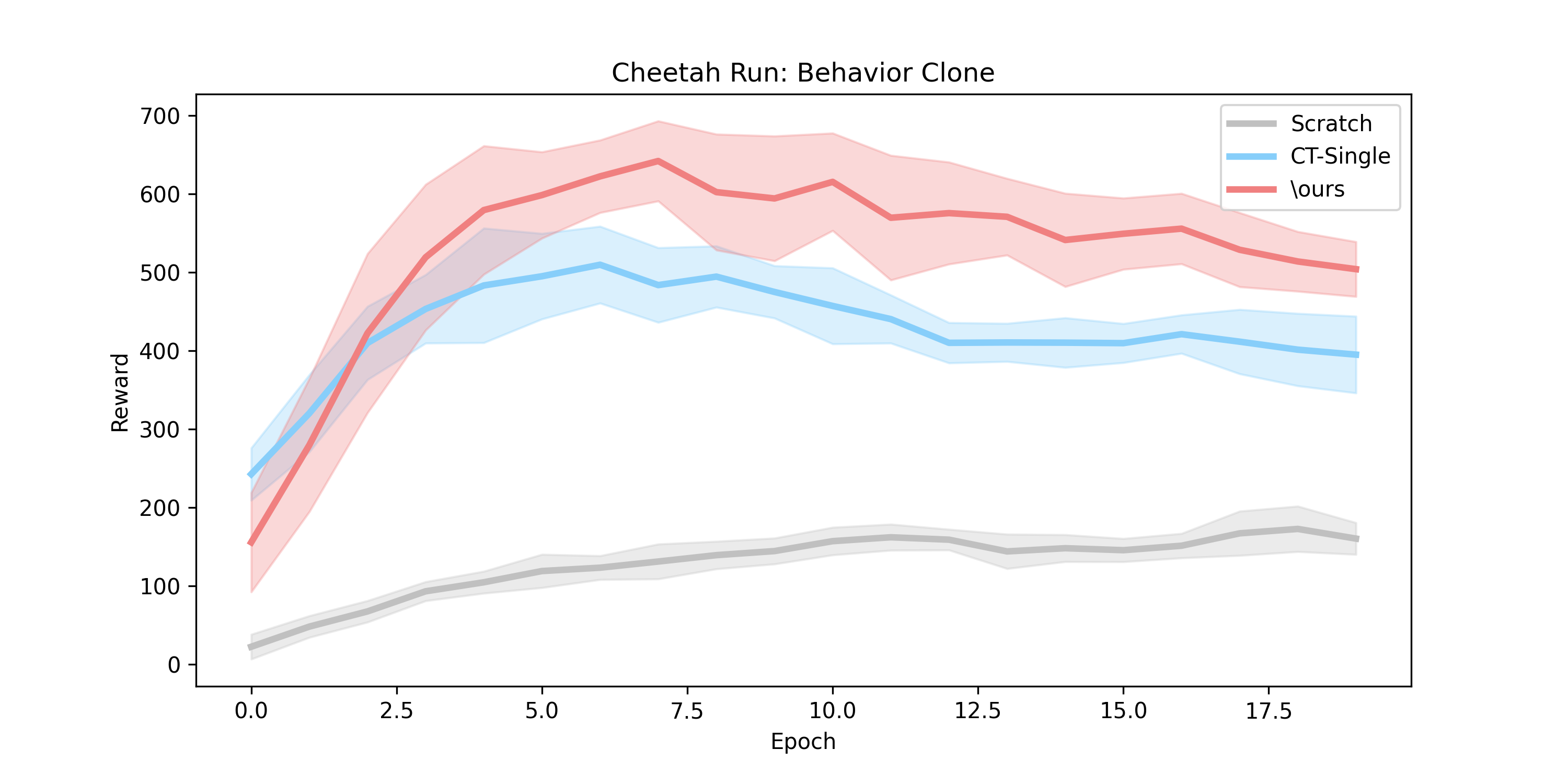}}
%   \vspace{-1.5em}
%   \caption{Disc. \& Adaptive}
 \end{subfigure}
 \hfill
 \begin{subfigure}[t]{0.19\columnwidth}
  \resizebox{\textwidth}{!}{\input{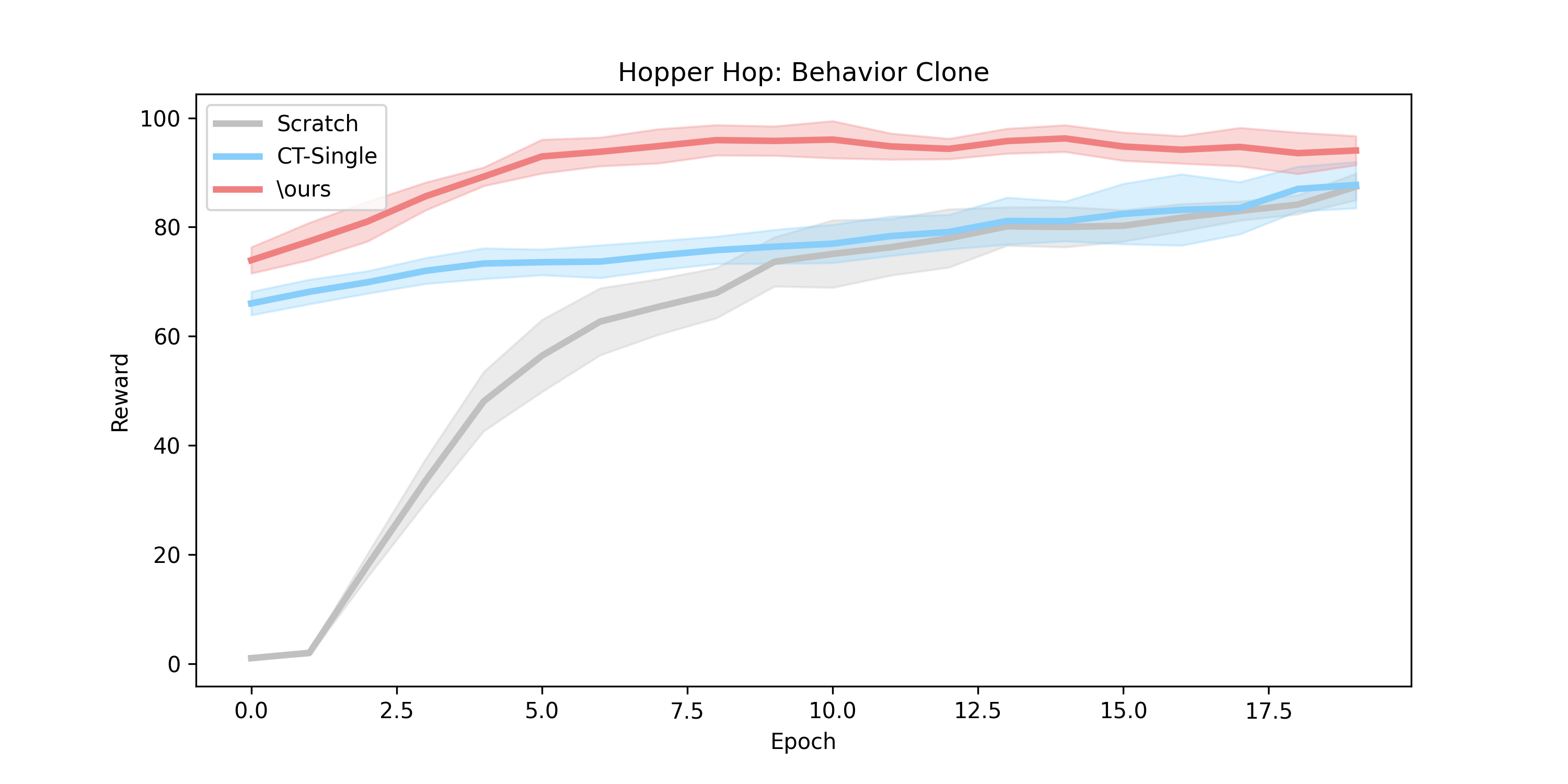}}
%   \vspace{-1.5em}
%   \caption{Cont. \& Non-adaptive}
 \end{subfigure}
 \hfill
 \begin{subfigure}[t]{0.19\columnwidth}
  \resizebox{\textwidth}{!}{\input{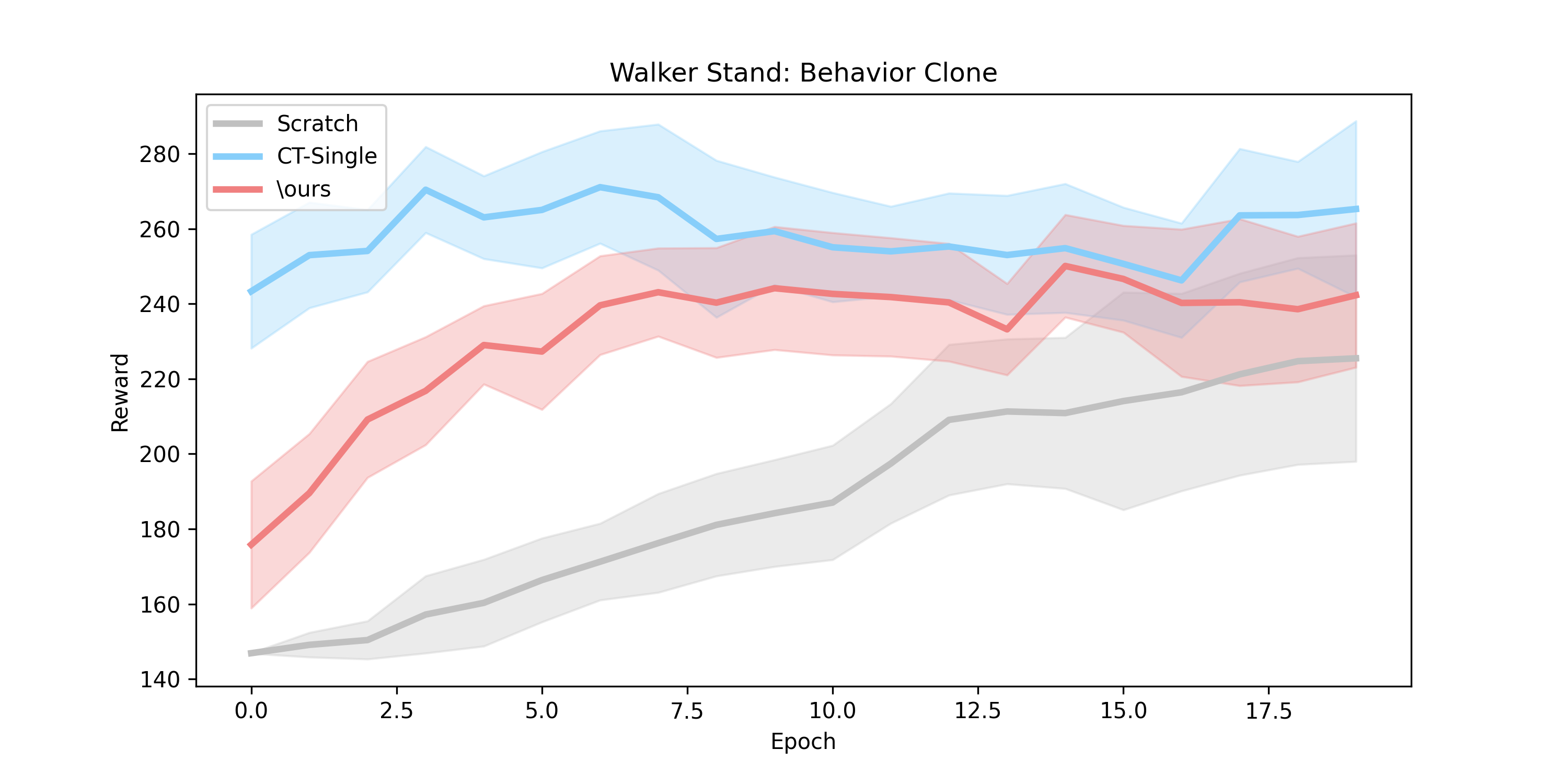}}
%   \vspace{-1.5em}
%   \caption{Cont. \& Adaptive}
 \end{subfigure} 
 \hfill
 \begin{subfigure}[t]{0.19\columnwidth}
  \resizebox{\textwidth}{!}{\input{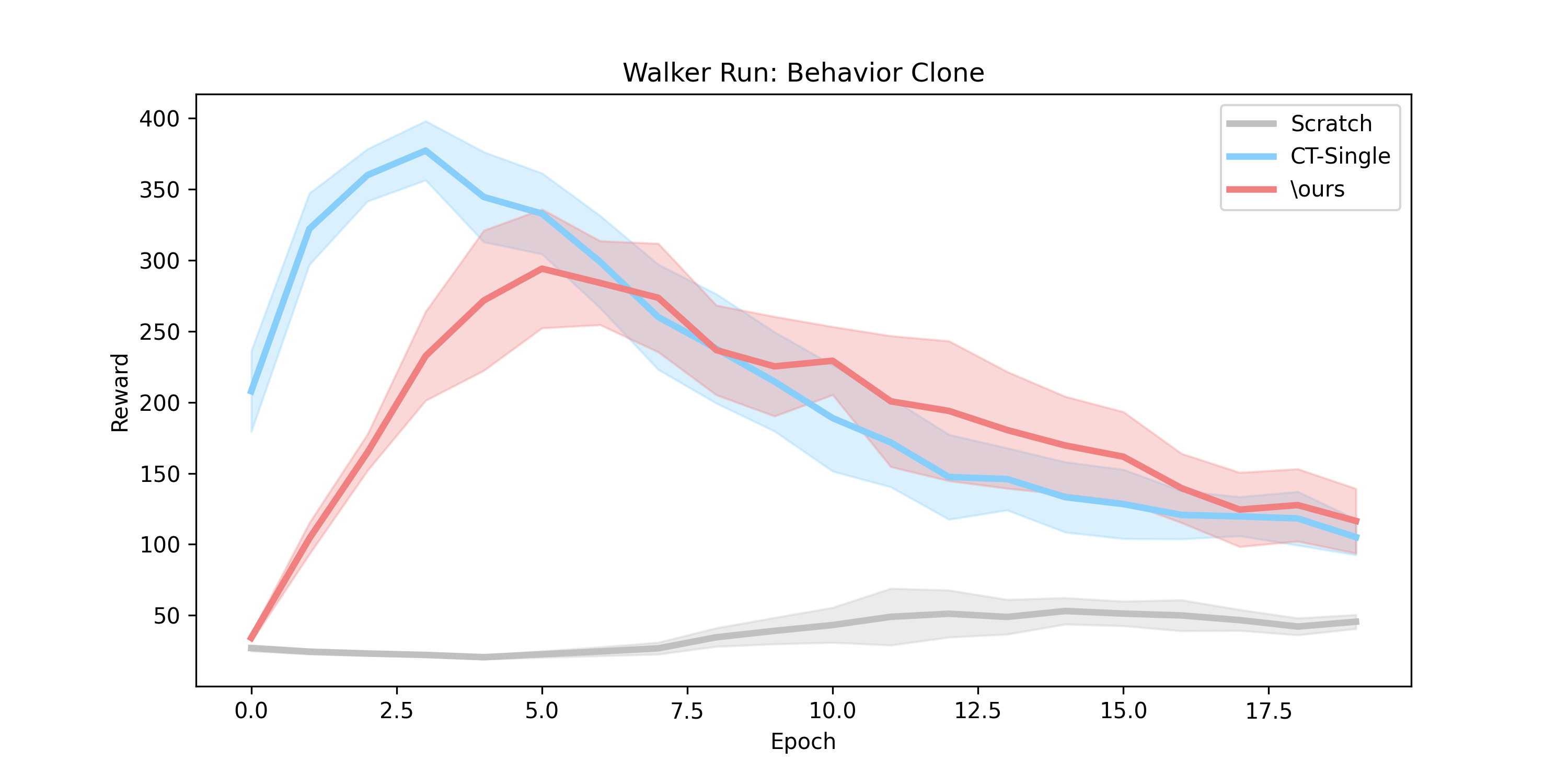}}
%   \vspace{-1.5em}
%   \caption{Cont. \& Adaptive}
 \end{subfigure} 
 \vspace{-1em}
\caption{Downstream learning rewards of \ours (\textcolor{red}{red}) compared with pretraining \ourmod with single-task data (\textcolor{cyan}{blue}) and training from scratch (\textcolor{gray}{gray}). Results are averaged over 3 random seeds.
}
% \vspace{-1em}
\label{fig:curves_seen}
\end{figure}

\begin{figure}[htbp]
\vspace{-0.5em}
 \centering
 \begin{subfigure}[t]{0.5\columnwidth}
  \resizebox{\textwidth}{!}{\input{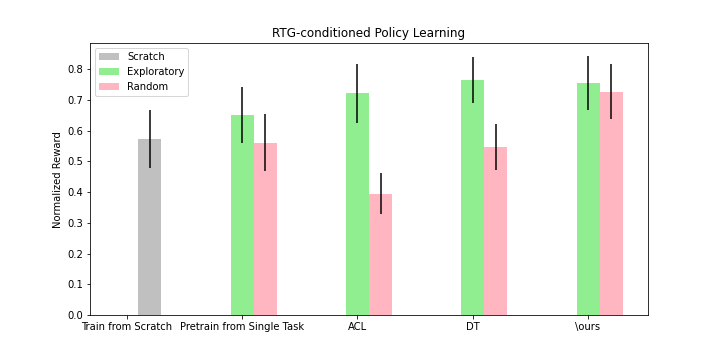}}
 \end{subfigure}
 \hfill
 \begin{subfigure}[t]{0.45\columnwidth}
  \resizebox{\textwidth}{!}{\input{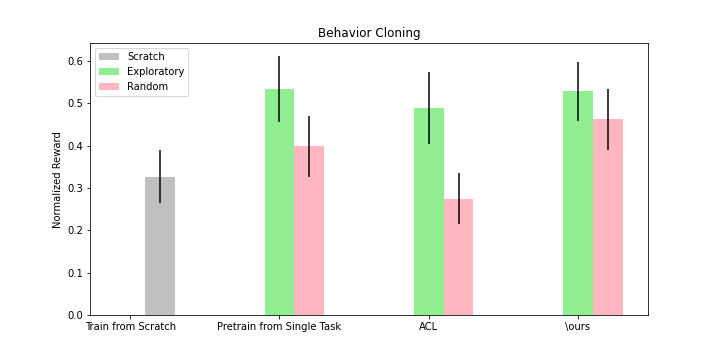}}
 \end{subfigure}
 \vspace{-1em}
\caption{Downstream learning rewards (normalized by expert score) of all methods using \texttt{Exploratory} and \texttt{Random} dataset. The gap between each pair of green and red bars corresponds to the resilience of each method to pretraining data quality, and our \ours shows the best resilience among all baselines.
% to low-quality data among all baselines.
}
% \vspace{-1.5em}
\label{fig:data_baseline}
\end{figure}

% We first evaluate the \textbf{Versatility} of \ours: whether the pretrained model can be used to learn multiple downstream tasks with different learning objectives.
% More specifically, we pretrain \ourmod on 5 tasks: cartpole-swingup, hopper-hop, cheetah-run, walker-stand and walker-run. 
% For downstream tasks, we use the trained policy to interact with each environment for 50 episodes and obtain the average rewards after every training epoch. 
We first evaluate the \textbf{versatility} of \ours: 
1) whether a single pretrained model can be finetuned with different downstream learning methods (i.e. RTG and BC); and 
2) whether the pretrained model can adapt towards various downstream tasks.
% More specifically, 
% To do so, we pretrain \ourmod on 5 tasks: 
% cartpole-swingup, hopper-hop, cheetah-run, walker-stand and walker-run. 
% For downstream evaluation, we use the trained policies to interact with each environment for 50 episodes and report the average rewards after every training epoch. 
\cref{fig:curves_seen} compares the the reward curve of \ours with \texttt{Scratch} and \texttt{CT-Single}, where models are pretrained with \texttt{Exploratory} dataset\footnote{\label{note1}Models pretrained with \texttt{Random} dataset show similar results, as can be seen in \cref{app:exp_full}}.
% Models pretrained with \texttt{Random} dataset show similar trends, as can be seen in \cref{app:exp_full}. 
To avoid overlapping of curves, comparison with \texttt{ACL} and \texttt{DT} is shown and discussed later in \cref{fig:data_baseline}.
It can be seen that pretrained CT from both single-task dataset (\texttt{CT-single}) and multi-task dataset (\ours) can achieve much better results than training from scratch. 
In general, under both RTG and BC finetuning, pretrained models have a warm start, a faster convergence rate, and a relatively better asymptotic performance in a variety of downstream tasks. 
In most cases, pretraining \ourmod from multi-task dataset (\ours) yields better results than pretraining with only in-task data (\texttt{CT-single}), although it is harder to accommodate multiple different tasks with the same model capacity, which suggests that \ours can extract common knowledge from diverse tasks.

\begin{figure}[t]
\vspace{-1em}
 \centering

\rotatebox{90}{\scriptsize{\hspace{1cm}\textbf{RTG}}}
 \begin{subfigure}[t]{0.19\columnwidth}
  \resizebox{1.01\textwidth}{!}{\input{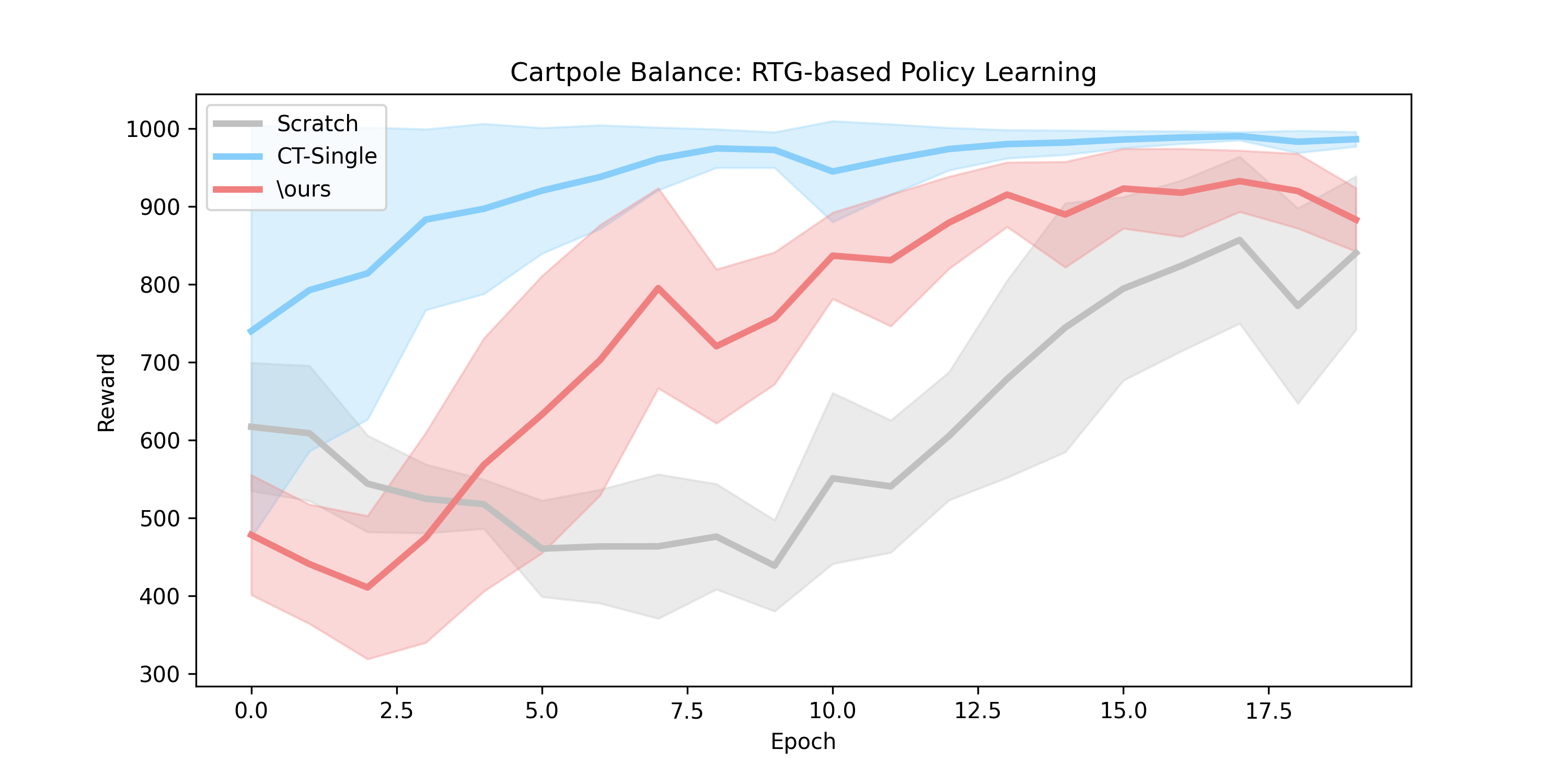}}
%   \vspace{-1.5em}
%   \caption{FoodCollector}
 \end{subfigure}
 \hfill
 \begin{subfigure}[t]{0.19\columnwidth}
  \resizebox{\textwidth}{!}{\input{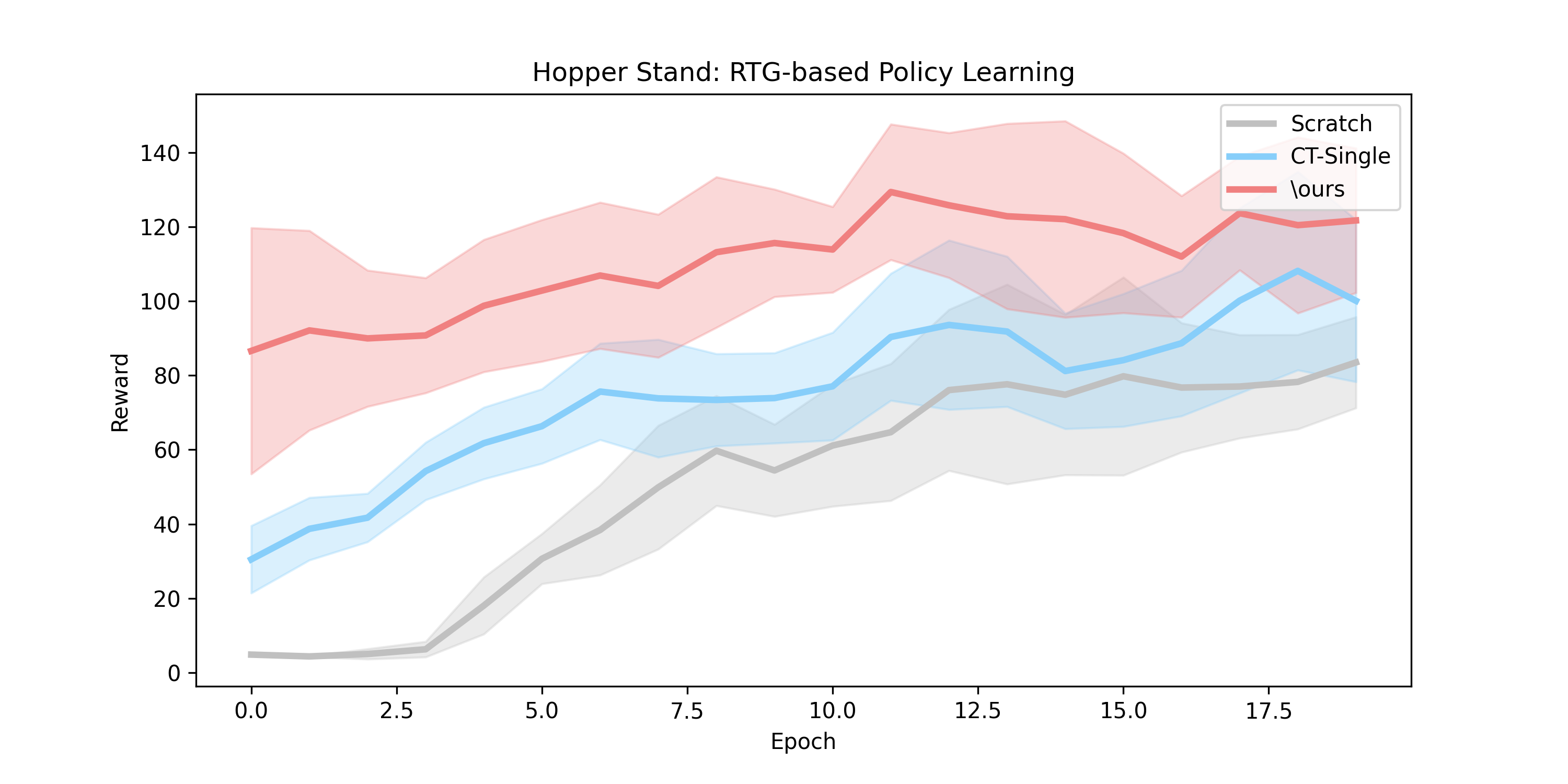}}
%   \vspace{-1.5em}
%   \caption{Disc. \& Adaptive}
 \end{subfigure}
 \hfill
 \begin{subfigure}[t]{0.19\columnwidth}
  \resizebox{\textwidth}{!}{\input{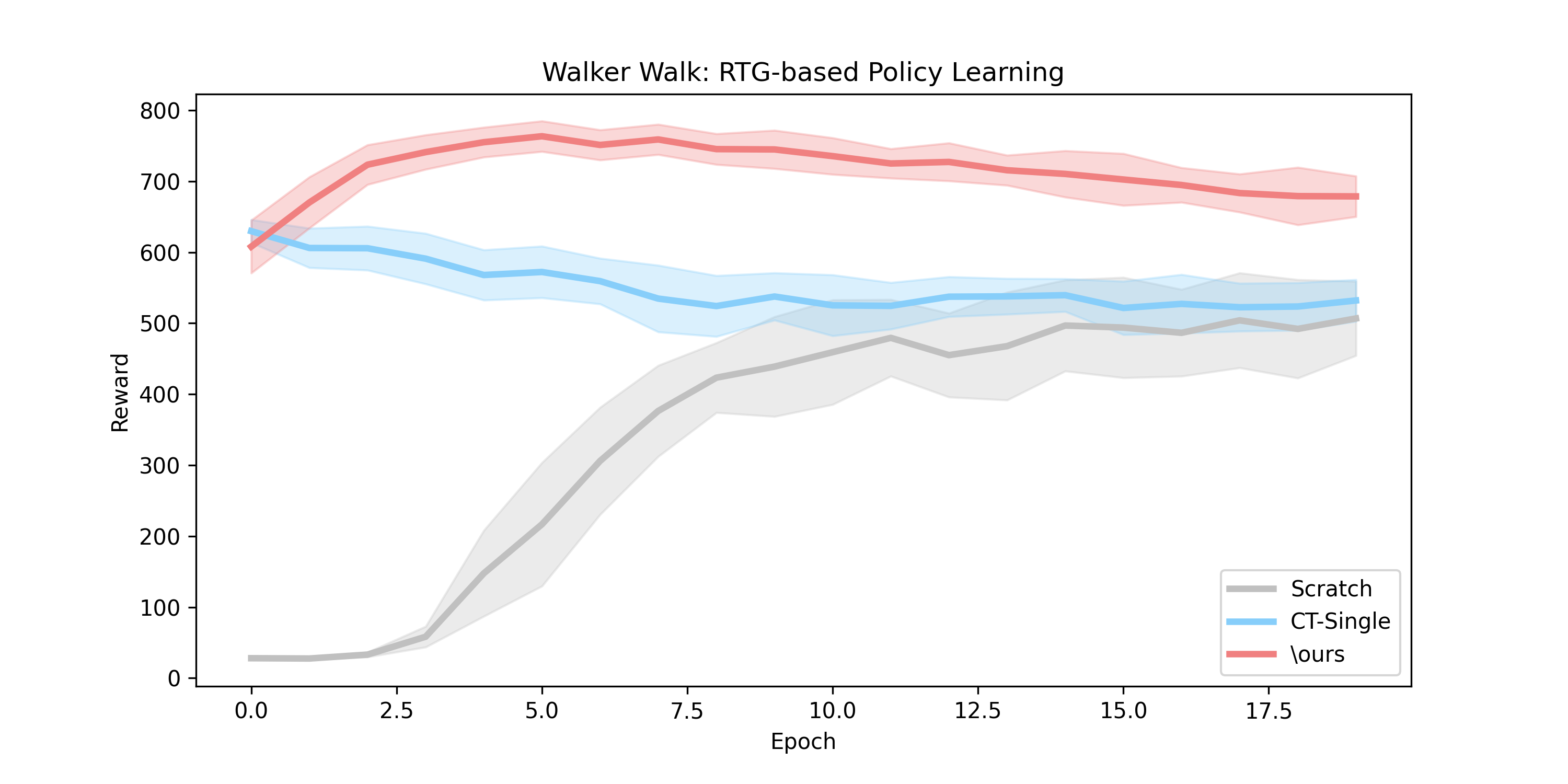}}
%   \vspace{-1.5em}
%   \caption{Cont. \& Non-adaptive}
 \end{subfigure}
 \hfill
 \begin{subfigure}[t]{0.19\columnwidth}
  \resizebox{\textwidth}{!}{\input{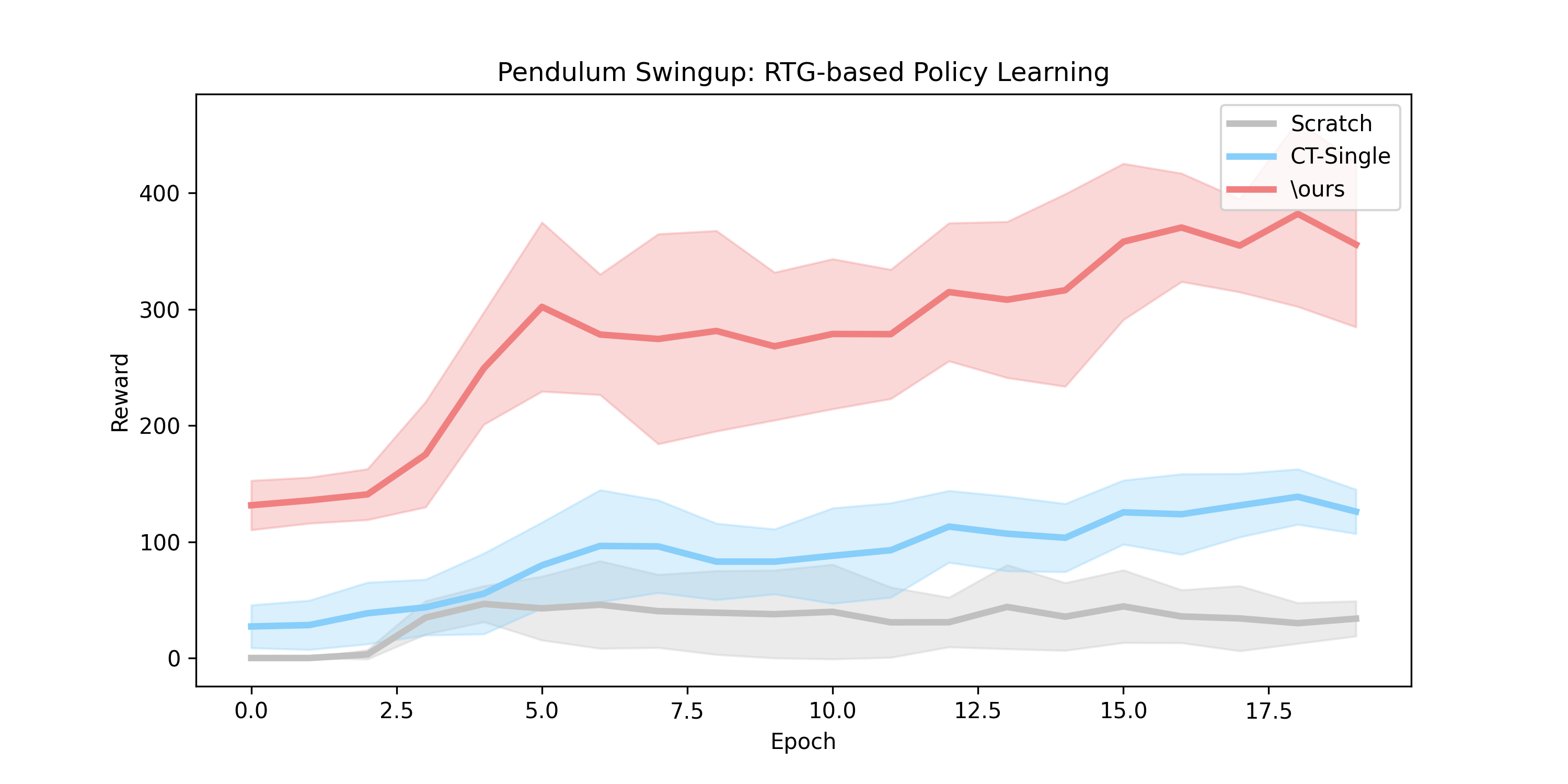}}
%   \vspace{-1.5em}
%   \caption{Cont. \& Adaptive}
 \end{subfigure} 
 \hfill
 \begin{subfigure}[t]{0.19\columnwidth}
  \resizebox{\textwidth}{!}{\input{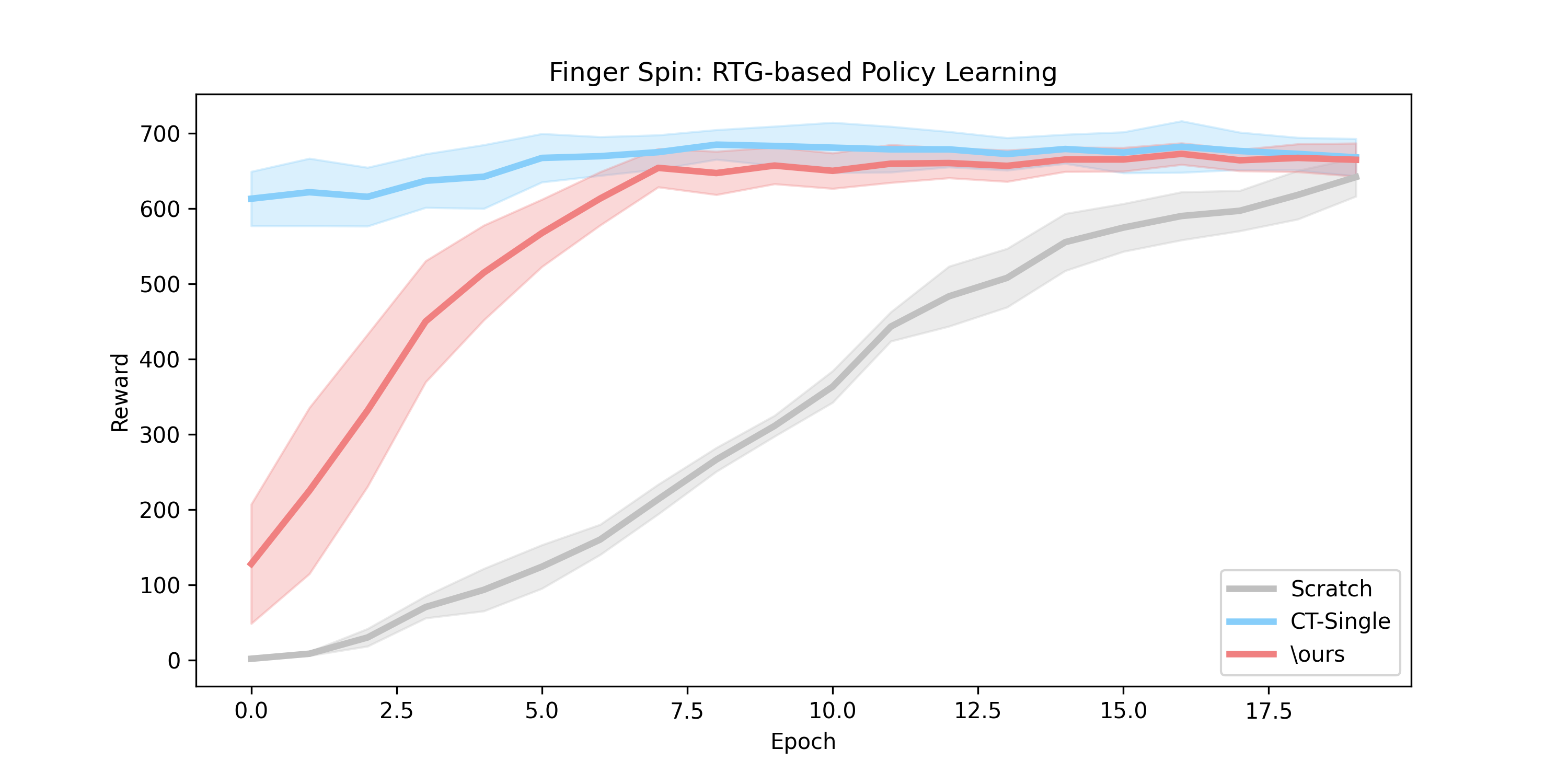}}
%   \vspace{-1.5em}
%   \caption{Cont. \& Adaptive}
 \end{subfigure} 
%  \vspace{0.5em}

\rotatebox{90}{\scriptsize{\hspace{1cm}\textbf{BC}}}
 \begin{subfigure}[t]{0.19\columnwidth}
  \resizebox{1.01\textwidth}{!}{\input{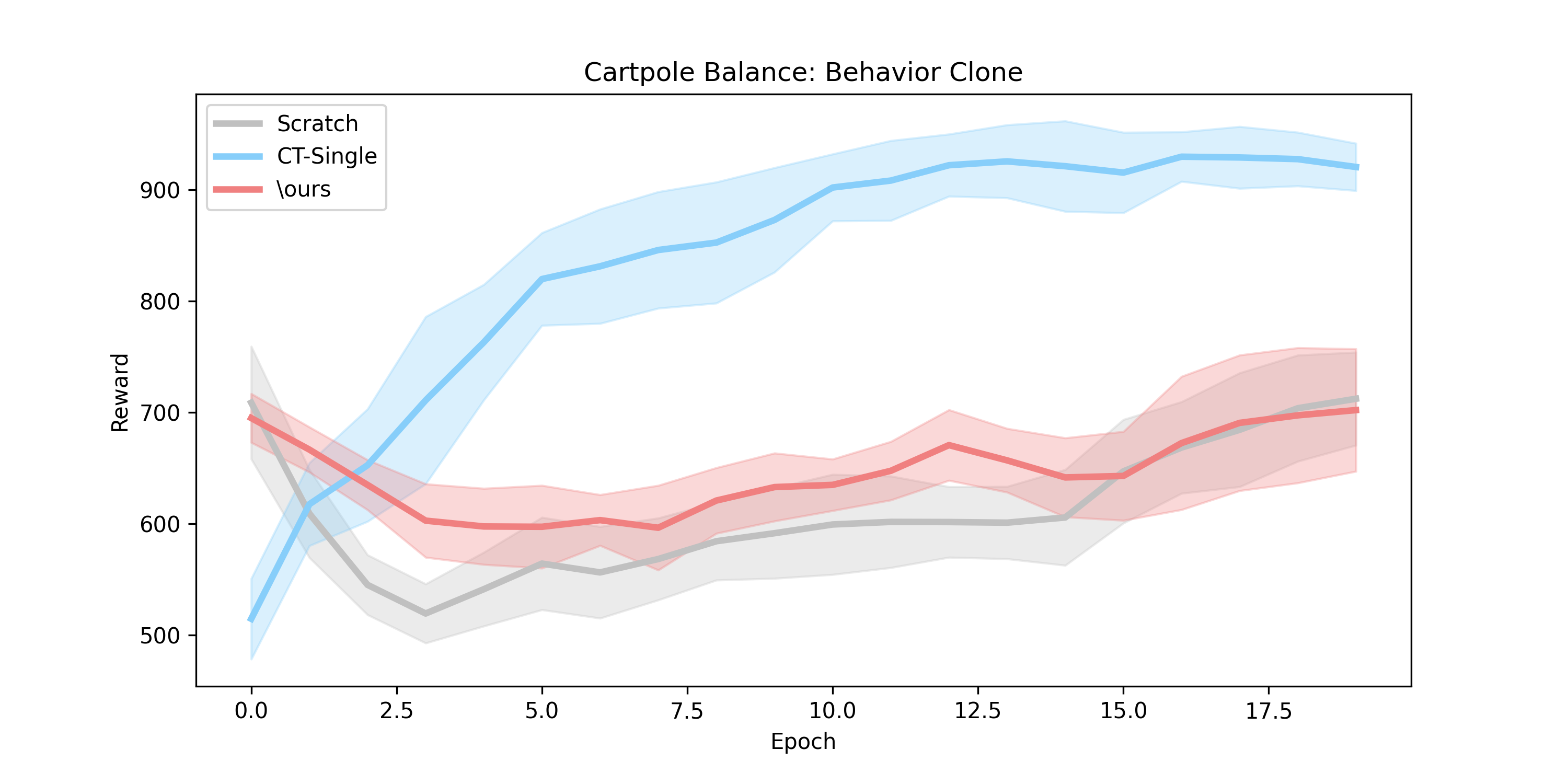}}
%   \vspace{-1.5em}
%   \caption{FoodCollector}
 \end{subfigure}
 \hfill
 \begin{subfigure}[t]{0.19\columnwidth}
  \resizebox{\textwidth}{!}{\input{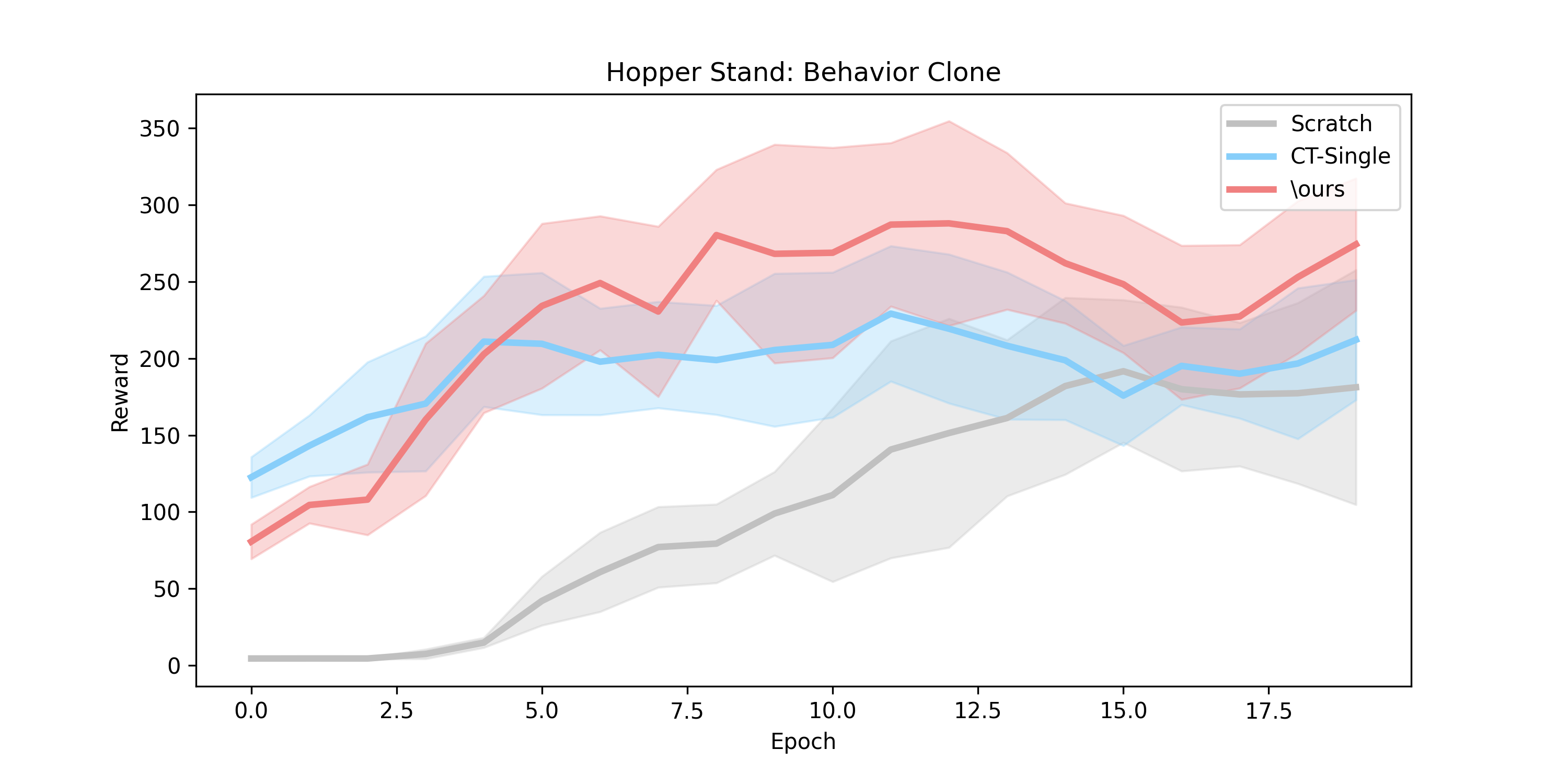}}
%   \vspace{-1.5em}
%   \caption{Disc. \& Adaptive}
 \end{subfigure}
 \hfill
 \begin{subfigure}[t]{0.19\columnwidth}
  \resizebox{\textwidth}{!}{\input{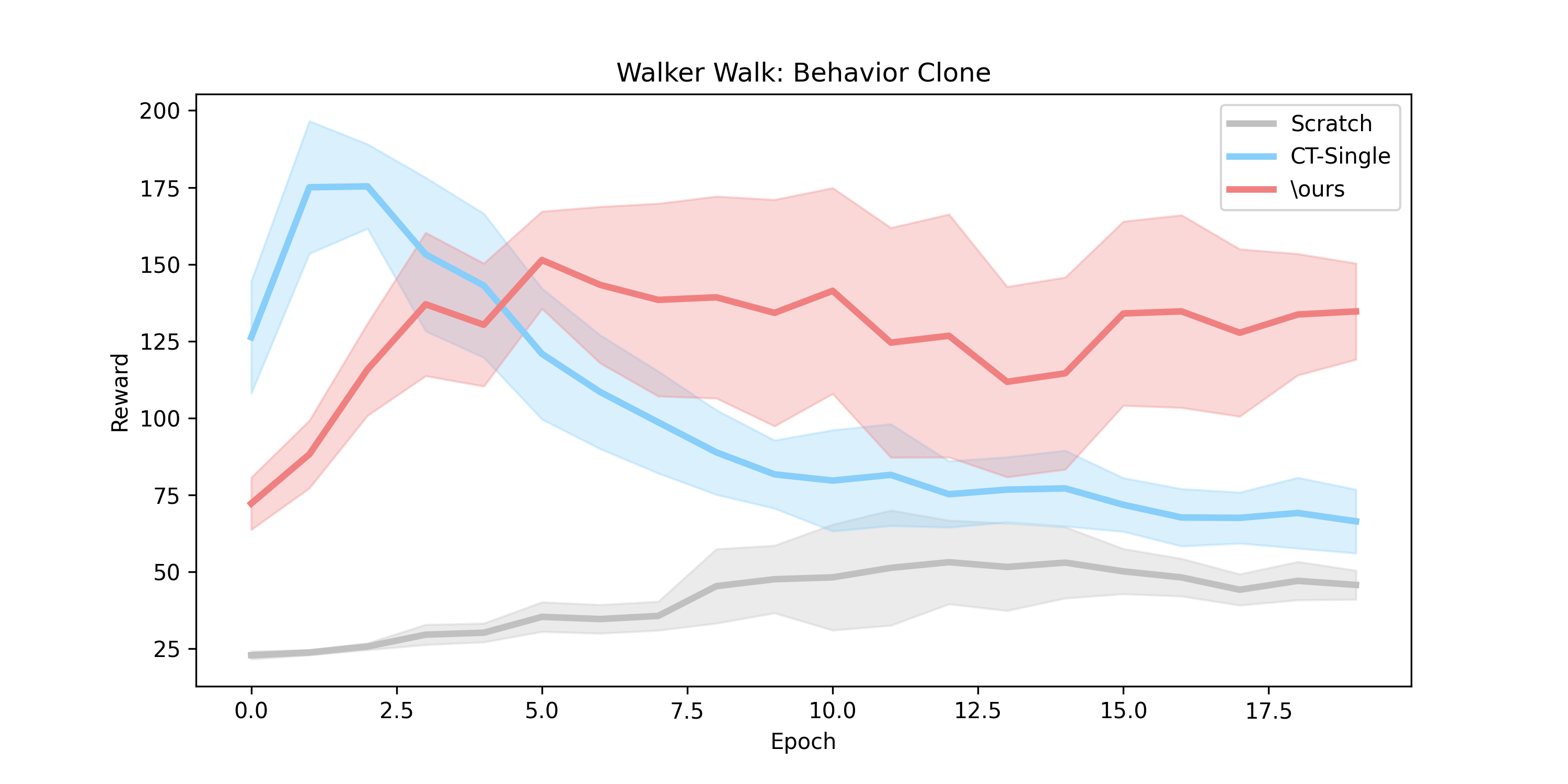}}
%   \vspace{-1.5em}
%   \caption{Cont. \& Non-adaptive}
 \end{subfigure}
 \hfill
 \begin{subfigure}[t]{0.19\columnwidth}
  \resizebox{\textwidth}{!}{\input{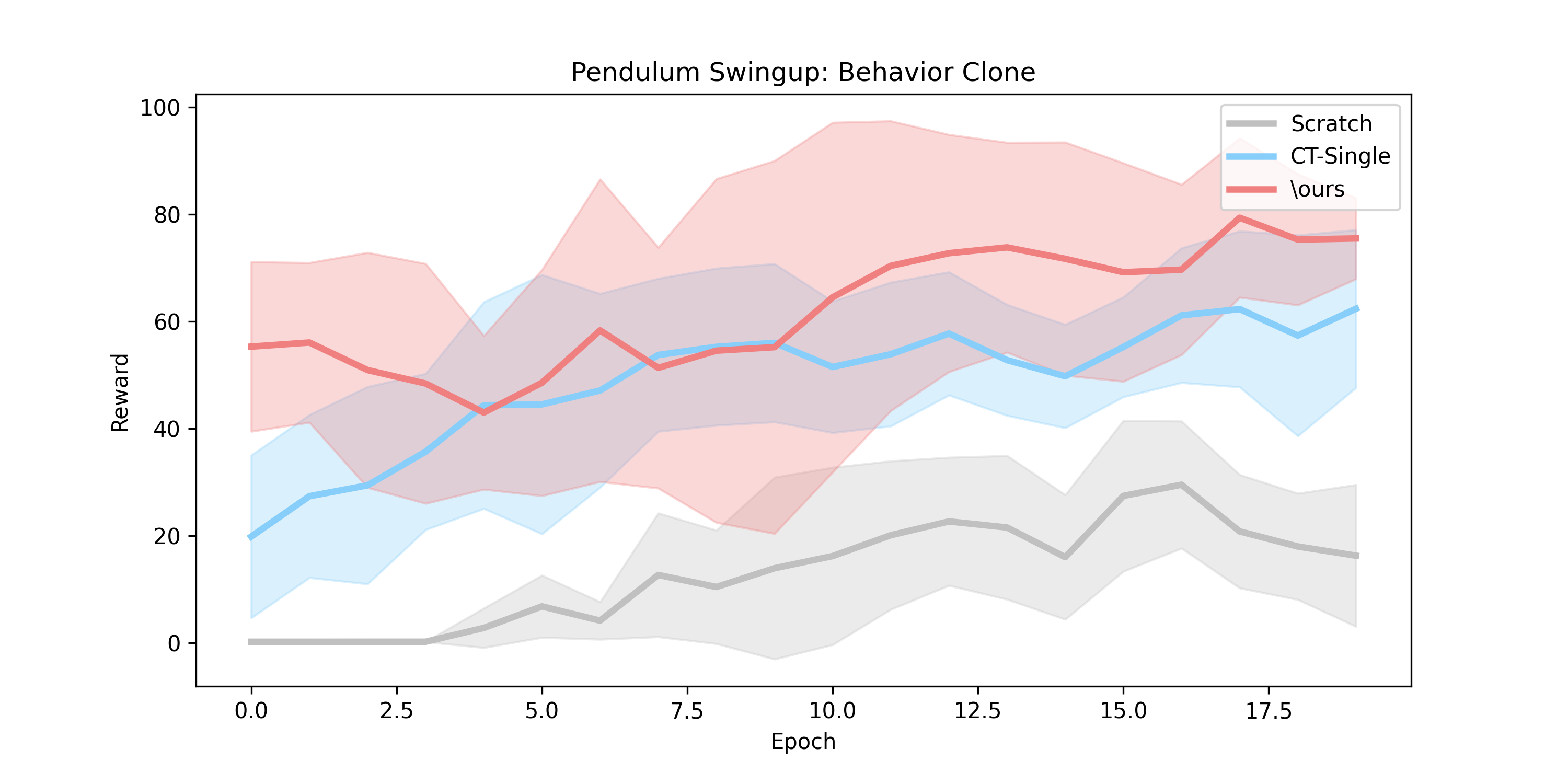}}
%   \vspace{-1.5em}
%   \caption{Cont. \& Adaptive}
 \end{subfigure} 
 \hfill
 \begin{subfigure}[t]{0.19\columnwidth}
  \resizebox{\textwidth}{!}{\input{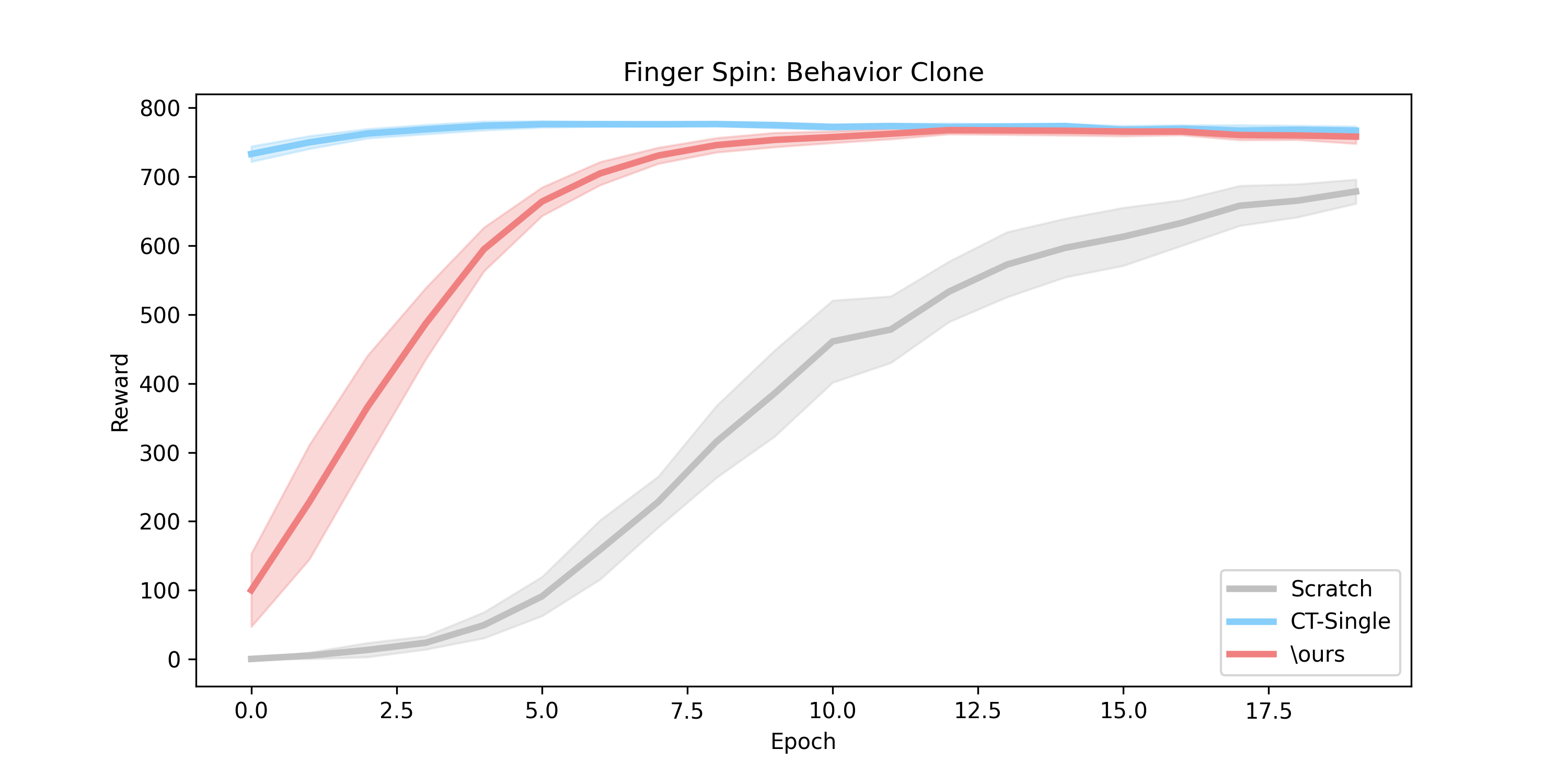}}
%   \vspace{-1.5em}
%   \caption{Cont. \& Adaptive}
 \end{subfigure} 
 \vspace{-1em}
\caption{Downstream learning rewards \textbf{in unseen tasks and domains} of \ours (\textcolor{red}{red}) compared with pretraining \ourmod with single-task data (\textcolor{cyan}{blue}) and training from scratch (\textcolor{gray}{gray}). Results are averaged over 3 seeds.
}
\vspace{-3mm}
\label{fig:curves_unseen}
\end{figure}

% \begin{wrapfigure}{r}{0.3\textwidth}
% \vspace{-2em}
%   \resizebox{0.3\textwidth}{!}{\input{figures/data_baseline/data_single_rtg}}
%   \vspace{-2em}
%   \caption{Normalized reward averaged across downstream RTG-conditioned tasks, with exploratory dataset and random dataset for pretraining.}
%   \label{fig:data_single}
% \vspace{-1em}
% \end{wrapfigure}
% \cref{fig:curves_seen} shows that pretraining \ourmod from multi-task dataset usually yields better results than pretraining with only in-task data, although different tasks have different state/action spaces and dynamics.
% This result suggests that \ours extracts common knowledge from different tasks. More importantly, when the pretraining phase is based on the lower-quality Random dataset, pretraining from single-task data becomes even worse than training from scratch due to possible overfitting to the pretraining data distribution. On the contrary, \ours still maintains a high performance, verifying the \textbf{resistance} of \ours to distribution shifts.

Next, we show the \textbf{generalizability} of \ours. 
\cref{fig:curves_unseen} shows the performance of \ours pretrained on \texttt{Exploratory} dataset$^{\ref{note1}}$,
% \footnote{Results of the models pretrained with \texttt{Random} dataset are similar as shown in in \cref{app:exp_full}.}
compared to \texttt{Scratch} and \texttt{CT-single} on 5 unseen tasks.
% : cartpole-balance, hopper-stand, walker-walk, pendulum-swingup and finger-spin. 
% Note that last two tasks are from unseen domains with state-action spaces different from tasks in the pretraining dataset.
We can see that \ours is able to generalize to unseen tasks and even unseen domains, whose distributions have a larger discrepancy as compared to the pretraining dataset. 
Surprisingly, \ours achieves better performance than \texttt{CT-single} in most tasks, even though \texttt{CT-single} has already seen the downstream environments. 
% It suggests that the good generalization ability can benefit from learning underlying common knowledge among a diverse set of distribution.
This suggests that good generalization ability can be obtained from learning underlying information which might be shared among multiple tasks and domains, spanning a diverse set of distributions. 
\cref{app:exp_full} provides additional results in other challenging tasks which have a larger discrepancy with pretraining tasks, where \ours still performs well in such challenging cases.
% provides additional results in other challenging tasks which are less similar to pretraining tasks, where \ours still performs well.

Next we evaluate the \textbf{resilience} of \ours by comparing with all aforementioned baselines, as visualized in \cref{fig:data_baseline}. 
We aggregate the results in all tasks by averaging the normalized reward (dividing raw scores by expert scores) in both RTG and BC settings. 
When using the \texttt{Exploratory} dataset for pretraining, \ours outperforms \texttt{ACL}, and is comparable to \texttt{DT} which has extra information of reward. 
When pretrained with the \texttt{Random} dataset, \ours is significantly better than \texttt{DT} and \texttt{ACL}, while \texttt{ACL} fails to outperform training from scratch. 
This result show that \ours is robust to low-quality data as compared to other baseline methods.
% This result validates the challenge of distribution shift, while \ours is much more robust to such shift than baseline methods.

% We can see that \ours can tolerate varying-quality of pretraining data and large distribution shift between pretraining data and downstream data. We demonstrate this advantage by comparing the results of using relatively good exploratory dataset and using low-quality random dataset. We 

\vspace{-3mm}
\subsection{Ablation and Discussion}
\label{ssec:exp_abl}
\vspace{-1mm}

\begin{figure}[htbp]
    \centering
    \begin{subfigure}[t]{0.48\columnwidth}
        \resizebox{\textwidth}{!}{\input{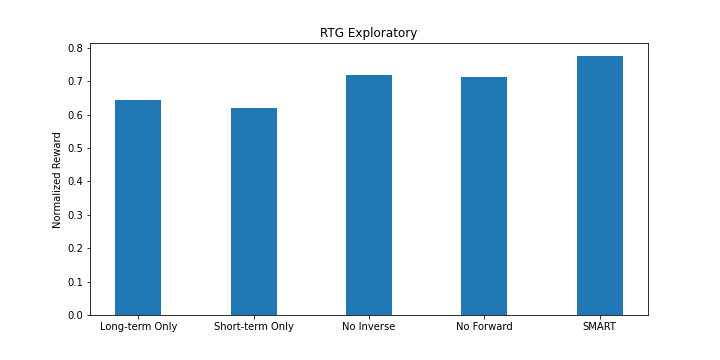}}
        % \caption{Ablation of objectives.}
    \end{subfigure}
    \hfill
    \begin{subfigure}[t]{0.48\columnwidth}
        \resizebox{\textwidth}{!}{\input{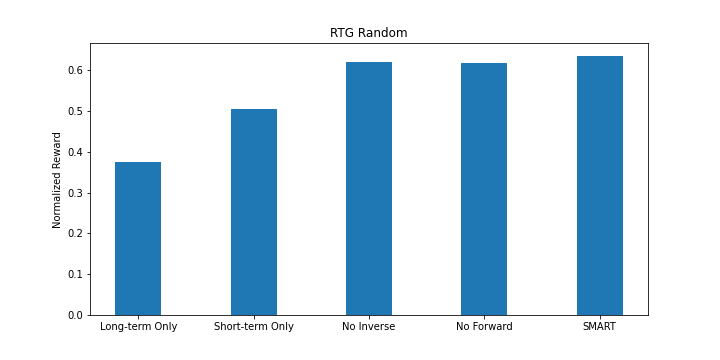}}
        % \caption{Ablation of long-term and short-term objectives.}
    \end{subfigure}
    \vspace{-3mm}
    \caption{Ablation study on proposed pretraining objective. Rewards are averaged over tasks. Both long-term control information (\randinvshort) and short-term control information (Forward and Inverse) are important.}
    \label{fig:ablation}
    % \vspace{-3mm}
\end{figure}

\begin{figure}[htbp]
\centering
\begin{minipage}{.48\textwidth}
  \centering
  \begin{subfigure}[t]{0.45\columnwidth}
  \resizebox{\textwidth}{!}{\input{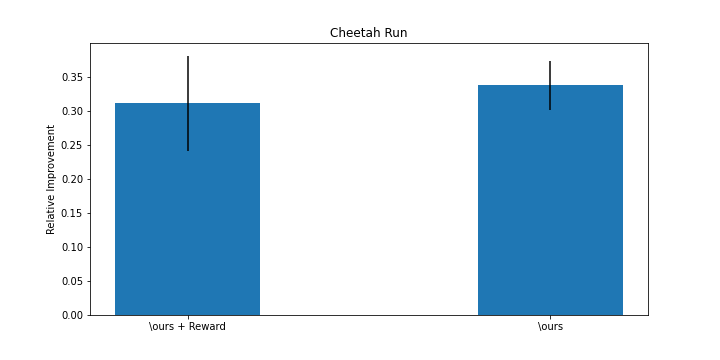}}
 \end{subfigure}
 \hfill
 \begin{subfigure}[t]{0.45\columnwidth}
  \resizebox{\textwidth}{!}{\input{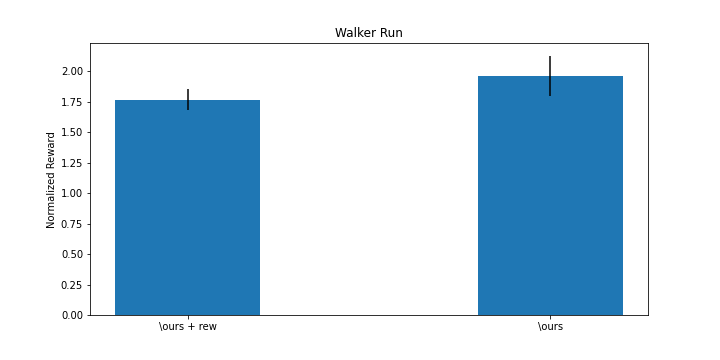}}
 \end{subfigure}
%   \captionof{figure}{Comparison betwen reward-based pretraining and reward-free pretraining.}
\vspace{-0.5em}
  \subcaption{Reward-based v.s. reward-free pretraining.}
  \label{fig:reward_ablation}
\end{minipage}%
\hfill
\begin{minipage}{.48\textwidth}
  \centering
  \begin{subfigure}[t]{0.45\columnwidth}
  \resizebox{\textwidth}{!}{\input{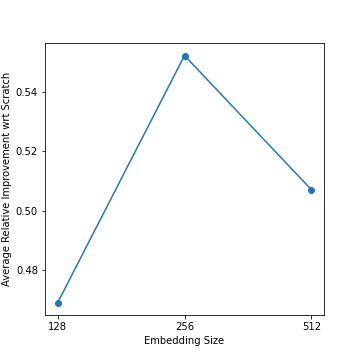}}
 \end{subfigure}
 \hfill
 \begin{subfigure}[t]{0.45\columnwidth}
  \resizebox{\textwidth}{!}{\input{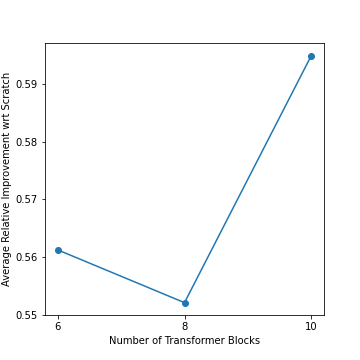}}
 \end{subfigure}
 \vspace{-0.5em}
%   \captionof{figure}{Relationship between overall performance and model capacity. }
  \subcaption{Overall performance v.s. model capacity.}
  \label{fig:capacity_line}
\end{minipage}
\vspace{-3mm}
\caption{Ablation study of the effect of reward in pretraining and comparison of various model capacity.}
\vspace{-1.5em}
\end{figure}

\textbf{Ablation of Pretraining Objectives.}
As we analyzed in \cref{ssec:obj}, forward prediction and inverse prediction aim to learn the information of short-term control, while the \randinv (in short, \randinvshort) learns the long-term control information. To show the effectiveness of combining these two kinds of information, we conduct ablation study over these three terms. 
\cref{fig:ablation} demonstrates their relative improvements wrt \texttt{Scratch} as defined in \cref{app:imple_metric}. 
We can see that only learning long-term control information (Only \randinvshort) or only learning short-term control information (w/o \randinvshort) renders much lower performance than the original \ours that leverages both types of information. 
In addition, we find that both the performance slightly drops if either forward prediction or inverse prediction is removed, as the combination of these two terms leads to more stable and comprehensive representation for short-term control. This ablation study verifies the effectiveness of each component in the proposed pretraining objective. More results in other settings are provided in \cref{app:exp_ablation_obj}.

% To investigate the effectiveness of each term of our proposed objective, we conduct ablation studies over three of them. 
% We evaluate \ours pretrained with ablated objectives in two downstream tasks: cheetah-run (seen) and walker-walk (unseen). 
% \cref{fig:ablation} demonstrates their relative improvements wrt \texttt{Scratch} as defined in \cref{app:imple_metric}.
% According to \cref{sfig:abla_all_cheetah} and \cref{sfig:abla_all_walker}, we can see that forward prediction and \randinv are both functional. 
% Although removing the inverse prediction does not make much difference with exploratory data, it is important when pretraining with random data as suggested by \cref{sfig:abla_inv_cheetah} and \cref{sfig:abla_inv_walker}. 
% This is because the inverse prediction helps the model understand per-step transition dynamics, which is independent of the behavior policy and thus can better tolerate distribution shift.
\vspace{-1mm}

\textbf{Reward-free v.s. Reward-based Pretraining.}
As discussed in \cref{ssec:overview}, including reward information in pretraining objectives is not necessarily helpful. We study the effects of rewards in pretraining by adding two reward-based objectives in the pretraining phase: immediate reward prediction and RTG-based action prediction. We evaluate this reward-based variant of \ours using exploratory dataset and RTG-conditioned downstream learning. Note that the RTG-based action prediction is used both in pretraining and finetuning of the reward-based variant, providing strong supervision for the pretrained model. However, we (surprisingly) observe that such supervision does not improve the downstream performance in many tasks, as shown in \cref{fig:reward_ablation}. A potential reason is that reward-based objectives are more fragile to distribution shifts, which also explains the non-ideal performance of DT pretrained from random data.

\textbf{Discussion on Model Capacity.}
In large-scale training problems, performance usually benefits from larger model capacity~\citep{kaplan2020scaling}. 
We investigate if this also applies to sequential decision making tasks by varying the embedding size (width) and the number of layers (depth) in \ourmod.
The aggregated results averaged over all tasks are show in \cref{fig:capacity_line}.
From the comparison, we can see that in general, increasing the model depth leads to a better performance. 
However, when embedding size gets too large, the performance further drops, as a large representation space might allow for irrelevant information.
Per-task comparison is provided in \cref{app:exp_capacity}.

\textbf{Other Potential Variants of \ours.} 
Following the basic idea of \ours, there could be many other variants and extensions on the design of both architecture and pretraining objective (e.g., does an additional contrastive loss help?). 
% For example, one may add more terms such as a contrastive loss to the pretraining objective.
% can \ours be boosted by adding a contrastive loss? Does it hurt or help to also predict observations in the \randinv loss? 
We investigate several variants of \ours and discuss the results in \cref{app:exp_ablation_variant}. 
In summary, adding more components to \ours could further improve its performance in certain scenarios, although such improvement may be limited and task-dependent. Exploring more extensions of \ours in various settings is an interesting future work.

\vspace{-2mm}
\section{Conclusion}
\label{sec:conclusion}
\vspace{-2mm}

This paper studies how to pretrain a versatile, generalizable and resilient representation model for multi-task sequential decision making. 
We propose a self-supervised and control-centric objective that encourages the transformer-based model to capture control-relevant representation.
Empirical results in multiple domains and tasks demonstrate the effectiveness of the proposed method, as well as its robustness to distribution shift and low-quality data. 
% Future work includes incorporating other techniques such as contrastive losses and data augmentation into our pretraining paradigm, as well as applying the method to more real-life applications such as robotics.
Future work includes strengthen the attention mechanism on both spatial observation space and temporal state-observation interactions, as well as investigating its potential generalization in a wider range of application scenarios. 

\newpage

\section{Ethics Statement}
\ours{} is a method that learns pretrained representations for controls tasks involving sequencial decision-making. 
As such, our models will reproduce patterns and biases found in its pretraining and finetuning datasets.

\section{Reproducibility Statement}
% We have submitted the code of this work as supplementary materials.
We are working towards providing an open-sourced implementation of our datasets, simulators, models and evaluation framework in the near future. We will make the links available for the camera-ready version of this work.

% \bibliography{reference}

\begin{thebibliography}{43}
\providecommand{\natexlab}[1]{#1}
\providecommand{\url}[1]{\texttt{#1}}
\expandafter\ifx\csname urlstyle\endcsname\relax
  \providecommand{\doi}[1]{doi: #1}\else
  \providecommand{\doi}{doi: \begingroup \urlstyle{rm}\Url}\fi

\bibitem[Andrychowicz et~al.(2017)Andrychowicz, Wolski, Ray, Schneider, Fong,
  Welinder, McGrew, Tobin, Pieter~Abbeel, and
  Zaremba]{andrychowicz2017hindsight}
Marcin Andrychowicz, Filip Wolski, Alex Ray, Jonas Schneider, Rachel Fong,
  Peter Welinder, Bob McGrew, Josh Tobin, OpenAI Pieter~Abbeel, and Wojciech
  Zaremba.
\newblock Hindsight experience replay.
\newblock \emph{Advances in neural information processing systems}, 30, 2017.

\bibitem[Banino et~al.(2022)Banino, Badia, Walker, Scholtes, Mitrovic, and
  Blundell]{banino2022coberl}
Andrea Banino, Adria~Puigdomenech Badia, Jacob~C Walker, Tim Scholtes, Jovana
  Mitrovic, and Charles Blundell.
\newblock Co{BERL}: Contrastive {BERT} for reinforcement learning.
\newblock In \emph{International Conference on Learning Representations}, 2022.
\newblock URL \url{https://openreview.net/forum?id=sRZ3GhmegS}.

\bibitem[Bonatti et~al.(2022)Bonatti, Vemprala, Ma, Frujeri, Chen, and
  Kapoor]{bonatti2022pact}
Rogerio Bonatti, Sai Vemprala, Shuang Ma, Felipe Frujeri, Shuhang Chen, and
  Ashish Kapoor.
\newblock Pact: Perception-action causal transformer for autoregressive
  robotics pre-training.
\newblock \emph{arXiv preprint arXiv:2209.11133}, 2022.

\bibitem[Brown et~al.(2020)Brown, Mann, Ryder, Subbiah, Kaplan, Dhariwal,
  Neelakantan, Shyam, Sastry, Askell, et~al.]{gpt:brown2020language}
Tom Brown, Benjamin Mann, Nick Ryder, Melanie Subbiah, Jared~D Kaplan, Prafulla
  Dhariwal, Arvind Neelakantan, Pranav Shyam, Girish Sastry, Amanda Askell,
  et~al.
\newblock Language models are few-shot learners.
\newblock \emph{Advances in neural information processing systems},
  33:\penalty0 1877--1901, 2020.

\bibitem[Chen et~al.(2022)Chen, Wu, Yoon, and Ahn]{chen2022transdreamer}
Chang Chen, Yi-Fu Wu, Jaesik Yoon, and Sungjin Ahn.
\newblock Transdreamer: Reinforcement learning with transformer world models.
\newblock \emph{arXiv preprint arXiv:2202.09481}, 2022.

\bibitem[Chen et~al.(2021)Chen, Lu, Rajeswaran, Lee, Grover, Laskin, Abbeel,
  Srinivas, and Mordatch]{chen2021decisiontransformer}
Lili Chen, Kevin Lu, Aravind Rajeswaran, Kimin Lee, Aditya Grover, Michael
  Laskin, Pieter Abbeel, Aravind Srinivas, and Igor Mordatch.
\newblock Decision transformer: Reinforcement learning via sequence modeling.
\newblock \emph{arXiv preprint arXiv:2106.01345}, 2021.

\bibitem[Devlin et~al.(2019)Devlin, Chang, Lee, and Toutanova]{devlin2019bert}
Jacob Devlin, Ming{-}Wei Chang, Kenton Lee, and Kristina Toutanova.
\newblock {BERT:} pre-training of deep bidirectional transformers for language
  understanding.
\newblock In Jill Burstein, Christy Doran, and Thamar Solorio (eds.),
  \emph{Proceedings of the 2019 Conference of the North American Chapter of the
  Association for Computational Linguistics: Human Language Technologies,
  {NAACL-HLT} 2019, Minneapolis, MN, USA, June 2-7, 2019, Volume 1 (Long and
  Short Papers)}, pp.\  4171--4186. Association for Computational Linguistics,
  2019.
\newblock \doi{10.18653/v1/n19-1423}.
\newblock URL \url{https://doi.org/10.18653/v1/n19-1423}.

\bibitem[Eysenbach et~al.(2019)Eysenbach, Gupta, Ibarz, and
  Levine]{eysenbach2018diversity}
Benjamin Eysenbach, Abhishek Gupta, Julian Ibarz, and Sergey Levine.
\newblock Diversity is all you need: Learning skills without a reward function.
\newblock In \emph{International Conference on Learning Representations}, 2019.
\newblock URL \url{https://openreview.net/forum?id=SJx63jRqFm}.

\bibitem[Furuta et~al.(2022)Furuta, Matsuo, and Gu]{furuta2022generalized}
Hiroki Furuta, Yutaka Matsuo, and Shixiang~Shane Gu.
\newblock Generalized decision transformer for offline hindsight information
  matching.
\newblock In \emph{International Conference on Learning Representations}, 2022.
\newblock URL \url{https://openreview.net/forum?id=CAjxVodl_v}.

\bibitem[Haarnoja et~al.(2018)Haarnoja, Zhou, Abbeel, and
  Levine]{haarnoja2018soft}
Tuomas Haarnoja, Aurick Zhou, Pieter Abbeel, and Sergey Levine.
\newblock Soft actor-critic: Off-policy maximum entropy deep reinforcement
  learning with a stochastic actor.
\newblock In \emph{International conference on machine learning}, pp.\
  1861--1870. PMLR, 2018.

\bibitem[He et~al.(2020)He, Fan, Wu, Xie, and Girshick]{he2020momentum}
Kaiming He, Haoqi Fan, Yuxin Wu, Saining Xie, and Ross Girshick.
\newblock Momentum contrast for unsupervised visual representation learning.
\newblock In \emph{Proceedings of the IEEE/CVF conference on computer vision
  and pattern recognition}, pp.\  9729--9738, 2020.

\bibitem[Hussein et~al.(2017)Hussein, Gaber, Elyan, and
  Jayne]{hussein2017imitation}
Ahmed Hussein, Mohamed~Medhat Gaber, Eyad Elyan, and Chrisina Jayne.
\newblock Imitation learning: A survey of learning methods.
\newblock \emph{ACM Comput. Surv.}, 50\penalty0 (2), apr 2017.
\newblock ISSN 0360-0300.
\newblock \doi{10.1145/3054912}.
\newblock URL \url{https://doi.org/10.1145/3054912}.

\bibitem[Janner et~al.(2021)Janner, Li, and Levine]{janner2021offline}
Michael Janner, Qiyang Li, and Sergey Levine.
\newblock Offline reinforcement learning as one big sequence modeling problem.
\newblock \emph{Advances in neural information processing systems}, 34, 2021.

\bibitem[Kadavath et~al.(2021)Kadavath, Paradis, and
  Yao]{kadavath2021pretraining}
Saurav Kadavath, Samuel Paradis, and Brian Yao.
\newblock Pretraining \& reinforcement learning: Sharpening the axe before
  cutting the tree.
\newblock \emph{arXiv preprint arXiv:2110.02497}, 2021.

\bibitem[Kaplan et~al.(2020)Kaplan, McCandlish, Henighan, Brown, Chess, Child,
  Gray, Radford, Wu, and Amodei]{kaplan2020scaling}
Jared Kaplan, Sam McCandlish, Tom Henighan, Tom~B Brown, Benjamin Chess, Rewon
  Child, Scott Gray, Alec Radford, Jeffrey Wu, and Dario Amodei.
\newblock Scaling laws for neural language models.
\newblock \emph{arXiv preprint arXiv:2001.08361}, 2020.

\bibitem[Lamb et~al.(2022)Lamb, Islam, Efroni, Didolkar, Misra, Foster, Molu,
  Chari, Krishnamurthy, and Langford]{lamb2022guaranteed}
Alex Lamb, Riashat Islam, Yonathan Efroni, Aniket Didolkar, Dipendra Misra,
  Dylan Foster, Lekan Molu, Rajan Chari, Akshay Krishnamurthy, and John
  Langford.
\newblock Guaranteed discovery of controllable latent states with multi-step
  inverse models.
\newblock \emph{arXiv preprint arXiv:2207.08229}, 2022.

\bibitem[Lee et~al.(2022)Lee, Nachum, Yang, Lee, Freeman, Xu, Guadarrama,
  Fischer, Jang, Michalewski, et~al.]{lee2022multi}
Kuang-Huei Lee, Ofir Nachum, Mengjiao Yang, Lisa Lee, Daniel Freeman, Winnie
  Xu, Sergio Guadarrama, Ian Fischer, Eric Jang, Henryk Michalewski, et~al.
\newblock Multi-game decision transformers.
\newblock \emph{arXiv preprint arXiv:2205.15241}, 2022.

\bibitem[Lee et~al.(2021)Lee, Seo, Lee, Abbeel, and
  Shin]{lee2021offlinetoonline}
Seunghyun Lee, Younggyo Seo, Kimin Lee, Pieter Abbeel, and Jinwoo Shin.
\newblock Offline-to-online reinforcement learning via balanced replay and
  pessimistic q-ensemble.
\newblock In \emph{5th Annual Conference on Robot Learning}, 2021.
\newblock URL \url{https://openreview.net/forum?id=AlJXhEI6J5W}.

\bibitem[Liu \& Abbeel(2021)Liu and Abbeel]{liu2021behavior}
Hao Liu and Pieter Abbeel.
\newblock Behavior from the void: Unsupervised active pre-training.
\newblock \emph{Advances in Neural Information Processing Systems},
  34:\penalty0 18459--18473, 2021.

\bibitem[Loshchilov \& Hutter(2019)Loshchilov and
  Hutter]{loshchilov2018decoupled}
Ilya Loshchilov and Frank Hutter.
\newblock Decoupled weight decay regularization.
\newblock In \emph{International Conference on Learning Representations}, 2019.
\newblock URL \url{https://openreview.net/forum?id=Bkg6RiCqY7}.

\bibitem[Mendonca et~al.(2021)Mendonca, Rybkin, Daniilidis, Hafner, and
  Pathak]{mendonca2021discovering}
Russell Mendonca, Oleh Rybkin, Kostas Daniilidis, Danijar Hafner, and Deepak
  Pathak.
\newblock Discovering and achieving goals via world models.
\newblock \emph{Advances in Neural Information Processing Systems}, 34, 2021.

\bibitem[Micheli et~al.(2022)Micheli, Alonso, and
  Fleuret]{micheli2022transformers}
Vincent Micheli, Eloi Alonso, and Fran{\c{c}}ois Fleuret.
\newblock Transformers are sample efficient world models.
\newblock \emph{arXiv preprint arXiv:2209.00588}, 2022.

\bibitem[Mnih et~al.(2015)Mnih, Kavukcuoglu, Silver, Rusu, Veness, Bellemare,
  Graves, Riedmiller, Fidjeland, Ostrovski, et~al.]{mnih2015human}
Volodymyr Mnih, Koray Kavukcuoglu, David Silver, Andrei~A Rusu, Joel Veness,
  Marc~G Bellemare, Alex Graves, Martin Riedmiller, Andreas~K Fidjeland, Georg
  Ostrovski, et~al.
\newblock Human-level control through deep reinforcement learning.
\newblock \emph{nature}, 518\penalty0 (7540):\penalty0 529--533, 2015.

\bibitem[Nair et~al.(2022)Nair, Rajeswaran, Kumar, Finn, and
  Gupta]{nair2022r3m}
Suraj Nair, Aravind Rajeswaran, Vikash Kumar, Chelsea Finn, and Abhinav Gupta.
\newblock R3m: A universal visual representation for robot manipulation.
\newblock \emph{arXiv preprint arXiv:2203.12601}, 2022.

\bibitem[Oord et~al.(2018)Oord, Li, and Vinyals]{CPC:oord2018representation}
Aaron van~den Oord, Yazhe Li, and Oriol Vinyals.
\newblock Representation learning with contrastive predictive coding.
\newblock \emph{arXiv preprint arXiv:1807.03748}, 2018.

\bibitem[Osa et~al.(2018)Osa, Pajarinen, Neumann, Bagnell, Abbeel, Peters,
  et~al.]{osa2018algorithmic}
Takayuki Osa, Joni Pajarinen, Gerhard Neumann, J~Andrew Bagnell, Pieter Abbeel,
  Jan Peters, et~al.
\newblock An algorithmic perspective on imitation learning.
\newblock \emph{Foundations and Trends{\textregistered} in Robotics},
  7\penalty0 (1-2):\penalty0 1--179, 2018.

\bibitem[Parisi et~al.(2022)Parisi, Rajeswaran, Purushwalkam, and
  Gupta]{parisi2022unsurprising}
Simone Parisi, Aravind Rajeswaran, Senthil Purushwalkam, and Abhinav Gupta.
\newblock The unsurprising effectiveness of pre-trained vision models for
  control.
\newblock \emph{arXiv preprint arXiv:2203.03580}, 2022.

\bibitem[Pathak et~al.(2017)Pathak, Agrawal, Efros, and
  Darrell]{pathak2017curiosity}
Deepak Pathak, Pulkit Agrawal, Alexei~A Efros, and Trevor Darrell.
\newblock Curiosity-driven exploration by self-supervised prediction.
\newblock In \emph{International conference on machine learning}, pp.\
  2778--2787. PMLR, 2017.

\bibitem[Radford et~al.(2018)Radford, Narasimhan, Salimans, Sutskever,
  et~al.]{radford2018improving}
Alec Radford, Karthik Narasimhan, Tim Salimans, Ilya Sutskever, et~al.
\newblock Improving language understanding by generative pre-training.
\newblock 2018.

\bibitem[Radford et~al.(2021)Radford, Kim, Hallacy, Ramesh, Goh, Agarwal,
  Sastry, Askell, Mishkin, Clark, et~al.]{radford2021learning}
Alec Radford, Jong~Wook Kim, Chris Hallacy, Aditya Ramesh, Gabriel Goh,
  Sandhini Agarwal, Girish Sastry, Amanda Askell, Pamela Mishkin, Jack Clark,
  et~al.
\newblock Learning transferable visual models from natural language
  supervision.
\newblock In \emph{International Conference on Machine Learning}, pp.\
  8748--8763. PMLR, 2021.

\bibitem[Radosavovic et~al.()Radosavovic, Xiao, James, Abbeel, Malik, and
  Darrell]{radosavovicreal}
Ilija Radosavovic, Tete Xiao, Stephen James, Pieter Abbeel, Jitendra Malik, and
  Trevor Darrell.
\newblock Real world robot learning with masked visual pre-training.
\newblock In \emph{6th Annual Conference on Robot Learning}.

\bibitem[Rakelly et~al.(2021)Rakelly, Gupta, Florensa, and
  Levine]{rakelly2021which}
Kate Rakelly, Abhishek Gupta, Carlos Florensa, and Sergey Levine.
\newblock Which mutual-information representation learning objectives are
  sufficient for control?
\newblock In M.~Ranzato, A.~Beygelzimer, Y.~Dauphin, P.S. Liang, and J.~Wortman
  Vaughan (eds.), \emph{Advances in Neural Information Processing Systems},
  volume~34, pp.\  26345--26357. Curran Associates, Inc., 2021.
\newblock URL
  \url{https://proceedings.neurips.cc/paper/2021/file/dd45045f8c68db9f54e70c67048d32e8-Paper.pdf}.

\bibitem[Reed et~al.(2022)Reed, Zolna, Parisotto, Colmenarejo, Novikov,
  Barth-Maron, Gimenez, Sulsky, Kay, Springenberg, et~al.]{reed2022generalist}
Scott Reed, Konrad Zolna, Emilio Parisotto, Sergio~Gomez Colmenarejo, Alexander
  Novikov, Gabriel Barth-Maron, Mai Gimenez, Yury Sulsky, Jackie Kay,
  Jost~Tobias Springenberg, et~al.
\newblock A generalist agent.
\newblock \emph{arXiv preprint arXiv:2205.06175}, 2022.

\bibitem[Schwarzer et~al.(2021)Schwarzer, Rajkumar, Noukhovitch, Anand,
  Charlin, Hjelm, Bachman, and Courville]{schwarzer2021pretraining}
Max Schwarzer, Nitarshan Rajkumar, Michael Noukhovitch, Ankesh Anand, Laurent
  Charlin, R~Devon Hjelm, Philip Bachman, and Aaron~C Courville.
\newblock Pretraining representations for data-efficient reinforcement
  learning.
\newblock In M.~Ranzato, A.~Beygelzimer, Y.~Dauphin, P.S. Liang, and J.~Wortman
  Vaughan (eds.), \emph{Advances in Neural Information Processing Systems},
  volume~34, pp.\  12686--12699. Curran Associates, Inc., 2021.
\newblock URL
  \url{https://proceedings.neurips.cc/paper/2021/file/69eba34671b3ef1ef38ee85caae6b2a1-Paper.pdf}.

\bibitem[Seo et~al.(2022)Seo, Lee, James, and Abbeel]{seo2022reinforcement}
Younggyo Seo, Kimin Lee, Stephen James, and Pieter Abbeel.
\newblock Reinforcement learning with action-free pre-training from videos.
\newblock \emph{arXiv preprint arXiv:2203.13880}, 2022.

\bibitem[Shah \& Kumar(2021)Shah and Kumar]{shah2021rrl}
Rutav Shah and Vikash Kumar.
\newblock Rrl: Resnet as representation for reinforcement learning.
\newblock In \emph{International Conference on Machine Learning}. PMLR, 2021.

\bibitem[Stooke et~al.(2021)Stooke, Lee, Abbeel, and
  Laskin]{stooke2021decoupling}
Adam Stooke, Kimin Lee, Pieter Abbeel, and Michael Laskin.
\newblock Decoupling representation learning from reinforcement learning.
\newblock In \emph{International Conference on Machine Learning}, pp.\
  9870--9879. PMLR, 2021.

\bibitem[Sutton \& Barto(2018)Sutton and Barto]{sutton2018reinforcement}
Richard~S Sutton and Andrew~G Barto.
\newblock \emph{Reinforcement learning: An introduction}.
\newblock MIT press, 2018.

\bibitem[Tassa et~al.(2018)Tassa, Doron, Muldal, Erez, Li, Casas, Budden,
  Abdolmaleki, Merel, Lefrancq, et~al.]{tassa2018deepmind}
Yuval Tassa, Yotam Doron, Alistair Muldal, Tom Erez, Yazhe Li, Diego de~Las
  Casas, David Budden, Abbas Abdolmaleki, Josh Merel, Andrew Lefrancq, et~al.
\newblock Deepmind control suite.
\newblock \emph{arXiv preprint arXiv:1801.00690}, 2018.

\bibitem[Tunyasuvunakool et~al.(2020)Tunyasuvunakool, Muldal, Doron, Liu,
  Bohez, Merel, Erez, Lillicrap, Heess, and Tassa]{tunyasuvunakool2020}
Saran Tunyasuvunakool, Alistair Muldal, Yotam Doron, Siqi Liu, Steven Bohez,
  Josh Merel, Tom Erez, Timothy Lillicrap, Nicolas Heess, and Yuval Tassa.
\newblock dm\_control: Software and tasks for continuous control.
\newblock \emph{Software Impacts}, 6:\penalty0 100022, 2020.
\newblock ISSN 2665-9638.
\newblock \doi{https://doi.org/10.1016/j.simpa.2020.100022}.
\newblock URL
  \url{https://www.sciencedirect.com/science/article/pii/S2665963820300099}.

\bibitem[Vaswani et~al.(2017)Vaswani, Shazeer, Parmar, Uszkoreit, Jones, Gomez,
  Kaiser, and Polosukhin]{vaswani2017attention}
Ashish Vaswani, Noam Shazeer, Niki Parmar, Jakob Uszkoreit, Llion Jones,
  Aidan~N Gomez, {\L}ukasz Kaiser, and Illia Polosukhin.
\newblock Attention is all you need.
\newblock \emph{Advances in neural information processing systems}, 30, 2017.

\bibitem[Yang \& Nachum(2021)Yang and Nachum]{yang2021representation}
Mengjiao Yang and Ofir Nachum.
\newblock Representation matters: offline pretraining for sequential decision
  making.
\newblock In \emph{International Conference on Machine Learning}, pp.\
  11784--11794. PMLR, 2021.

\bibitem[Zheng et~al.(2022)Zheng, Zhang, and Grover]{zheng2022online}
Qinqing Zheng, Amy Zhang, and Aditya Grover.
\newblock Online decision transformer.
\newblock In Kamalika Chaudhuri, Stefanie Jegelka, Le~Song, Csaba Szepesvari,
  Gang Niu, and Sivan Sabato (eds.), \emph{Proceedings of the 39th
  International Conference on Machine Learning}, volume 162 of
  \emph{Proceedings of Machine Learning Research}, pp.\  27042--27059. PMLR,
  17--23 Jul 2022.
\newblock URL \url{https://proceedings.mlr.press/v162/zheng22c.html}.

\end{thebibliography}

\bibliographystyle{iclr2023_conference}

\newpage
\appendix
{\centering{\Large Appendix}}

\section{Implementation Details}
\label{app:implementation}

\subsection{Implementation of Pretraining Objectives}
\label{app:imple_obj}

\textbf{I. Forward Dynamics Prediction.} \\
For each observation-action pair $(o_t, a_t)$ in a control sequence, the forward prediction loss is constructed as follows. 
Let $g_\theta$ be the observation tokenizer being trained where $\theta$ denotes the parameterization. 
We maintain a momentum encoder $\bar{g}_{\bar{\theta}}$ whose parameters are updated by $\bar{\theta} = \tau (\bar{\theta}) + (1-\tau) \theta $. With the next observation $o_{t+1}$, we have $\hat{s}_{t+1}:=\mathsf{SG}(\bar{g}_{\bar{\theta}}(o_{t+1}))$, where $\mathsf{SG}$ refers to stop gradient. 
Then the forward prediction loss is defined as: 
\begin{equation}
    \label{loss:forward}
    L_{\fwdmath} := \mathrm{MSE}\left( f_{\fwdmath}(\rep(o_t),\rep(a_t)), \hat{s}_{t+1} \right),
\end{equation}
where $f_{\fwdmath}$ is a linear prediction head.

\textbf{II. Inverse Dynamics Prediction.} \\
For each consecutive observation pair $(o_t, o_{t+1})$, inverse prediction tries to recover the action that leads $o_t$ to $o_{t+1}$, which gives the loss function
\begin{equation}
    \label{loss:inverse}
    L_{\invmath} := \mathrm{MSE}\left( f_{\invmath}(\rep(o_t),\rep(o_{t+1})), a_{t} \right),
\end{equation}
where $f_{\invmath}$ is a linear prediction head.

\textbf{III. \randinvcap.} \\
In our implementation, we use a predefined mask-token $\mathbf{m}=[-1,\cdots,-1]$ to replace the original tokens. 
Let $\tilde{h}$ denote the masked control sequence, and $0 \leq s_1,s_2,\cdots,s_{k} \leq L$ be the selected indices for masked actions.  Then the loss function can be defined as:
\begin{equation}
\label{loss:inverse}
\begin{aligned}
    &L_{\randinvmath} := \sum\nolimits_{i=1}^{k} l(s_i), \\
    & \text{where} \quad l(s_i) = 
    \begin{cases}
        \mathrm{MSE}\left( f_{\randinvmath}(\rep_M(\mathbf{m}_{t+s_i});\tilde{h}), a_{t+s_i} \right), & \text{if } s_i < L\\
        0 & \text{if } s_i = L
    \end{cases}
\end{aligned}
\end{equation}
where $f_{\randinvmath}$ is a linear prediction head, and $\rep_M$ is the transformer model without a causal mask. Note that we do not predict $a_{t+L}$ as it is infeasible to recover it without future observations.

\textit{Schedule of Masking Size.}
Theoretically, it is possible to recover a full action sequence for a given observation sequence, which implies that $k=L$ is a reasonable setup. But in environments with complex dynamics, directly recovering all actions is hard in the start of training. Hence, we adjust the difficulty of \randinv in a curriculum way, by gradually increasing the value of $k$ in the following schedule:
\begin{equation}
    k = \max\left(1, \mathrm{int}\left(L * \frac{\mathrm{current\_epoch} + 1}{\mathrm{total\_epochs}}\right)\right)
\end{equation}
On the other hand, if we predict actions from all observations, it is possible that the model mainly relies on local dynamics, i.e., predict $a_t$ mainly based on $o_{t-1}, o_{t-2}, o_{t-3}$, conflicting with our desire of learning long-term dependence. Therefore, we also mask a subset of observations with $k^\prime$. In the extreme case $k^\prime=L$, the objective becomes similar to goal-conditioned modeling with $o_t$ being the start and $o_{t+L}$ is the goal. However, in a control environment, there usually exist multiple paths from $o_t$ to $o_{t+L}$, making the action prediction ambiguous. \citet{schwarzer2021pretraining} uses a goal-conditioned objective which finds the shortest path. However, extra value fitting and planning are required, which may lead to higher cost and compounding errors. 
Therefore, we intentionally make $k^\prime$ smaller than the context length (half of $L$), such that the model is able to predict the masked actions based on revealed subsequence of observations and actions. The schedule of $k^\prime$ is as below.
\begin{equation}
    k^\prime = \max\left(1, \mathrm{int}\left(\frac{L}{2} * \frac{\mathrm{current\_epoch} + 1}{\mathrm{total\_epochs}}\right)\right).
\end{equation}

The overall pretraining objective is
\begin{equation}
    \label{loss:all}
    \min_{\rep, f_{\fwdmath}, f_{\invmath}, f_{\randinvmath}} L_{\fwdmath} + L_{\invmath} + L_{\randinvmath}.
\end{equation} 

\subsection{Environment and Dataset.}
\label{app:imple_env}

\begin{figure}[h]
    \centering
    \includegraphics[width=0.5\textwidth]{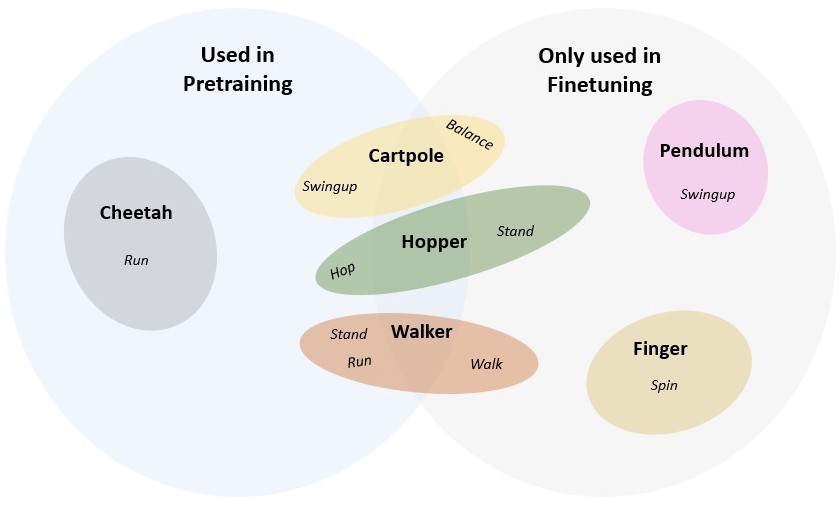}
    \caption{Graphical relations of all tasks involved.}
    \label{fig:tasks}
\end{figure}

\cref{tab:envs} lists all domains and tasks from DMC used in our experiments, and their relations are further depicted in \cref{fig:tasks}. 

\begin{table}[!htbp]
\centering
\begin{tabular}{|c|c|c|c|}
\toprule  
\textbf{Phase} & \textbf{Domain} & \textbf{Task} & \textbf{Expert Score by SAC} \\ \hline 
\multirow{5}{*}{Pretraining \& Finetuning} & cartpole & swingup & 875 \\ 
 & hopper & hop & 200 \\ 
 & cheetah & run & 850 \\ 
 & walker & stand & 980 \\ 
 & walker & run & 700 \\ \hline
\multirow{5}{*}{Finetuning only} & cartpole & balance & 1000 \\ 
& hopper & stand & 900 \\ 
& walker & walk & 950 \\ 
& pendulum & swingup & 1000\\ 
& finger & spin & 800 \\ 
\bottomrule
\end{tabular}
% \vspace{-1em}
\caption{A list of domains and tasks used in pretraining and finetuning. The first 5 tasks are used for pretraining. In the finetuning phase, we use the pretrained model to learn policies in all 10 tasks, including the last 5 tasks that are unseen during pretraining.}
\label{tab:envs} 
\end{table}

To generate our datasets for pretraining and finetuning, we first train 5 agents (corresponding to different random seeds) for each task with ground truth vector states using SAC~\citep{haarnoja2018soft} for 1M steps, and collect the full replay buffer with corresponding RGB images (3$\times$84$\times$84) rendered by the physical simulator.
% We use the open-source implementation by \citet{yarats2019improving} (the SAC algorithm with groundtruth observation).
% to learn expert policies and collect trajectories along-the-way. For each step of learning, we save the rendered RGB image, the action taken by the agent, the reward (only used in RTG finetuning and for the DT baseline) and the done flags. We also use uniformly random agents to produce the Random dataset. 
Then, we divide the replay buffers and create the following datasets of different qualities.
\vspace{-3mm}
\begin{itemize}[noitemsep,leftmargin=*]
    \setlist{nolistsep}
    \item \texttt{Random}: Randomly generated interaction trajectories. This dataset has 400K timesteps per task.
    \item \texttt{Exploratory}: The first 80K timesteps of each SAC learner, corresponding to the exploratory stage. For all 5 agents, this leads to a cumulative of 400K timesteps per task. 
\end{itemize}
\vspace{-3mm}
Note that different tasks have different difficulties for the SAC agent to converge. For example, in cartpole, the agent converges with much less samples than in walker. Therefore, we slightly adjust the proportion of data from different tasks in multi-task pretraining to avoid overfitting to simple tasks and underfitting to harder tasks. 
In pretraining, we use 40K timesteps from each cartpole behavior agent (200K by 5 agents), and 90K timesteps from each walker behavior agent (450K by 5 agents), and 80K for all other agents (400K by 5 agents).

Datasets for downstream learning:
\vspace{-3mm}
\begin{itemize}[noitemsep,leftmargin=*]
    \setlist{nolistsep}
    \item \texttt{Sampled Replay} (for RTG): We randomly sample 10\% trajectories from the full replay buffer of 1 SAC agent, resulting in a dataset of size 100K per task, with diverse return distribution.
    \item \texttt{Expert} (for BC): We select 10\% trajectories with the highest returns from the full replay buffer of 1 SAC agent, resulting in an expert dataset of size 100K per task.
\end{itemize}
\vspace{-2mm}

% Our preliminary experiments suggest that the diversity of pretraining tasks and domains is more important than the total amount of data used. 

\subsection{Model and Hyperparameters.}
\label{app:imple_model}
Following the implementation of Decision Transformer~\citep{chen2021decisiontransformer}, our transformer backbone is based on the minGPT implementation \href{https://github.com/karpathy/minGPT}{https://github.com/karpathy/minGPT} with the default AdamW optimizer~\citep{loshchilov2018decoupled}. 
For both pretraining and finetuning, the learning rate is set to be $6 \times 10^{-4}$ and batch size 256. For learning rate, linear warmup and cosine decay are used. A context length 30 is used in all tasks for both training and execution. We tested different context lengths in preliminary experiments, and a shorter context length (5/10/20) does not work as well as 30 when training from scratch.
We use 8 attention heads, 8 layer attention blocks, and an embedding size 256 in all experiments, for both our model and baselines. There are 10.8 M trainable parameters in the model.  
We test \ours with varying layer numbers and embedding sizes in \cref{app:exp_capacity}.

For the execution of learned RTG-conditioned policies, we set the expected RTG as the expert scores in \cref{tab:envs}. Tuning the RTG setting may further increase the results. But since our focus is to show the effectiveness of pretraining, we did not explore other possibilities. 

All models are trained for 10 epochs in pretraining, and 20 epochs for each downstream task. The performance of the best checkpoint is reported, as detailed in the next section.

\subsection{Evaluation Metrics.}
\label{app:imple_metric}

For both BC and RTG downstream learning, we report the average cumulative reward of the learned policy by interacting with the environment for 50 episodes. In \cref{fig:data_baseline}, we calculate the \textit{normalized reward} based on expert scores. 

In \cref{fig:ablation} and \cref{fig:reward_ablation}, we report the relative improvement of each ablated method calculated by the following formula.
\begin{equation}
    \mathrm{relative\_improvement} := 
    \frac{\mathrm{method\_reward}-\mathrm{scratch\_reward}}{\mathrm{scratch\_reward}},
\end{equation}
where the scratch\_reward is the best reward of training from scratch using the same learning configurations.

\section{Additional Experiment Results}
\label{app:exp}

\subsection{More Results of Downstream Learning}
\label{app:exp_full}

\begin{figure}[!htbp]
% \vspace{-1em}
 \centering

\rotatebox{90}{\scriptsize{\hspace{1cm}\textbf{RTG}}}
 \begin{subfigure}[t]{0.19\columnwidth}
  \resizebox{1\textwidth}{!}{\input{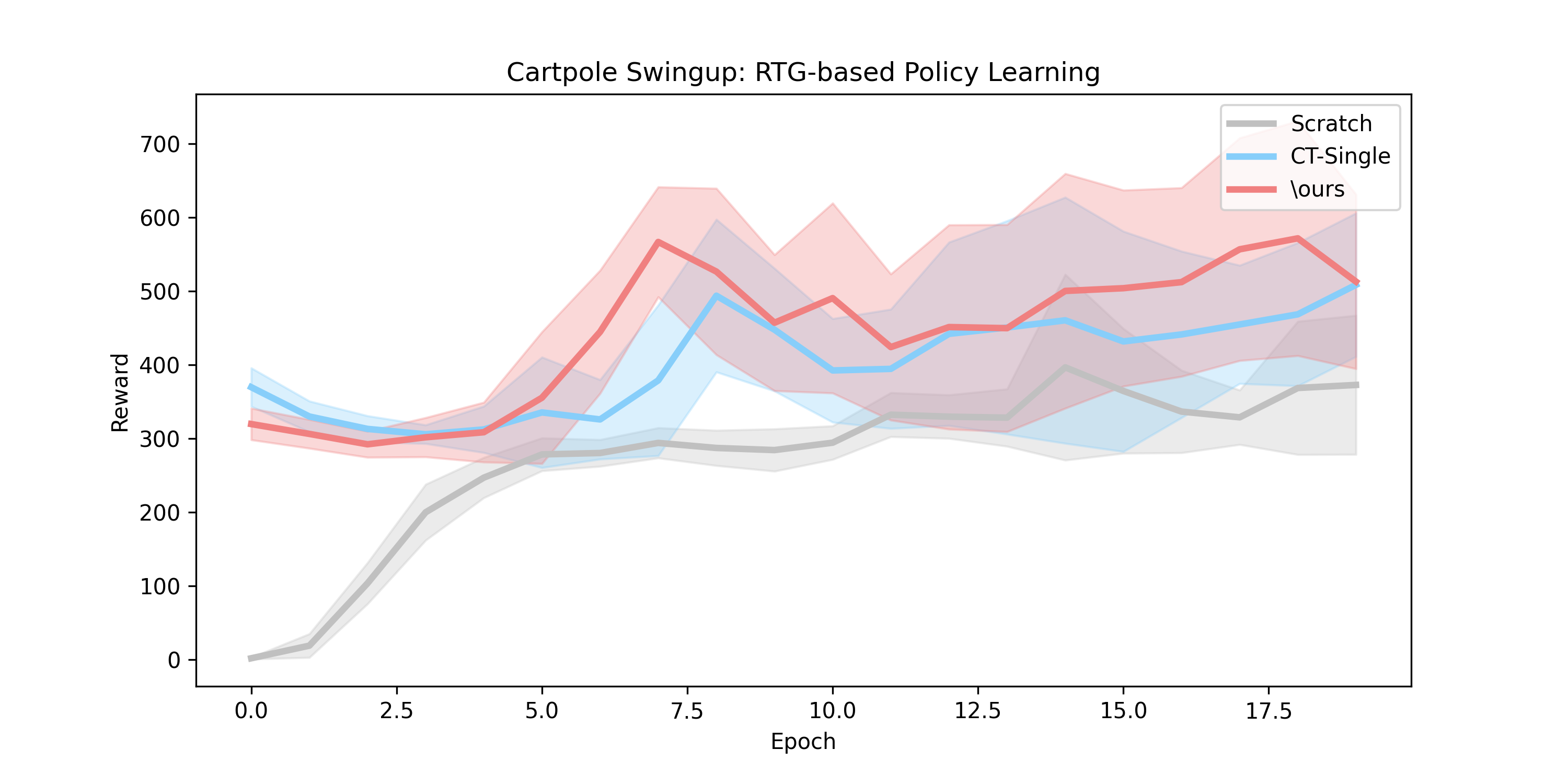}}
%   \vspace{-1.5em}
%   \caption{FoodCollector}
 \end{subfigure}
 \hfill
 \begin{subfigure}[t]{0.19\columnwidth}
  \resizebox{\textwidth}{!}{\input{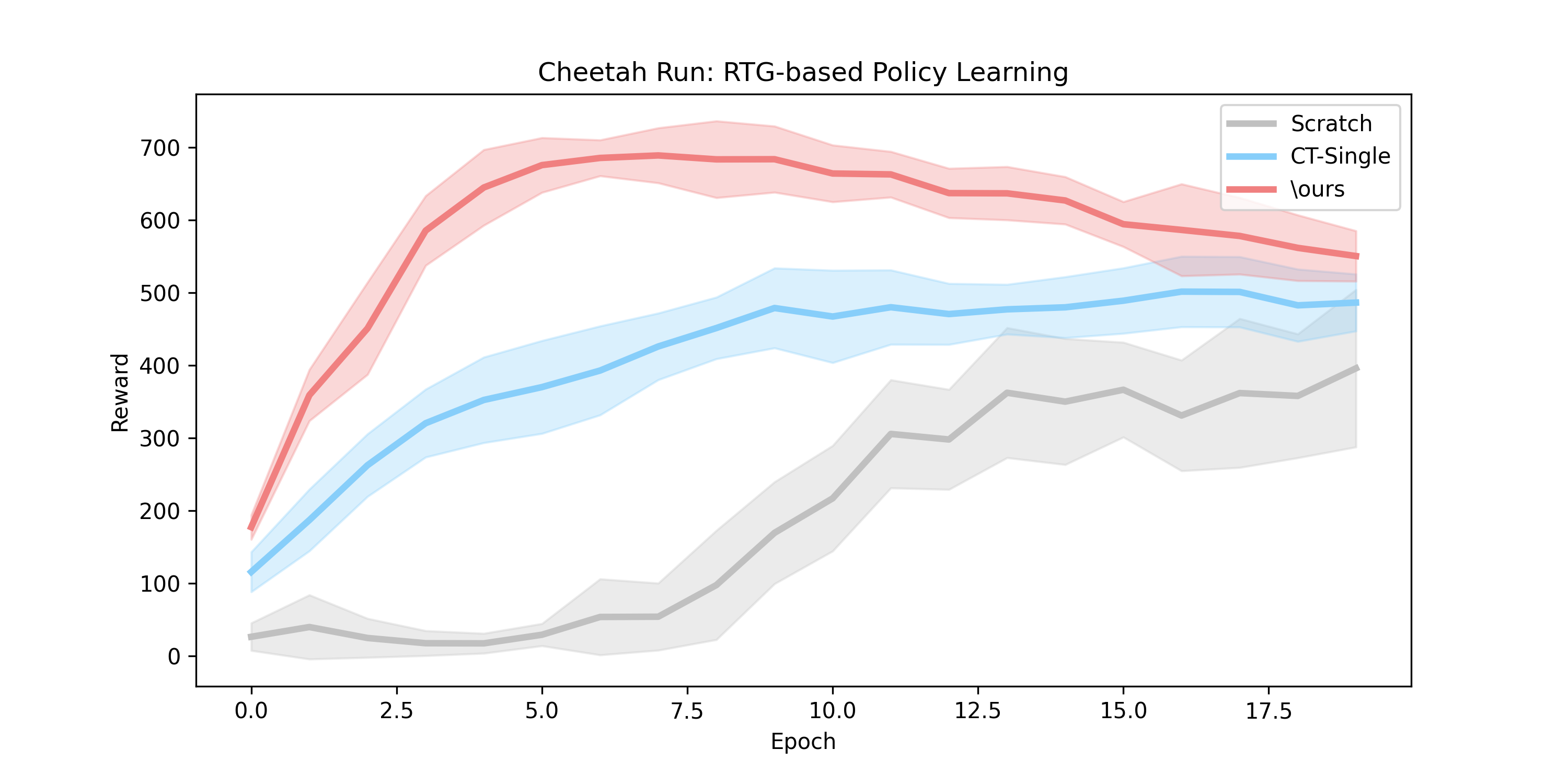}}
%   \vspace{-1.5em}
%   \caption{Disc. \& Adaptive}
 \end{subfigure}
 \hfill
 \begin{subfigure}[t]{0.19\columnwidth}
  \resizebox{\textwidth}{!}{\input{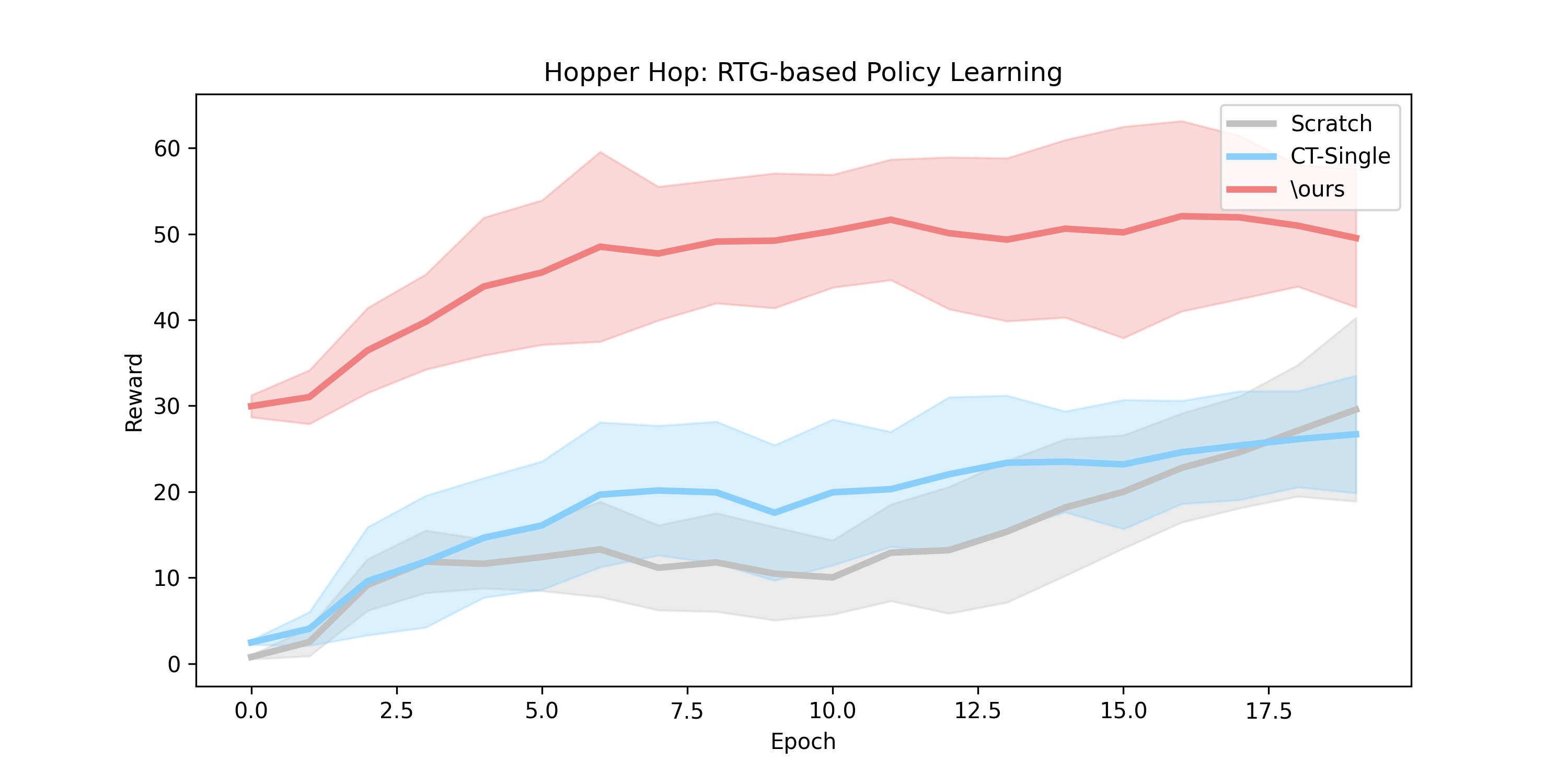}}
%   \vspace{-1.5em}
%   \caption{Cont. \& Non-adaptive}
 \end{subfigure}
 \hfill
 \begin{subfigure}[t]{0.19\columnwidth}
  \resizebox{\textwidth}{!}{\input{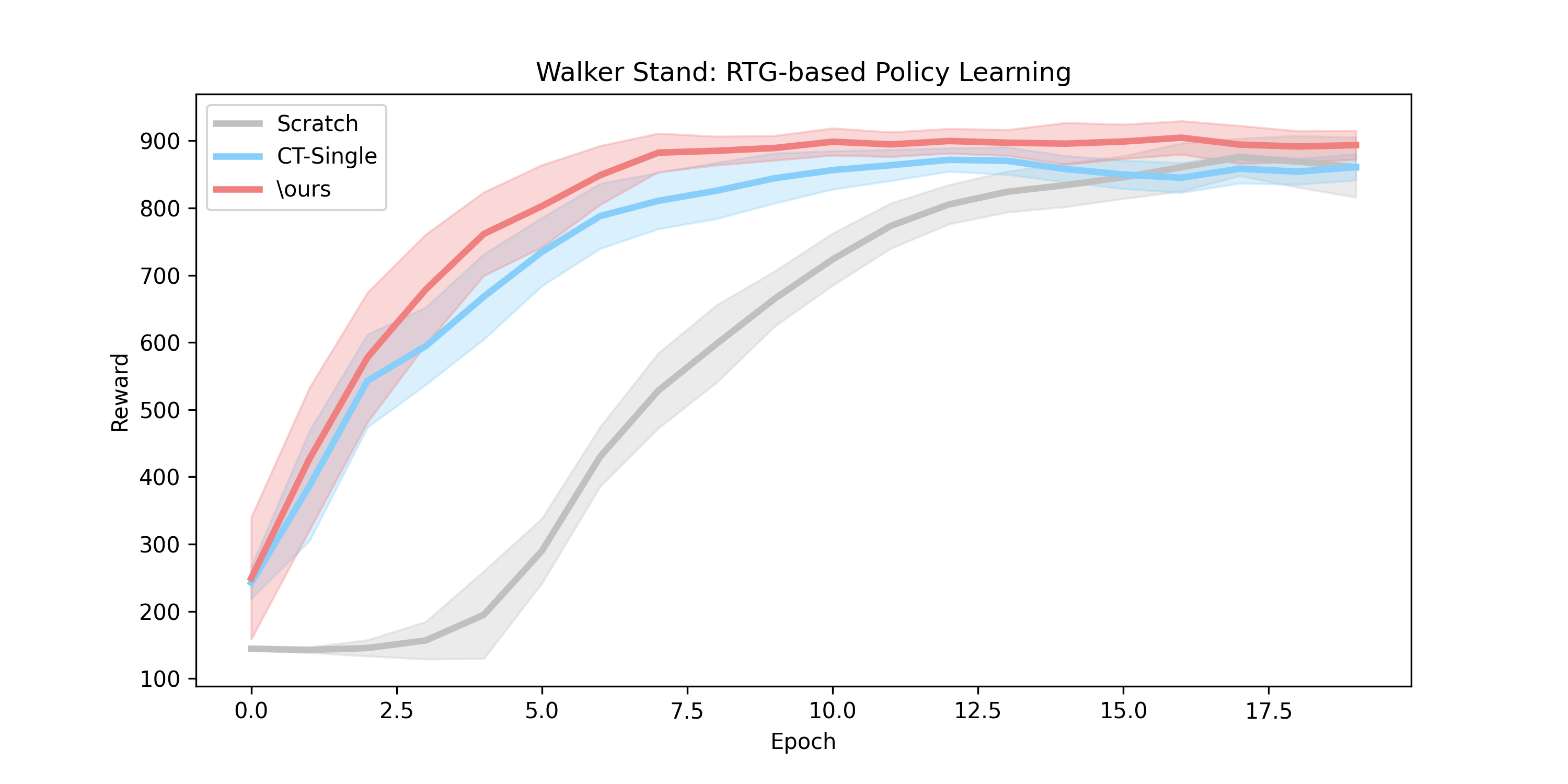}}
%   \vspace{-1.5em}
%   \caption{Cont. \& Adaptive}
 \end{subfigure} 
 \hfill
 \begin{subfigure}[t]{0.19\columnwidth}
  \resizebox{\textwidth}{!}{\input{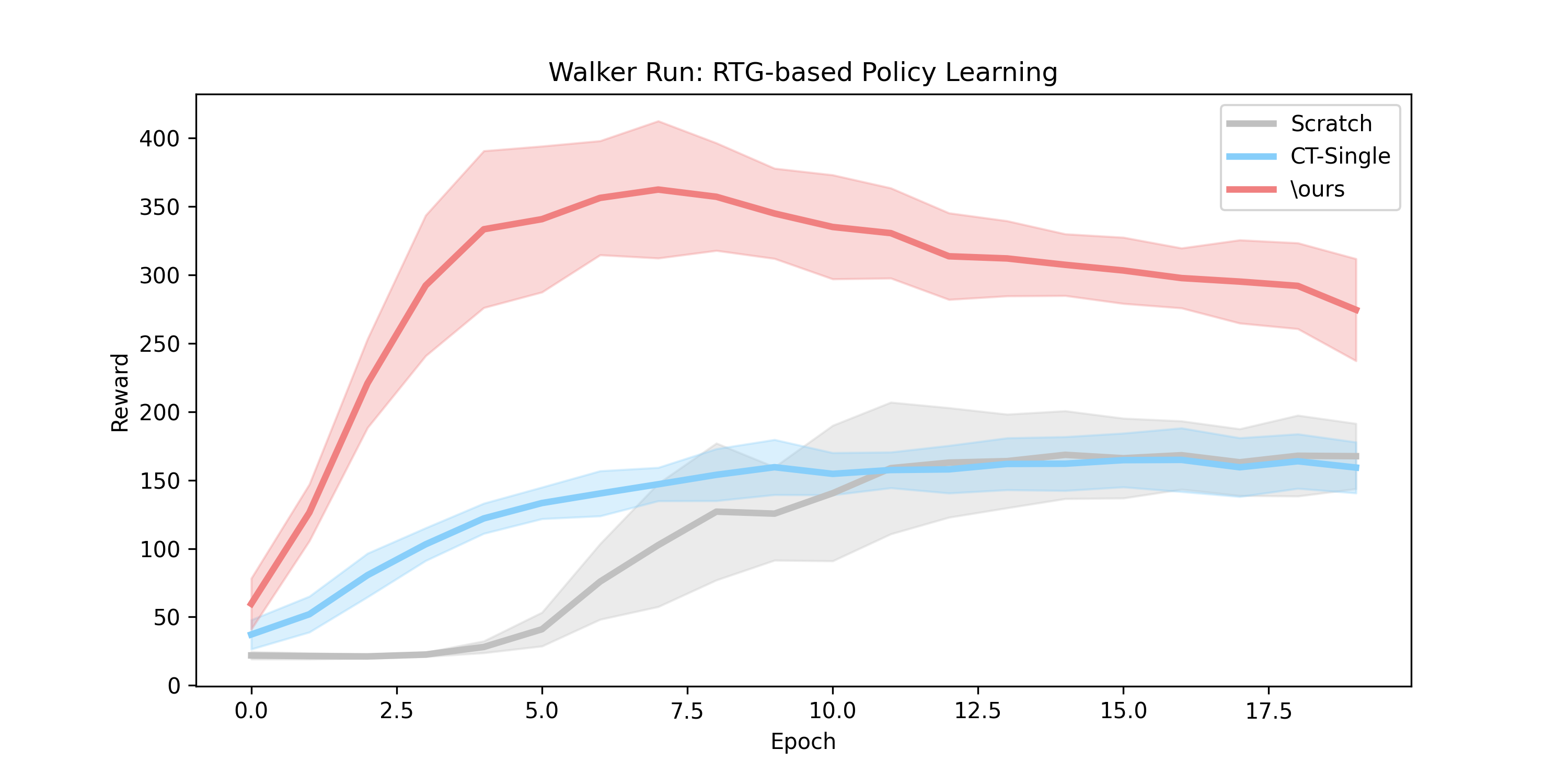}}
%   \vspace{-1.5em}
%   \caption{Cont. \& Adaptive}
 \end{subfigure} 
%  \vspace{0.5em}

\rotatebox{90}{\scriptsize{\hspace{1cm}\textbf{BC}}}
 \begin{subfigure}[t]{0.19\columnwidth}
  \resizebox{1.04\textwidth}{!}{\input{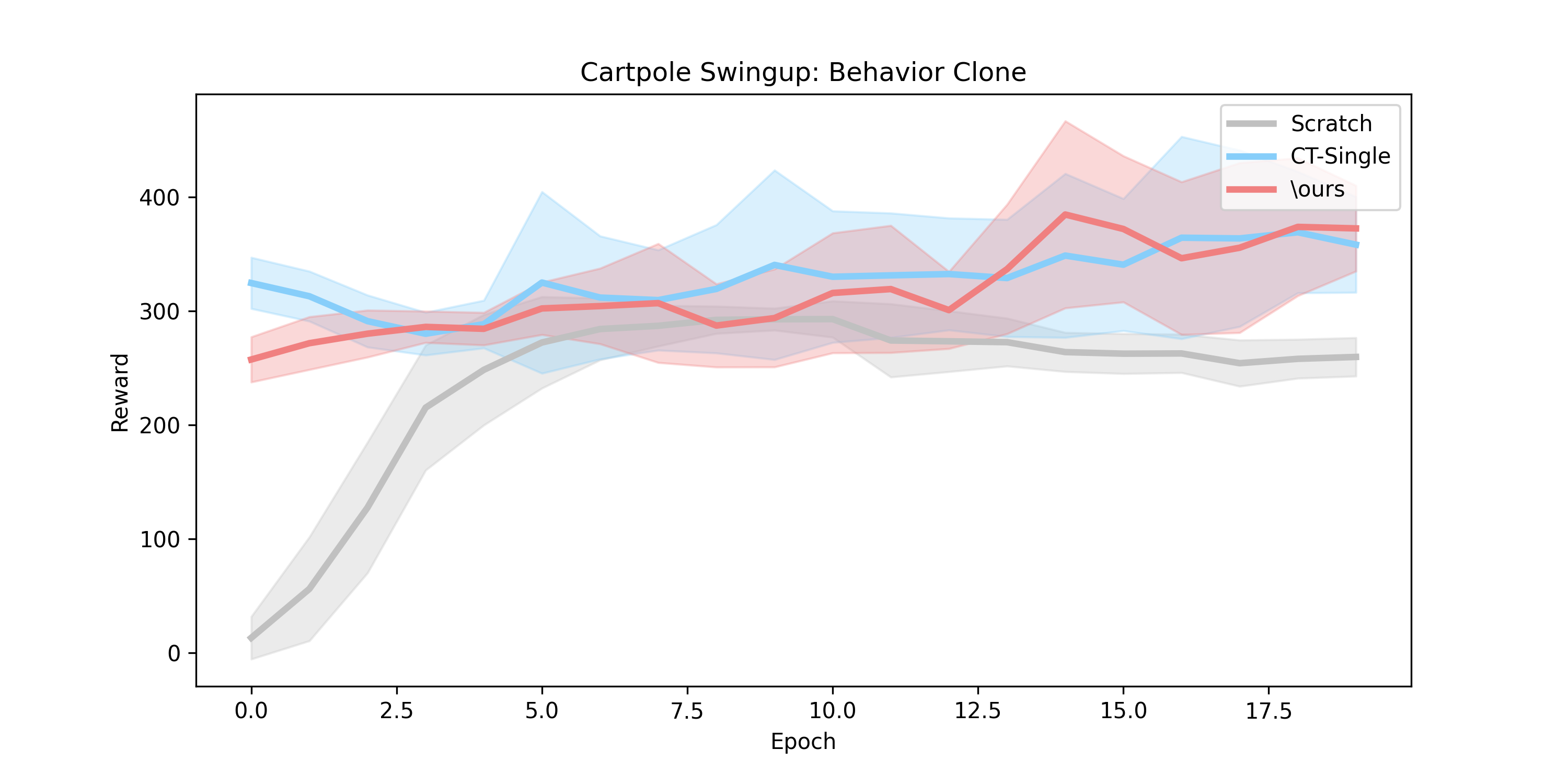}}
%   \vspace{-1.5em}
%   \caption{FoodCollector}
 \end{subfigure}
 \hfill
 \begin{subfigure}[t]{0.19\columnwidth}
  \resizebox{\textwidth}{!}{\input{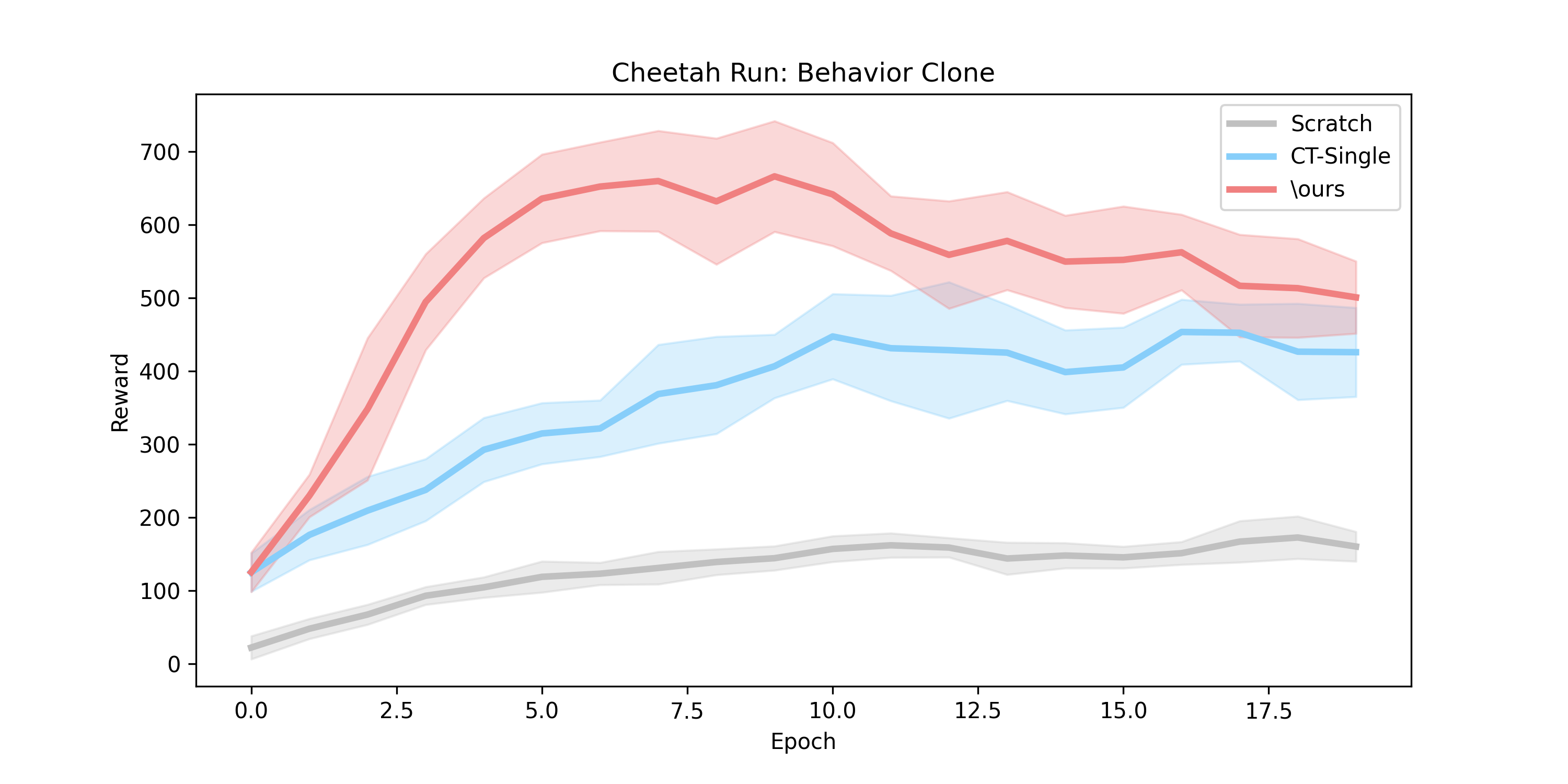}}
%   \vspace{-1.5em}
%   \caption{Disc. \& Adaptive}
 \end{subfigure}
 \hfill
 \begin{subfigure}[t]{0.19\columnwidth}
  \resizebox{\textwidth}{!}{\input{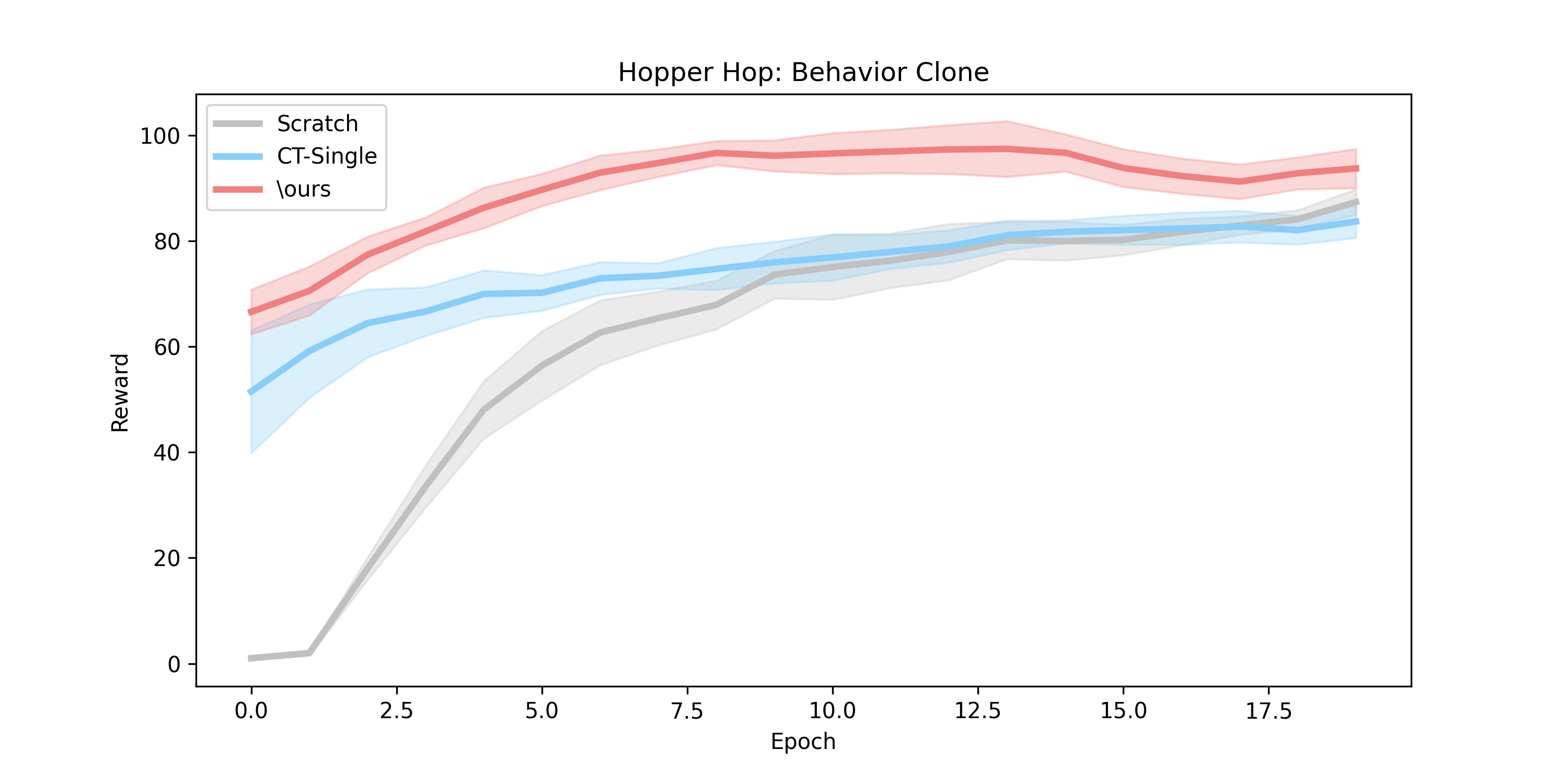}}
%   \vspace{-1.5em}
%   \caption{Cont. \& Non-adaptive}
 \end{subfigure}
 \hfill
 \begin{subfigure}[t]{0.19\columnwidth}
  \resizebox{\textwidth}{!}{\input{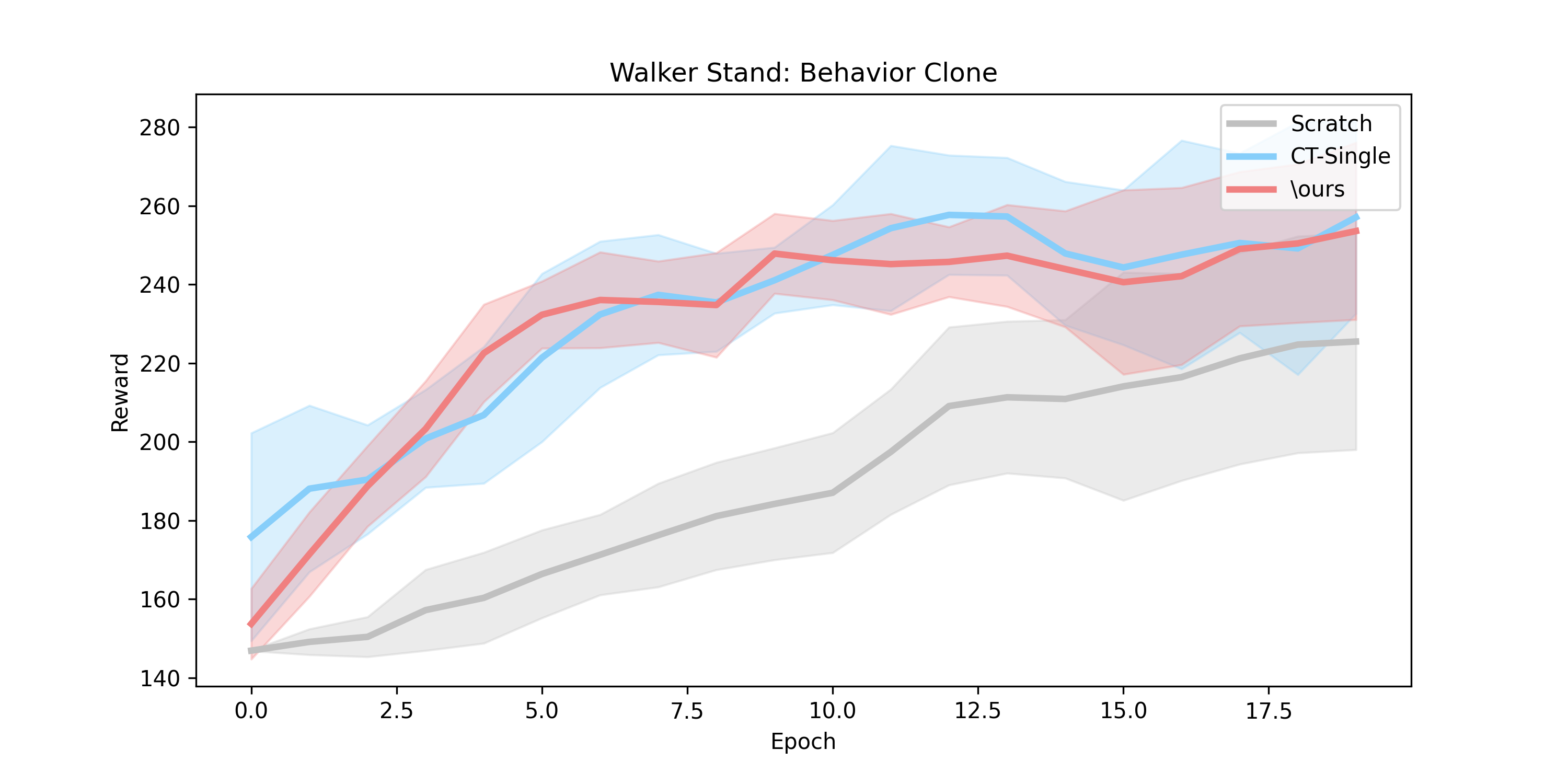}}
%   \vspace{-1.5em}
%   \caption{Cont. \& Adaptive}
 \end{subfigure} 
 \hfill
 \begin{subfigure}[t]{0.19\columnwidth}
  \resizebox{\textwidth}{!}{\input{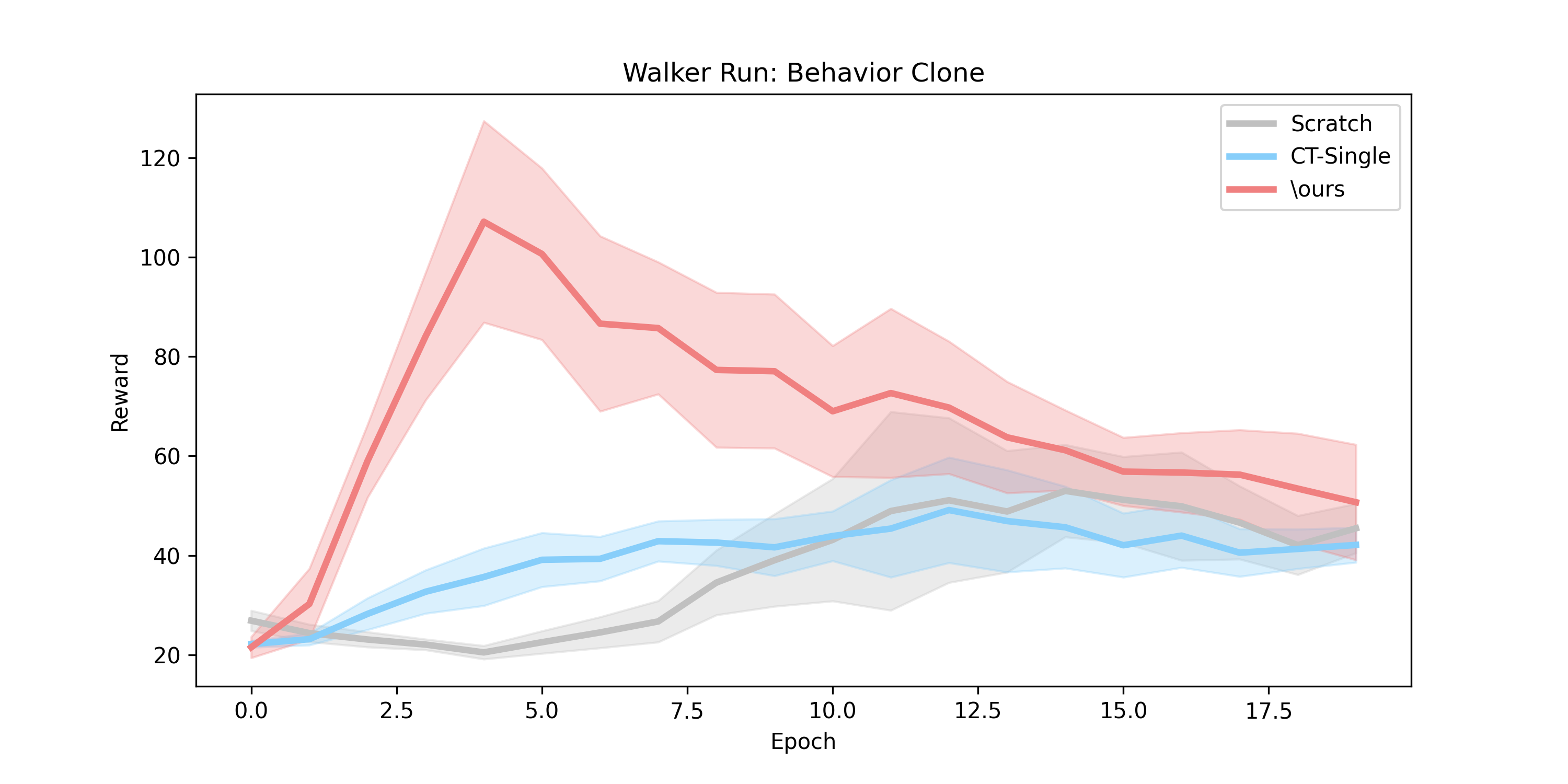}}
%   \vspace{-1.5em}
%   \caption{Cont. \& Adaptive}
 \end{subfigure} 
 \vspace{-0.5em}
\caption{Downstream learning rewards of \ours (\textcolor{red}{red}) compared with pretraining \ourmod with single-task data (\textcolor{cyan}{blue}) and training from scratch (\textcolor{gray}{gray}), using the \texttt{Random} pretraining dataset. Results are averaged over 3 random seeds.
}
% \vspace{-1em}
\label{fig:curves_seen_rand}
\end{figure}

\begin{figure}[!htbp]
\vspace{-1em}
 \centering

\rotatebox{90}{\scriptsize{\hspace{1cm}\textbf{RTG}}}
 \begin{subfigure}[t]{0.19\columnwidth}
  \resizebox{1\textwidth}{!}{\input{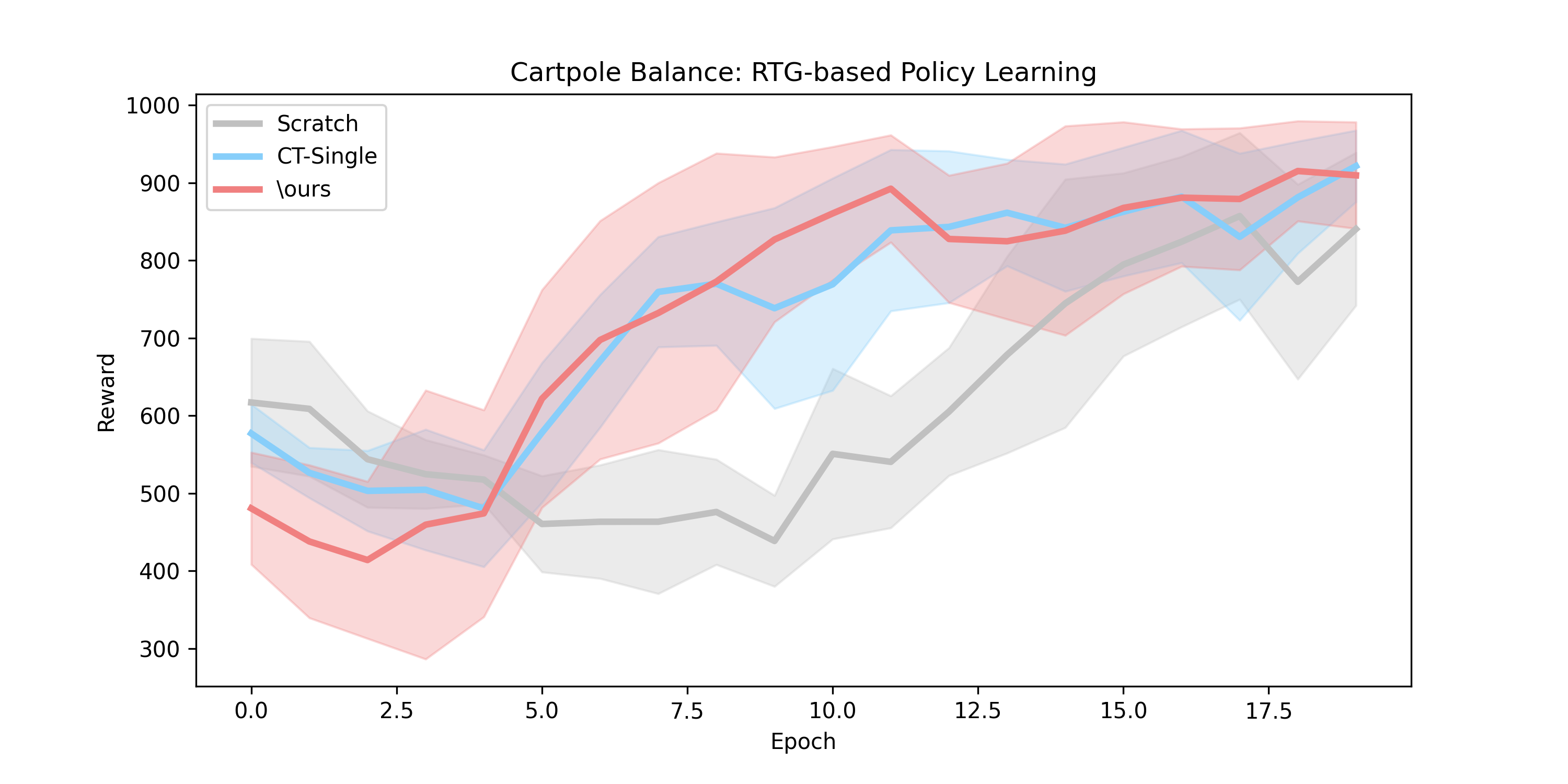}}
%   \vspace{-1.5em}
%   \caption{FoodCollector}
 \end{subfigure}
 \hfill
 \begin{subfigure}[t]{0.19\columnwidth}
  \resizebox{\textwidth}{!}{\input{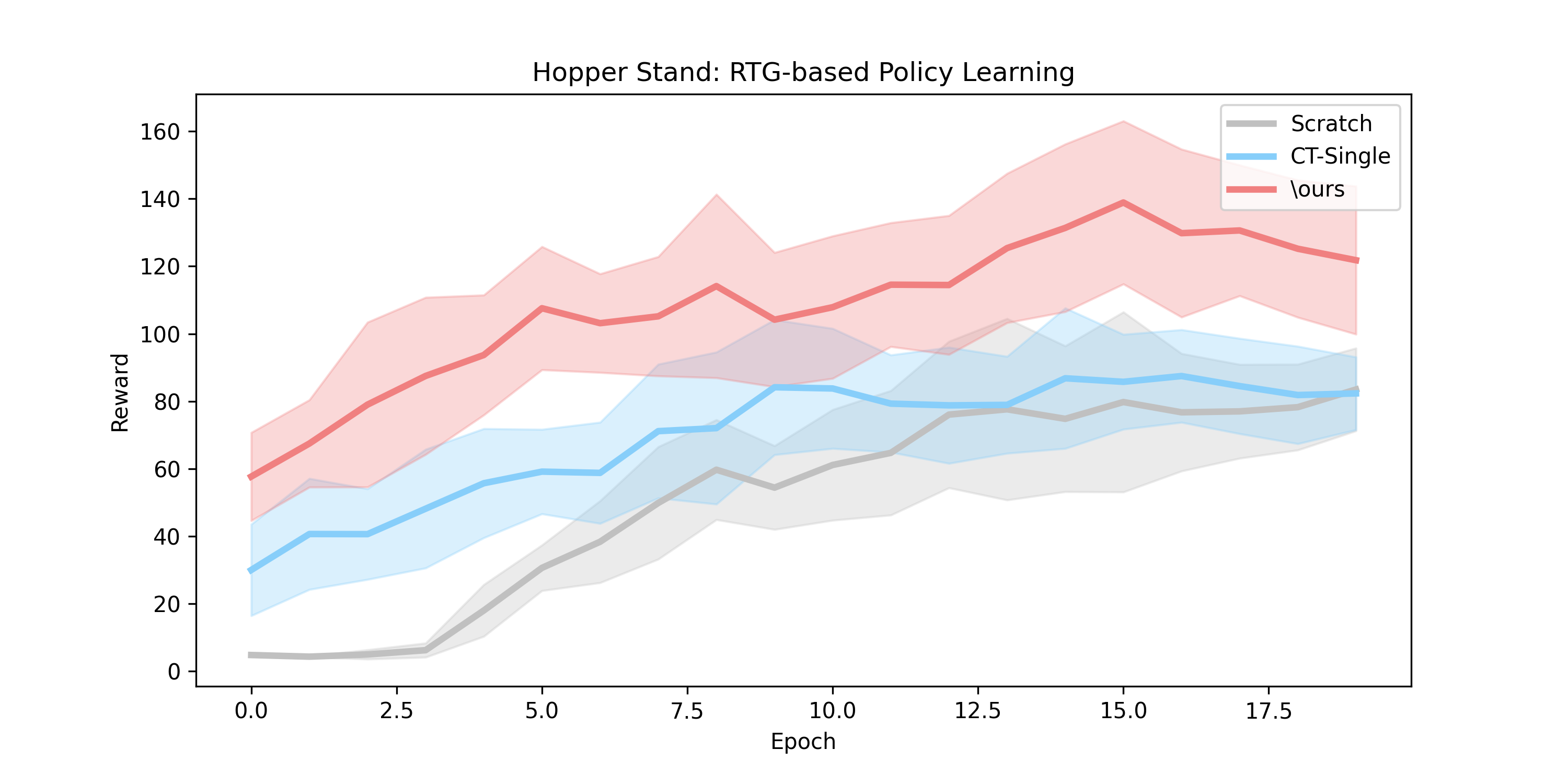}}
%   \vspace{-1.5em}
%   \caption{Disc. \& Adaptive}
 \end{subfigure}
 \hfill
 \begin{subfigure}[t]{0.19\columnwidth}
  \resizebox{\textwidth}{!}{\input{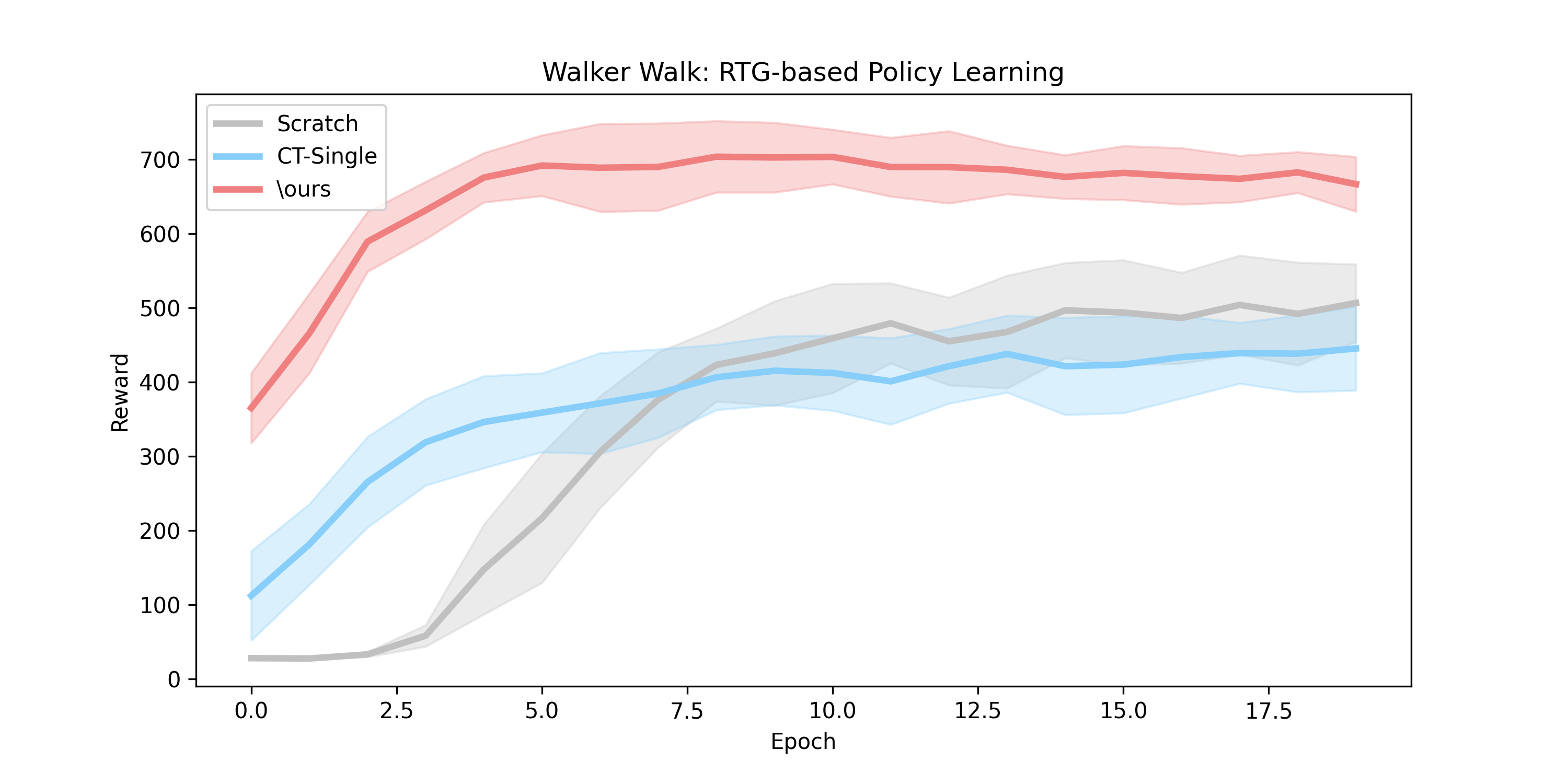}}
%   \vspace{-1.5em}
%   \caption{Cont. \& Non-adaptive}
 \end{subfigure}
 \hfill
 \begin{subfigure}[t]{0.19\columnwidth}
  \resizebox{\textwidth}{!}{\input{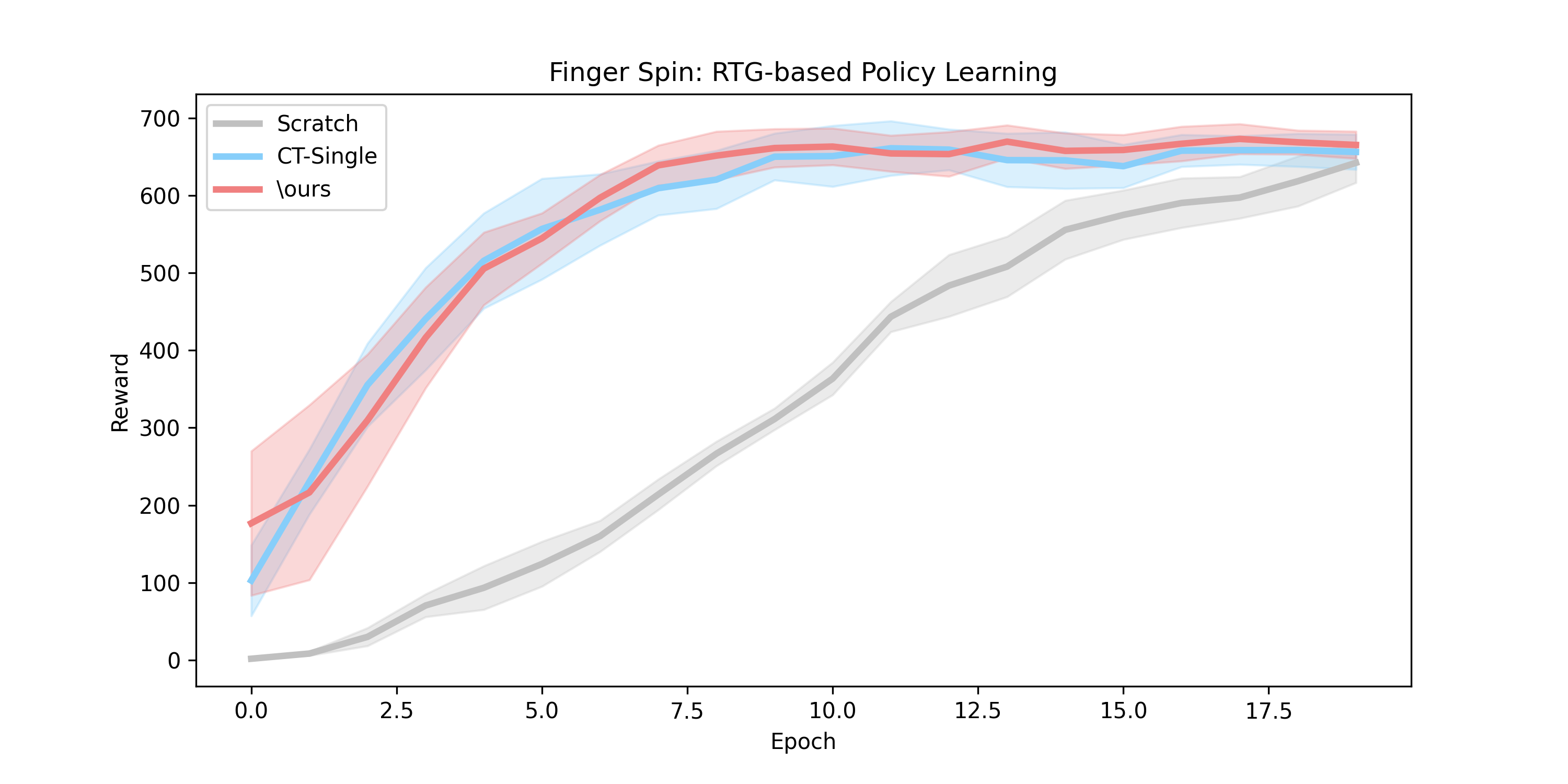}}
%   \vspace{-1.5em}
%   \caption{Cont. \& Adaptive}
 \end{subfigure} 
 \hfill
 \begin{subfigure}[t]{0.19\columnwidth}
  \resizebox{\textwidth}{!}{\input{figures/curves_rand/finger_spin_rtg_compare}}
%   \vspace{-1.5em}
%   \caption{Cont. \& Adaptive}
 \end{subfigure} 
%  \vspace{0.5em}

\rotatebox{90}{\scriptsize{\hspace{1cm}\textbf{BC}}}
 \begin{subfigure}[t]{0.19\columnwidth}
  \resizebox{1.04\textwidth}{!}{\input{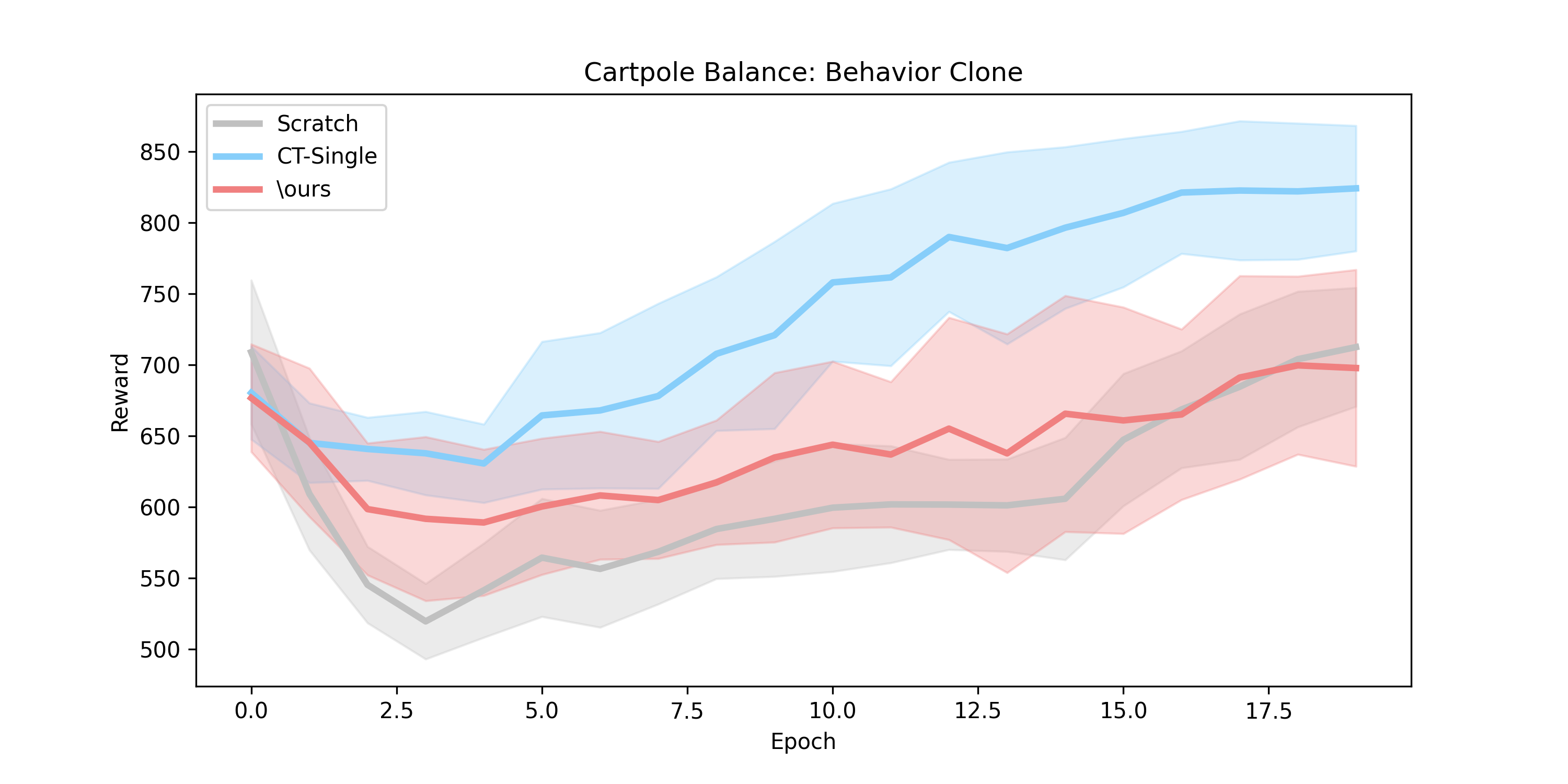}}
%   \vspace{-1.5em}
%   \caption{FoodCollector}
 \end{subfigure}
 \hfill
 \begin{subfigure}[t]{0.19\columnwidth}
  \resizebox{\textwidth}{!}{\input{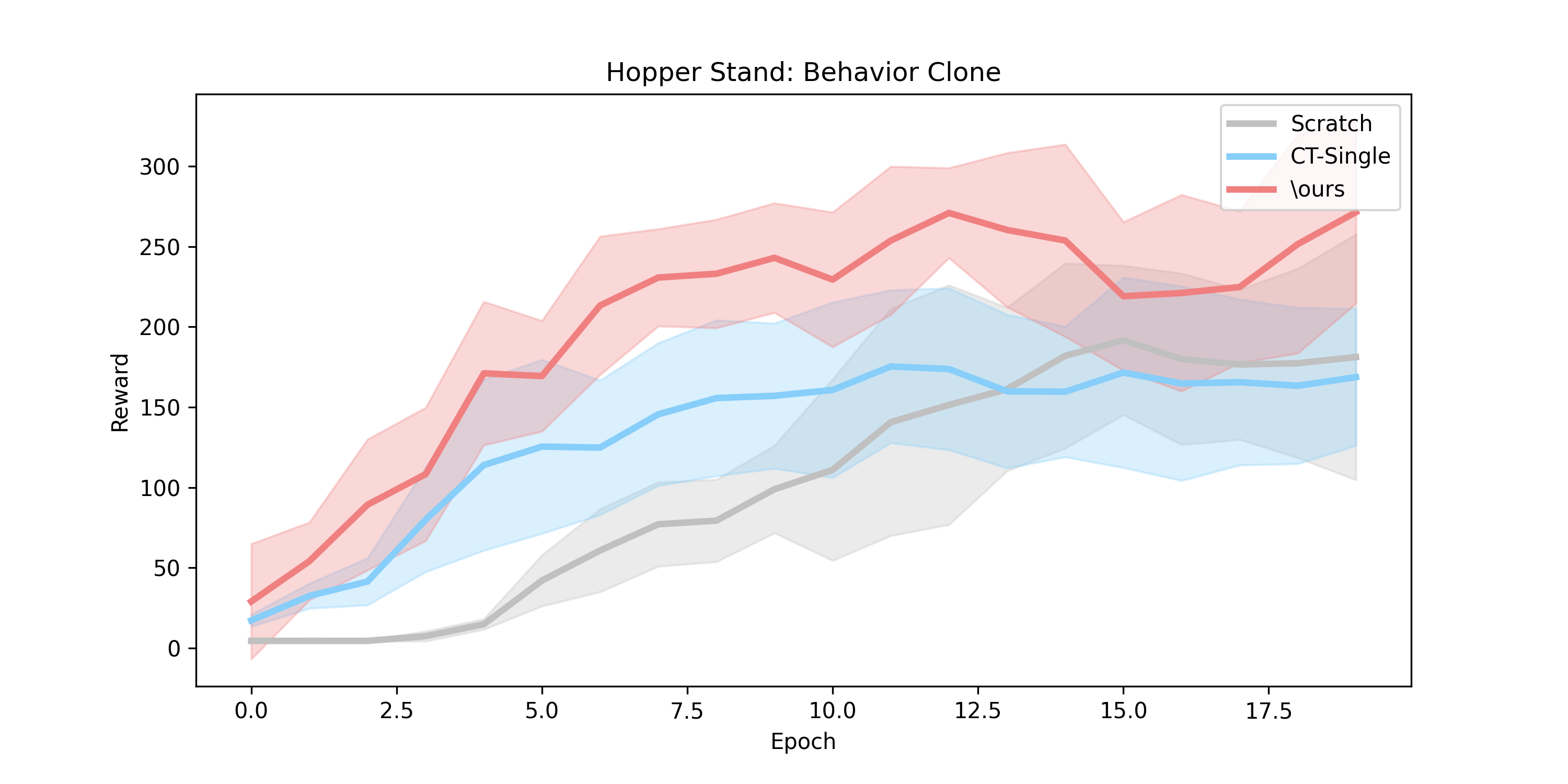}}
%   \vspace{-1.5em}
%   \caption{Disc. \& Adaptive}
 \end{subfigure}
 \hfill
 \begin{subfigure}[t]{0.19\columnwidth}
  \resizebox{\textwidth}{!}{\input{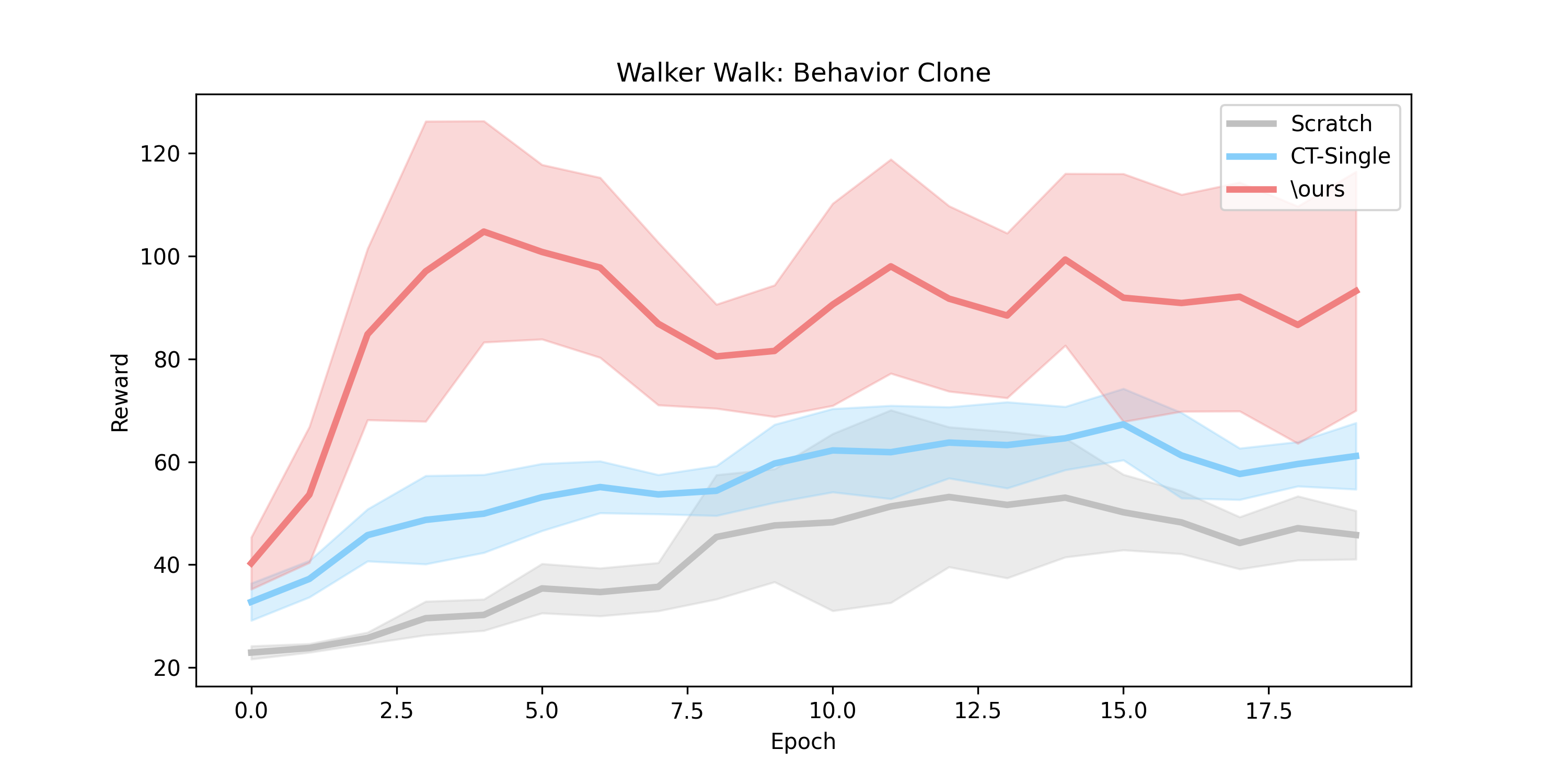}}
%   \vspace{-1.5em}
%   \caption{Cont. \& Non-adaptive}
 \end{subfigure}
 \hfill
 \begin{subfigure}[t]{0.19\columnwidth}
  \resizebox{\textwidth}{!}{\input{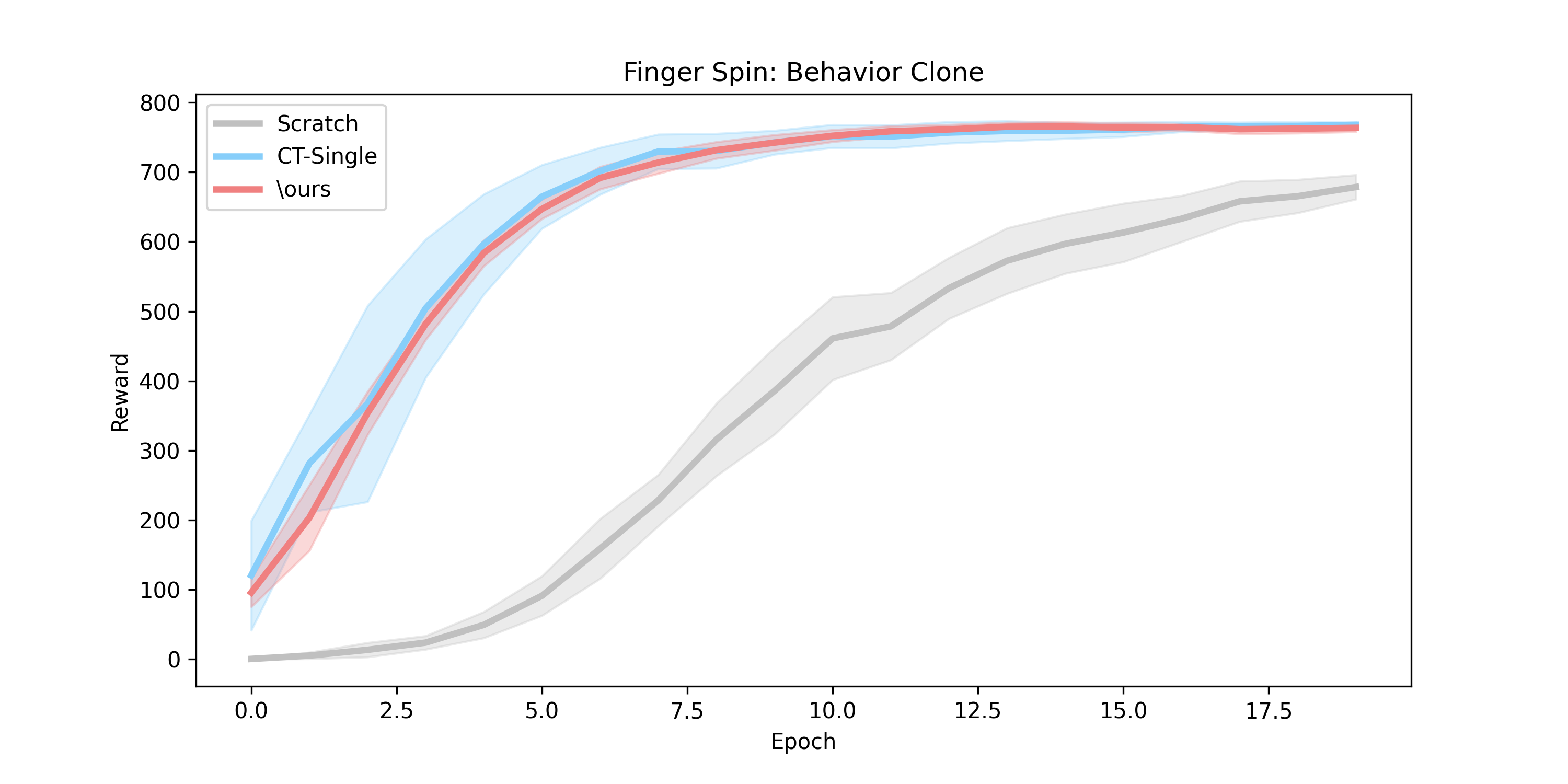}}
%   \vspace{-1.5em}
%   \caption{Cont. \& Adaptive}
 \end{subfigure} 
 \hfill
 \begin{subfigure}[t]{0.19\columnwidth}
  \resizebox{\textwidth}{!}{\input{figures/curves_rand/finger_spin_bc_compare}}
%   \vspace{-1.5em}
%   \caption{Cont. \& Adaptive}
 \end{subfigure} 
 \vspace{-0.5em}
\caption{Downstream learning rewards in \textbf{unseen tasks} and domains of \ours (\textcolor{red}{red}) compared with pretraining \ourmod with single-task data (\textcolor{cyan}{blue}) and training from scratch (\textcolor{gray}{gray}), using the \texttt{Random} pretraining dataset. Results are averaged over 3 random seeds.
}
% \vspace{-1em}
\label{fig:curves_unseen_rand}
\end{figure}

\textbf{Curves with the \texttt{Random} Dataset.}
The results in
\cref{fig:curves_seen} and \cref{fig:curves_unseen} are generated with the \texttt{Exploratory} pretraining dataset. Now, we show the performance of models pretrained using the \texttt{Random} dataset in seen tasks and unseen tasks in \cref{fig:curves_seen_rand} and \cref{fig:curves_unseen_rand}, respectively.

Although the single-task pretrained model consistently outperforms training from scratch when pretrained with the \texttt{Exploratory} dataset, it sometime gets worst-than-scratch downstream performance when pretrained with the Random dataset. Therefore, it is challenging to overcome the distribution shift problem. 
In contrast, \ours still achieves much better performance than training from scratch in all tested tasks, which verifies the resilience of \ours due to multi-task self-supervised pretraining.

\begin{wrapfigure}{r}{0.4\textwidth}
\vspace{-1.5em}
  \centering
  \begin{subfigure}[t]{0.18\textwidth}
      \includegraphics[width=\textwidth]{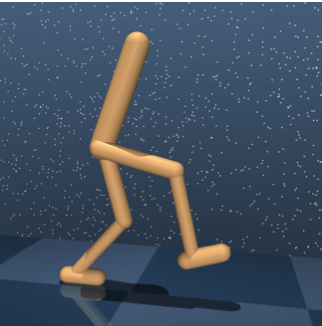}
  \end{subfigure}
  \hfill
  \begin{subfigure}[t]{0.18\textwidth}
      \includegraphics[width=\textwidth]{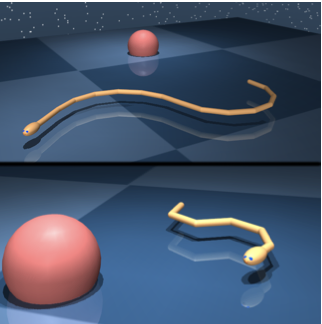}
  \end{subfigure}
 \vspace{-0.5em}
  \caption{Discrepancy between pretraining domains and selected downstream domains: \textbf{(left)} Walker domain. \textbf{(right)} Swimmer domain (6 and 15 links)}
  \label{fig:walker_swimmer}
 \vspace{-1em}
\end{wrapfigure}

\textbf{Generalizability of \ours: Tasks That Have Larger Discrepancy with Pretraining Tasks.}
In addition to the generalizability test we show in \cref{ssec:exp_res}, we evaluate the performance of the pretrained \ours model in other more challenging domains/tasks from DMC, that are significantly different from the pretraining tasks. These additional domain-tasks are: ball-in-cup-catch, finger-turn-hard, fish-swim, swimmer-swimmer6 and swimmer-swimmer15. Note that these agents have significantly different appearance and moving patterns compared to pretraining tasks, as visualized in \cref{fig:walker_swimmer}. 

\begin{figure}[!htbp]
\rotatebox{90}{\scriptsize{\hspace{1cm}\textbf{RTG}}}
 \begin{subfigure}[t]{0.19\columnwidth}
  \resizebox{1\textwidth}{!}{\input{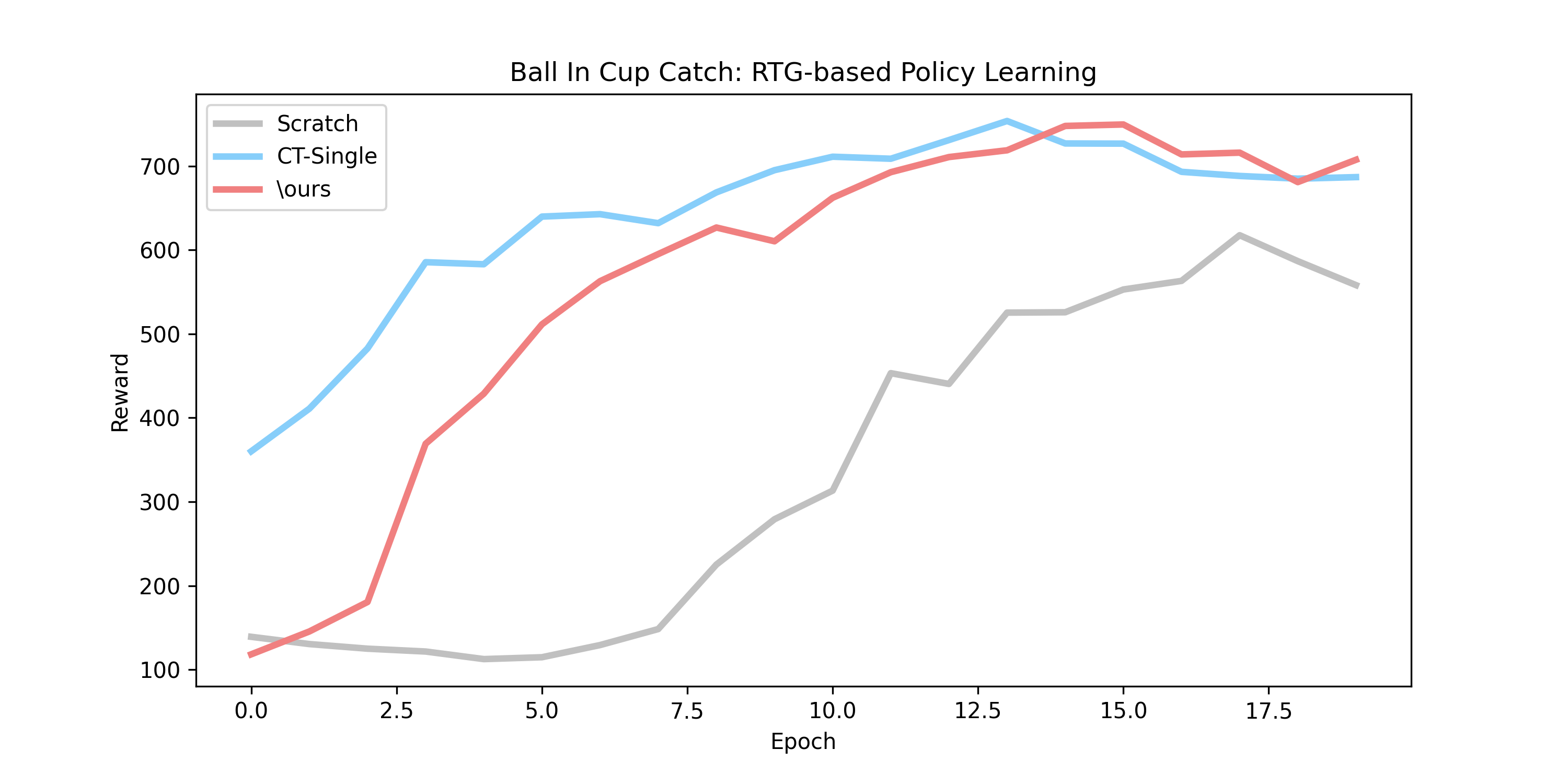}}
 \end{subfigure}
 \hfill
 \begin{subfigure}[t]{0.19\columnwidth}
  \resizebox{\textwidth}{!}{\input{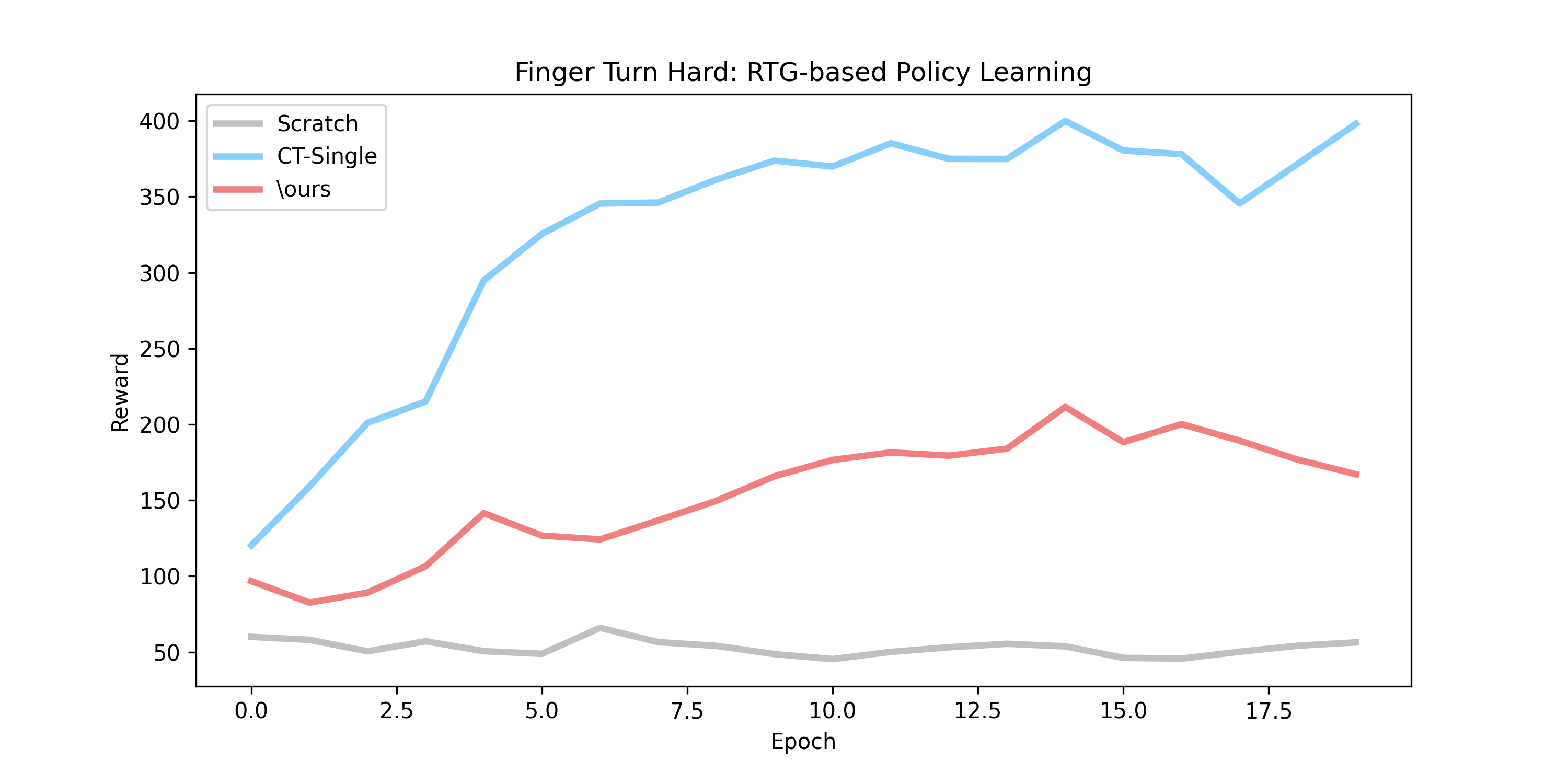}}
 \end{subfigure} 
 \hfill
 \begin{subfigure}[t]{0.19\columnwidth}
  \resizebox{\textwidth}{!}{\input{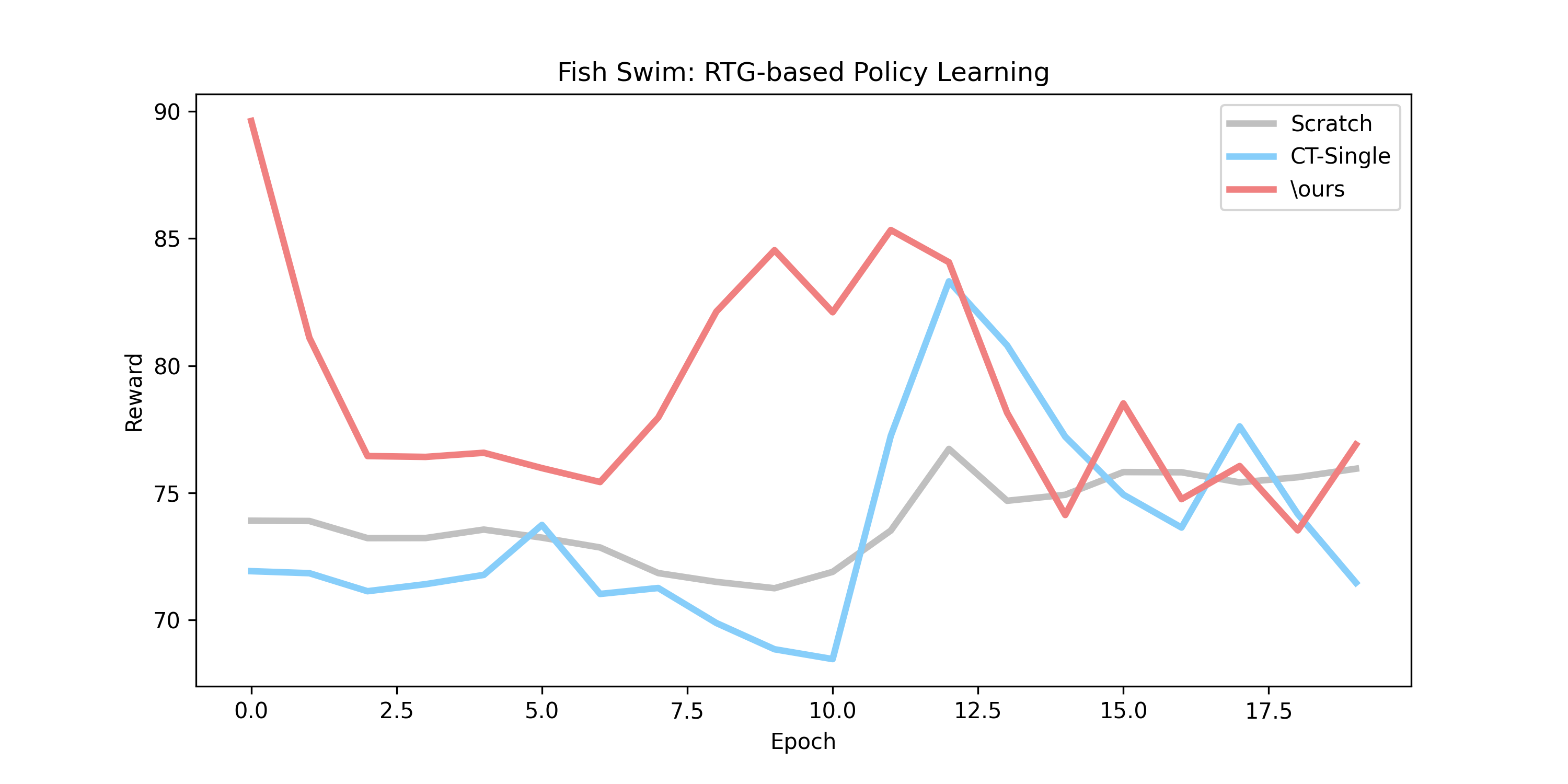}}
 \end{subfigure}
 \hfill
 \begin{subfigure}[t]{0.19\columnwidth}
  \resizebox{\textwidth}{!}{\input{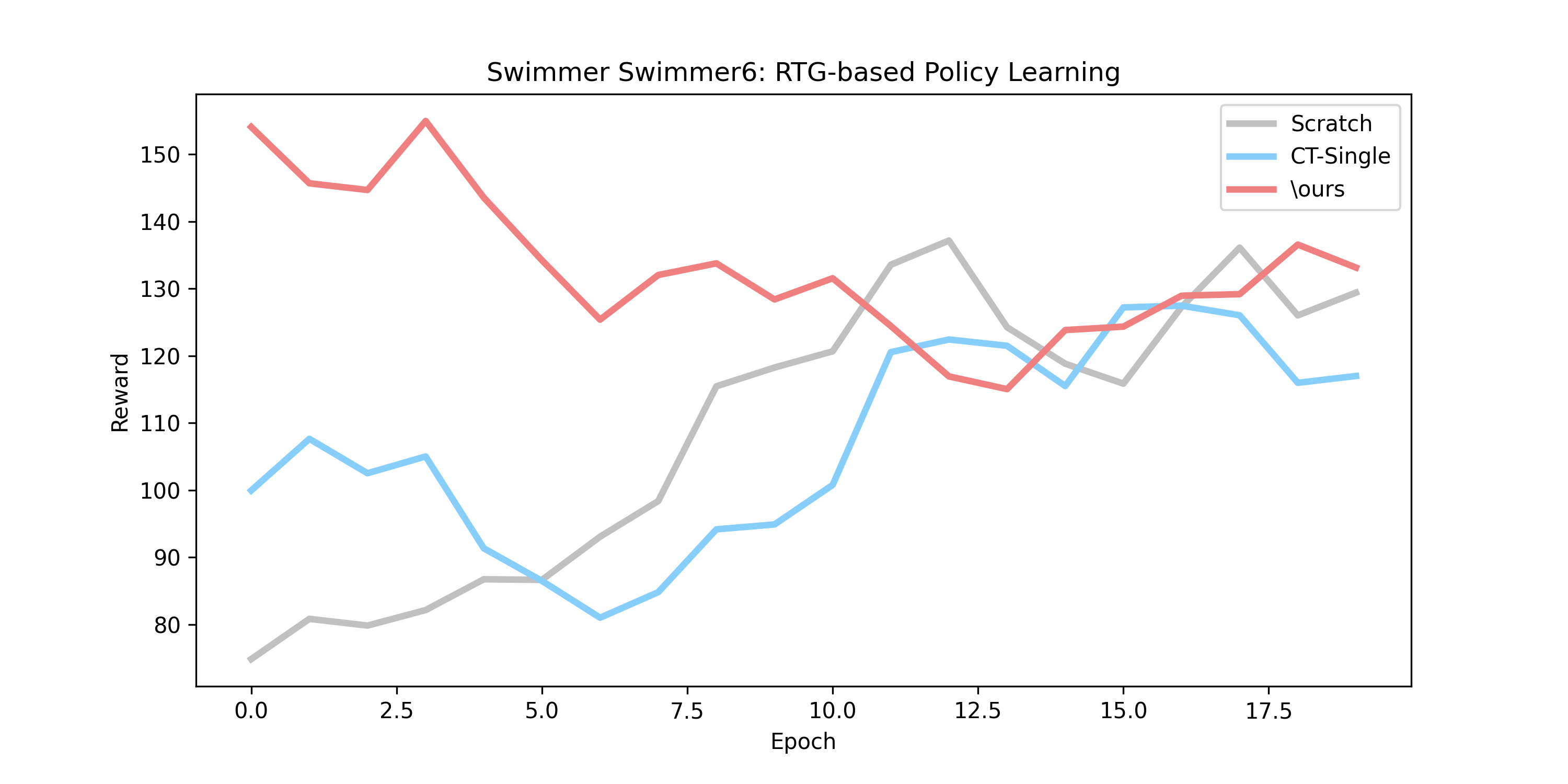}}
 \end{subfigure}
 \hfill
 \begin{subfigure}[t]{0.19\columnwidth}
  \resizebox{\textwidth}{!}{\input{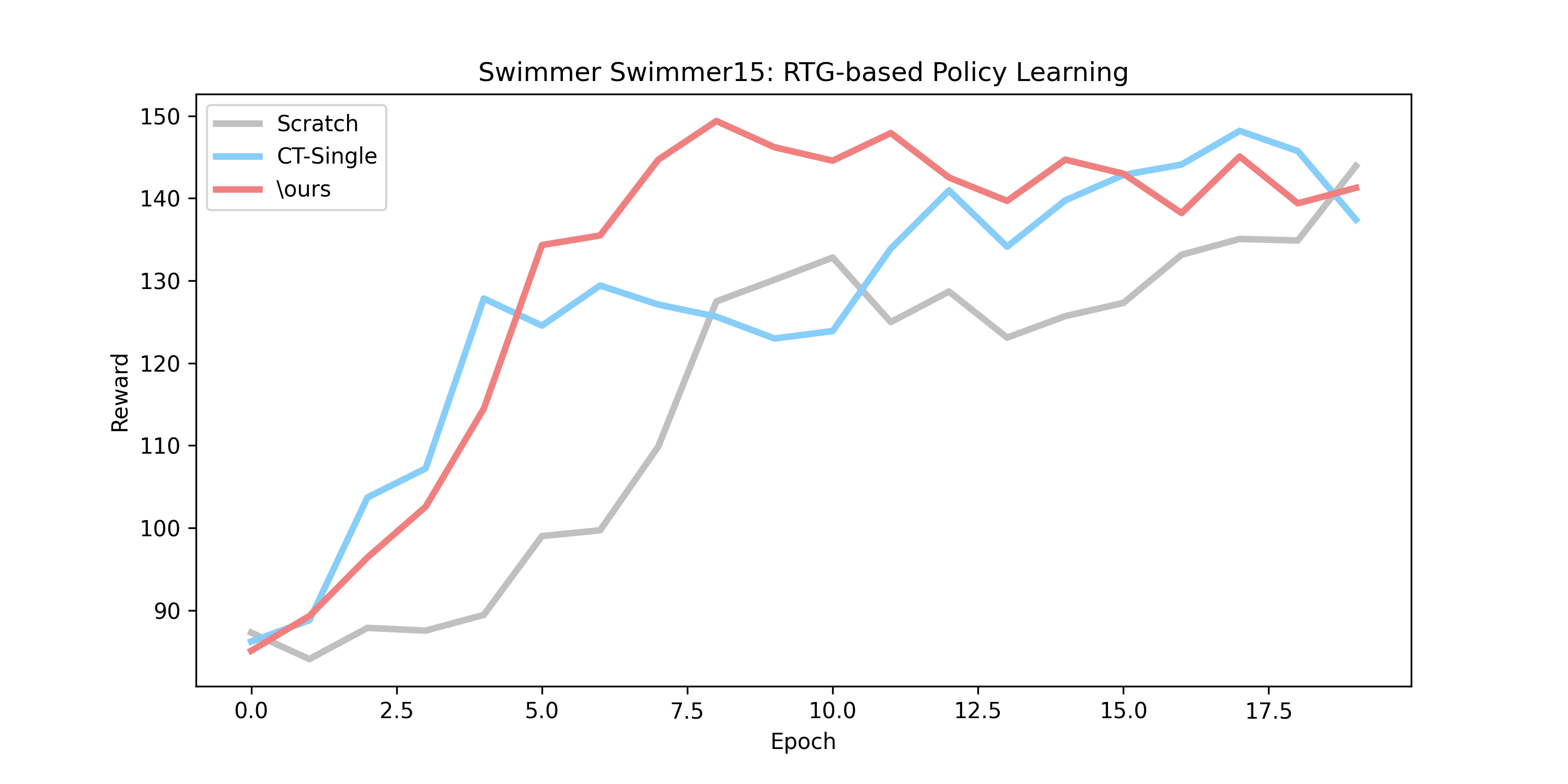}}
 \end{subfigure}

\rotatebox{90}{\scriptsize{\hspace{1cm}\textbf{BC}}}
 \begin{subfigure}[t]{0.19\columnwidth}
  \resizebox{1\textwidth}{!}{\input{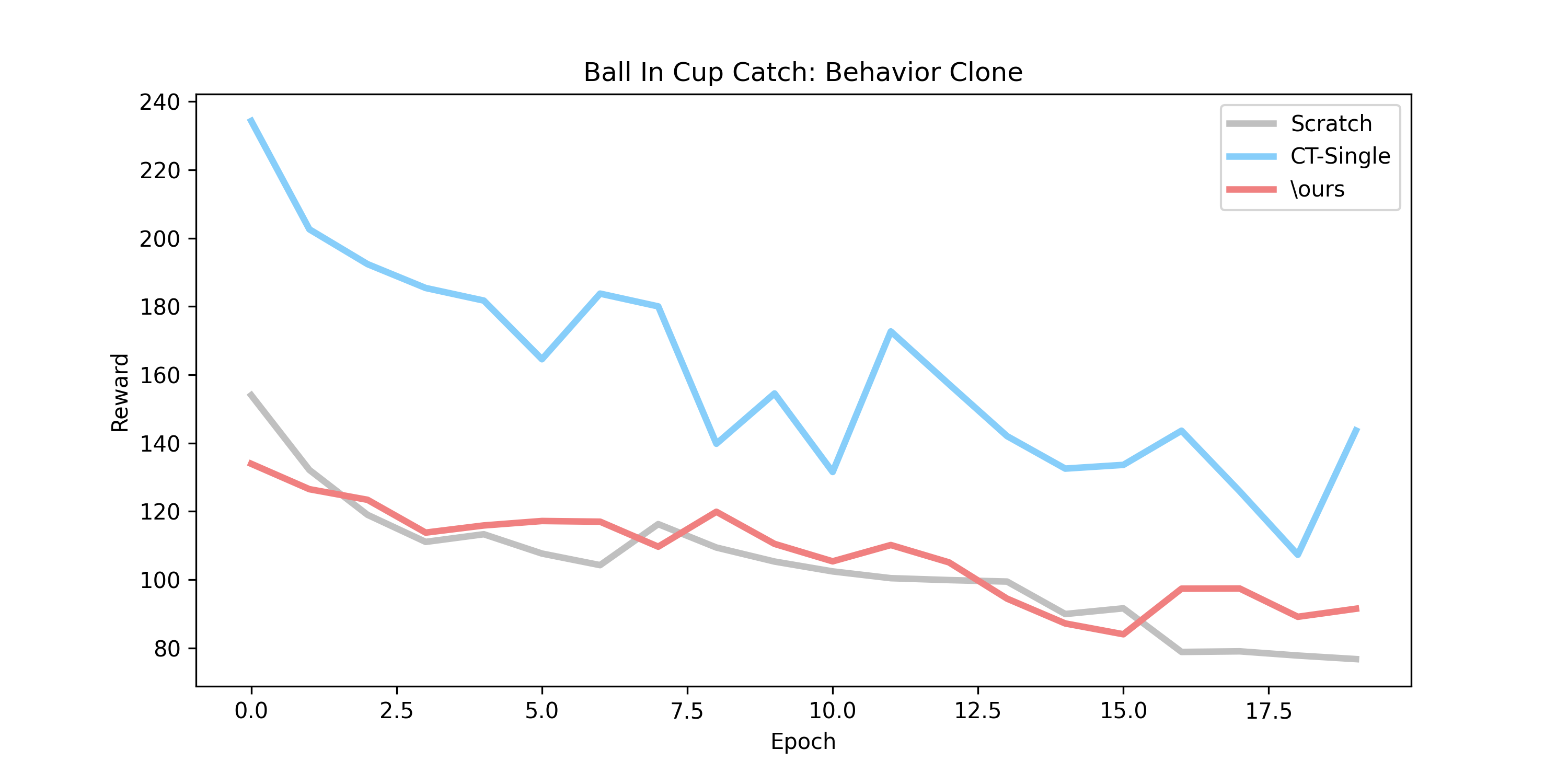}}
 \end{subfigure}
 \hfill
 \begin{subfigure}[t]{0.19\columnwidth}
  \resizebox{\textwidth}{!}{\input{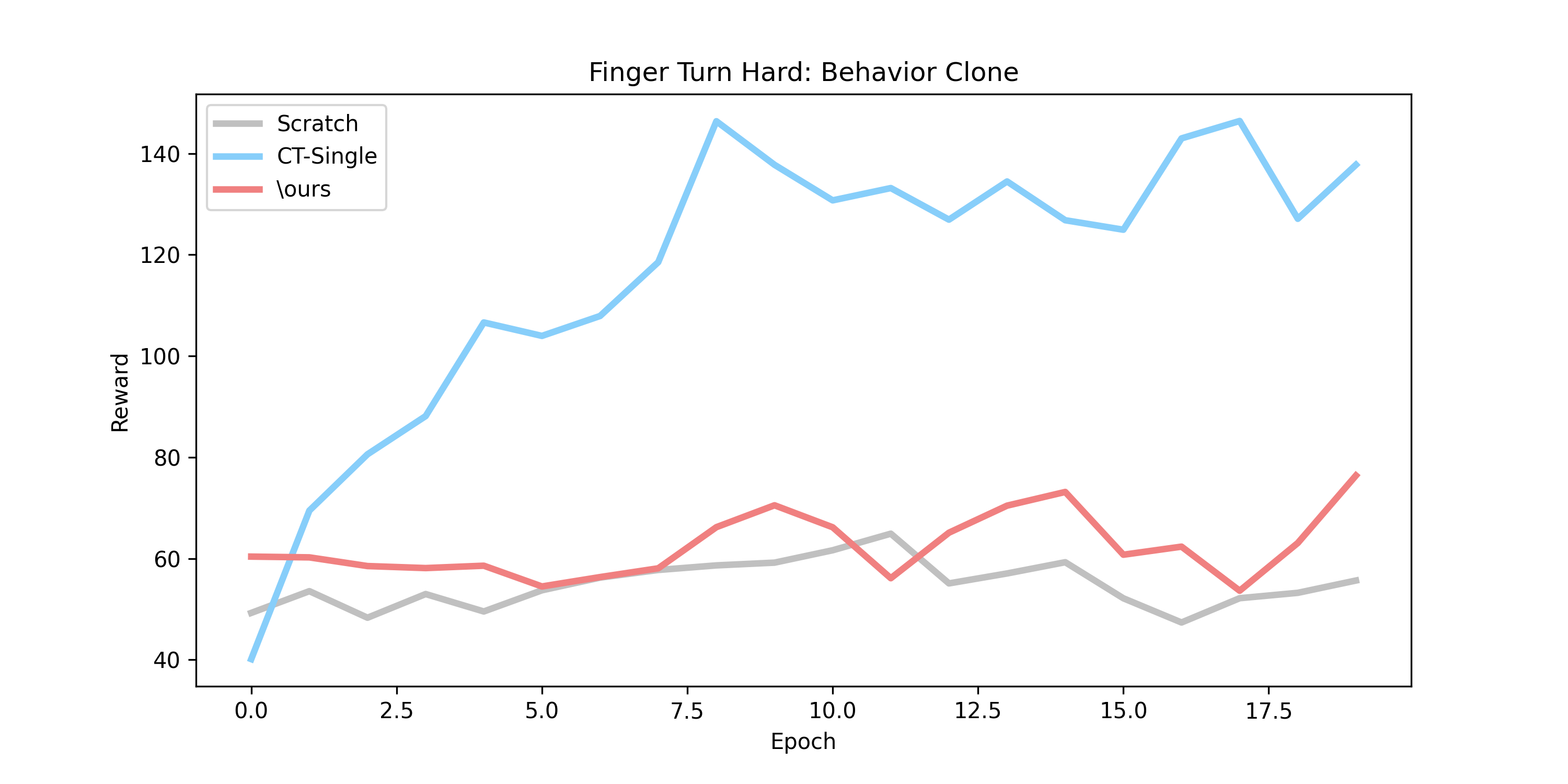}}
 \end{subfigure} 
 \hfill
 \begin{subfigure}[t]{0.19\columnwidth}
  \resizebox{\textwidth}{!}{\input{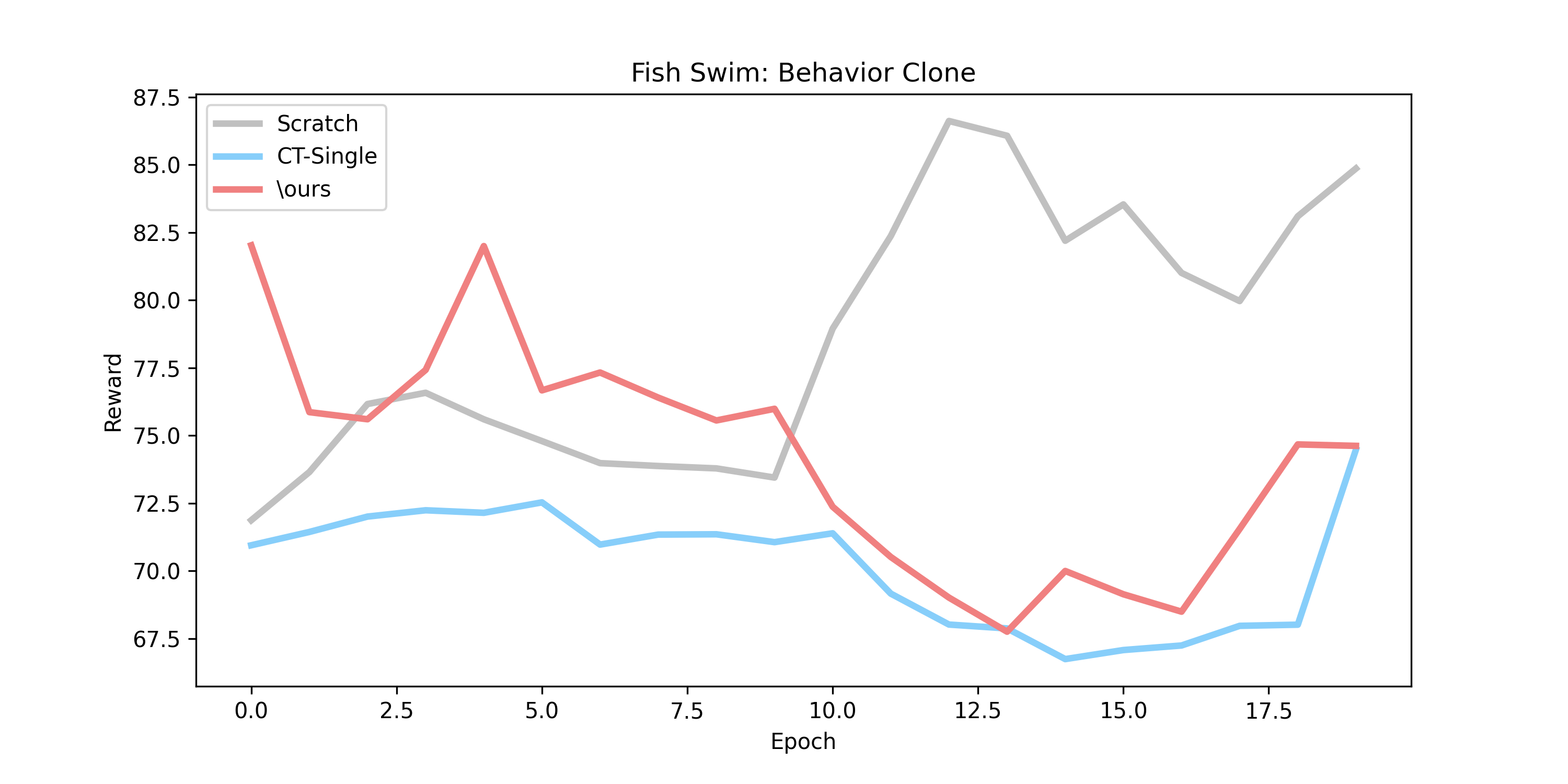}}
 \end{subfigure}
 \hfill
 \begin{subfigure}[t]{0.19\columnwidth}
  \resizebox{\textwidth}{!}{\input{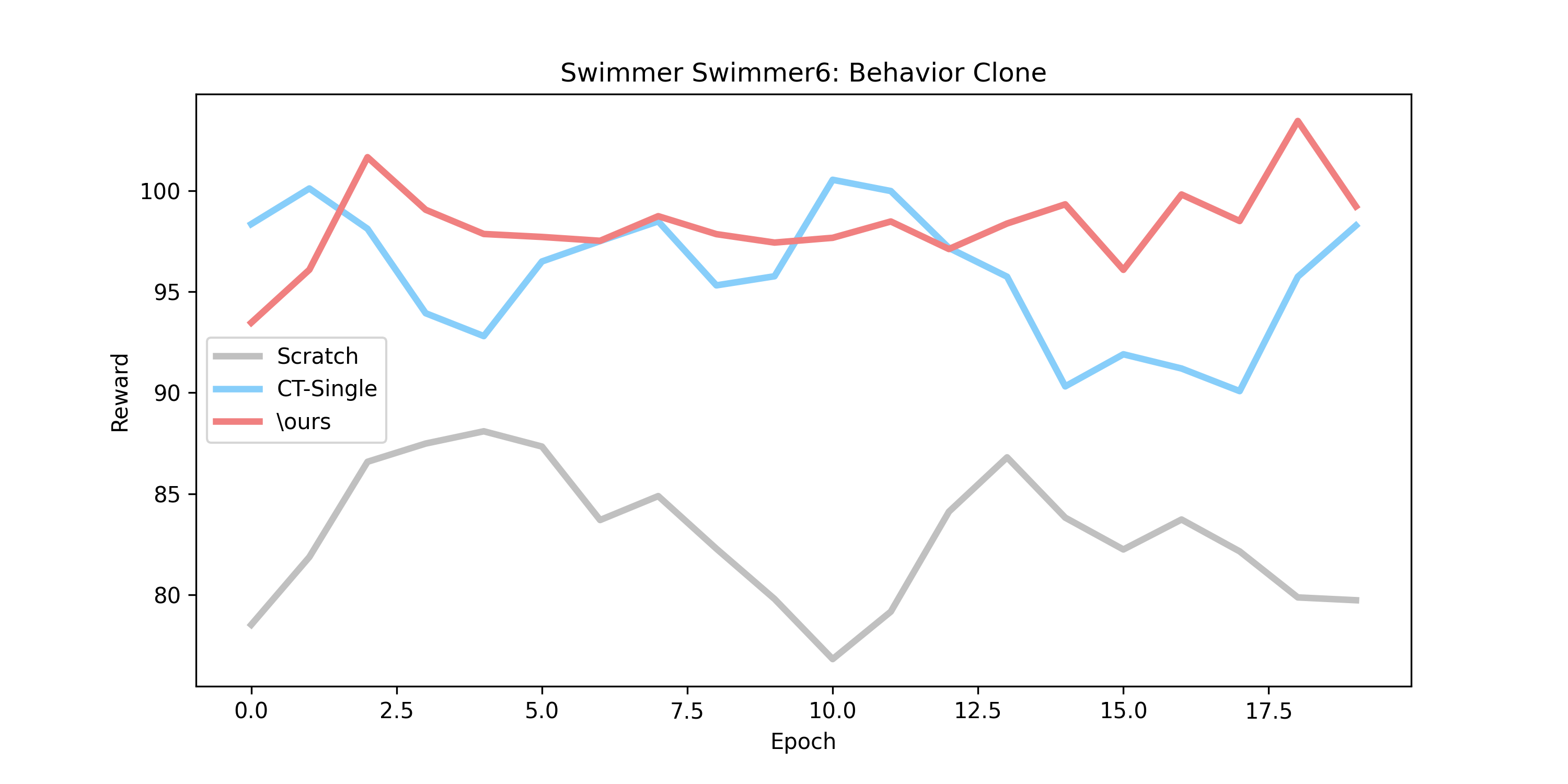}}
 \end{subfigure}
 \hfill
 \begin{subfigure}[t]{0.19\columnwidth}
  \resizebox{\textwidth}{!}{\input{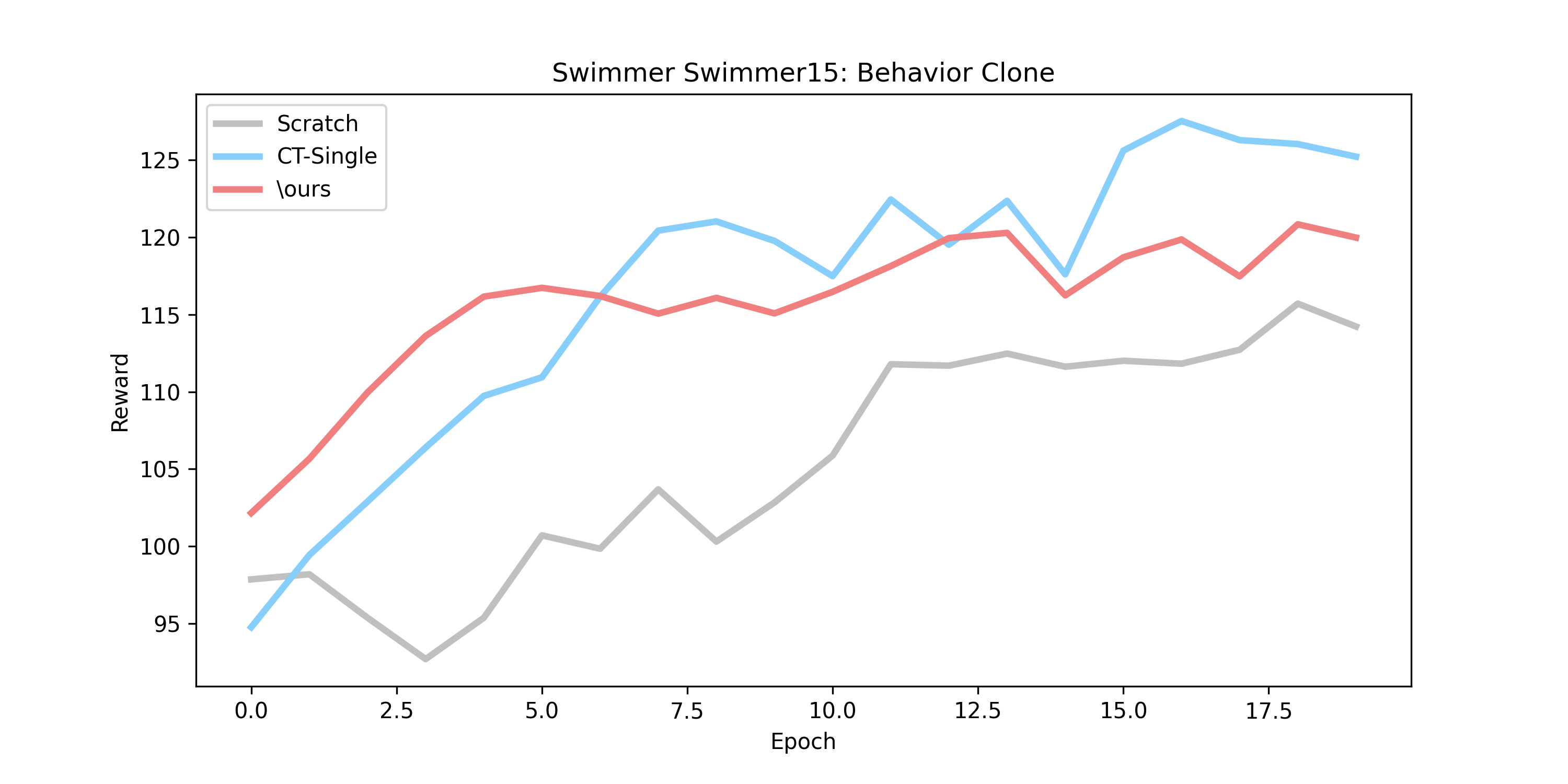}}
 \end{subfigure} 
 \vspace{-0.5em}
\caption{
Downstream learning rewards of \ours (\textcolor{red}{red}) in challenging tasks that have larger discrepancy with pretraining tasks, using the \texttt{Exploratory} pretraining dataset. Results are from 1 random seed.
}
\label{fig:curves_add_expl}
\end{figure}

\begin{figure}[!htbp]
\rotatebox{90}{\scriptsize{\hspace{1cm}\textbf{RTG}}}
 \begin{subfigure}[t]{0.19\columnwidth}
  \resizebox{1\textwidth}{!}{\input{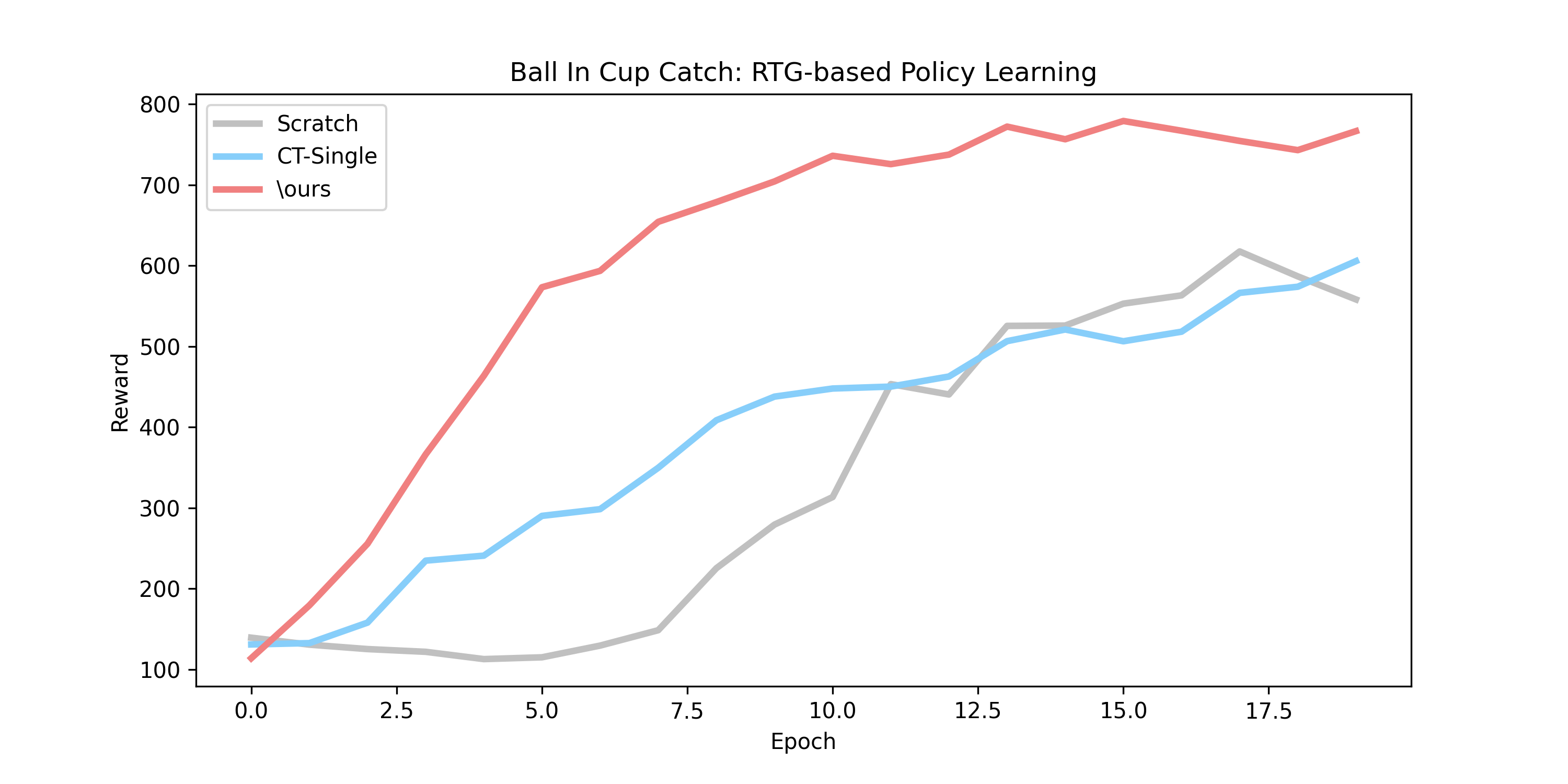}}
 \end{subfigure}
 \hfill
 \begin{subfigure}[t]{0.19\columnwidth}
  \resizebox{\textwidth}{!}{\input{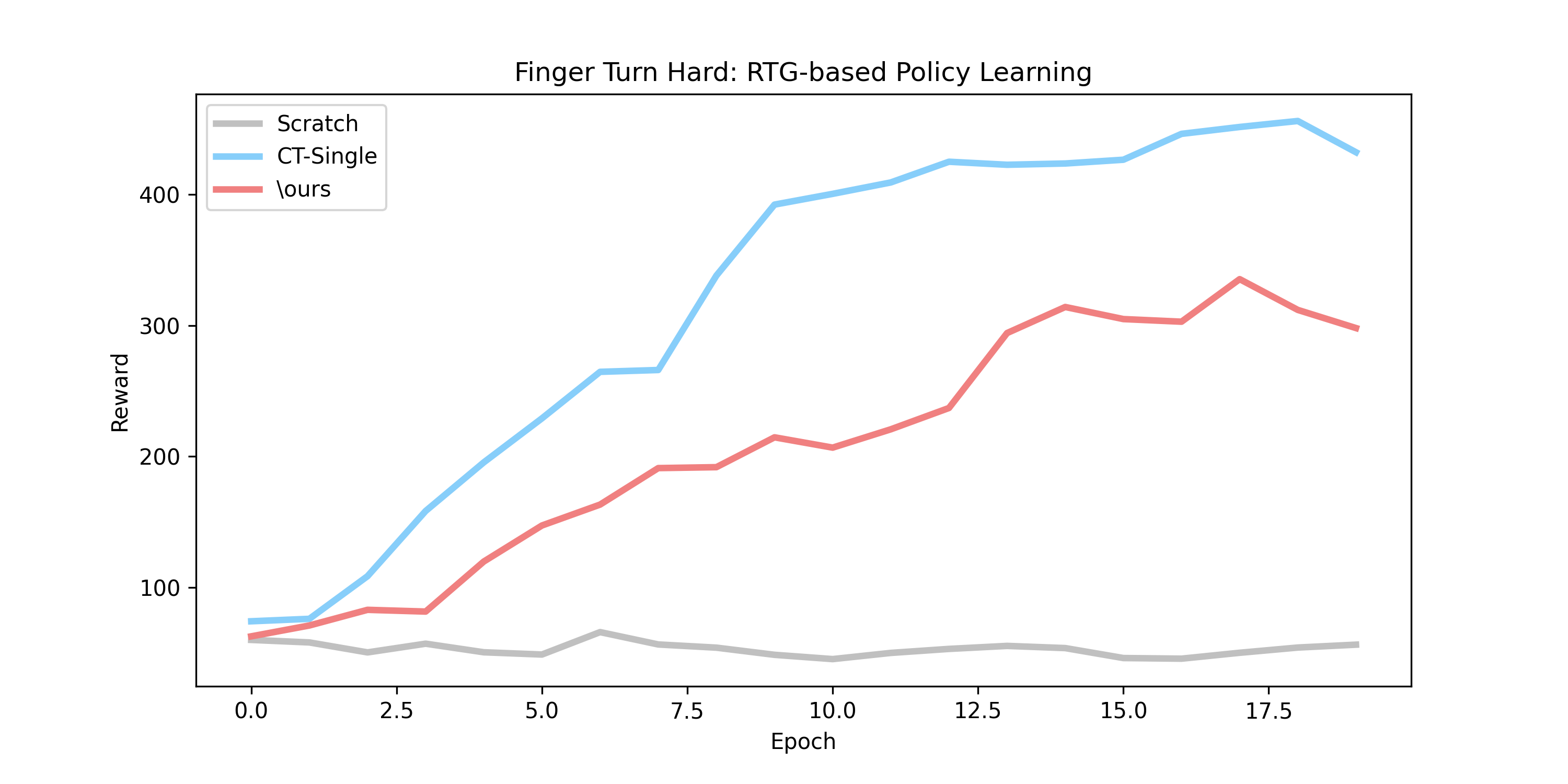}}
 \end{subfigure} 
 \hfill
 \begin{subfigure}[t]{0.19\columnwidth}
  \resizebox{\textwidth}{!}{\input{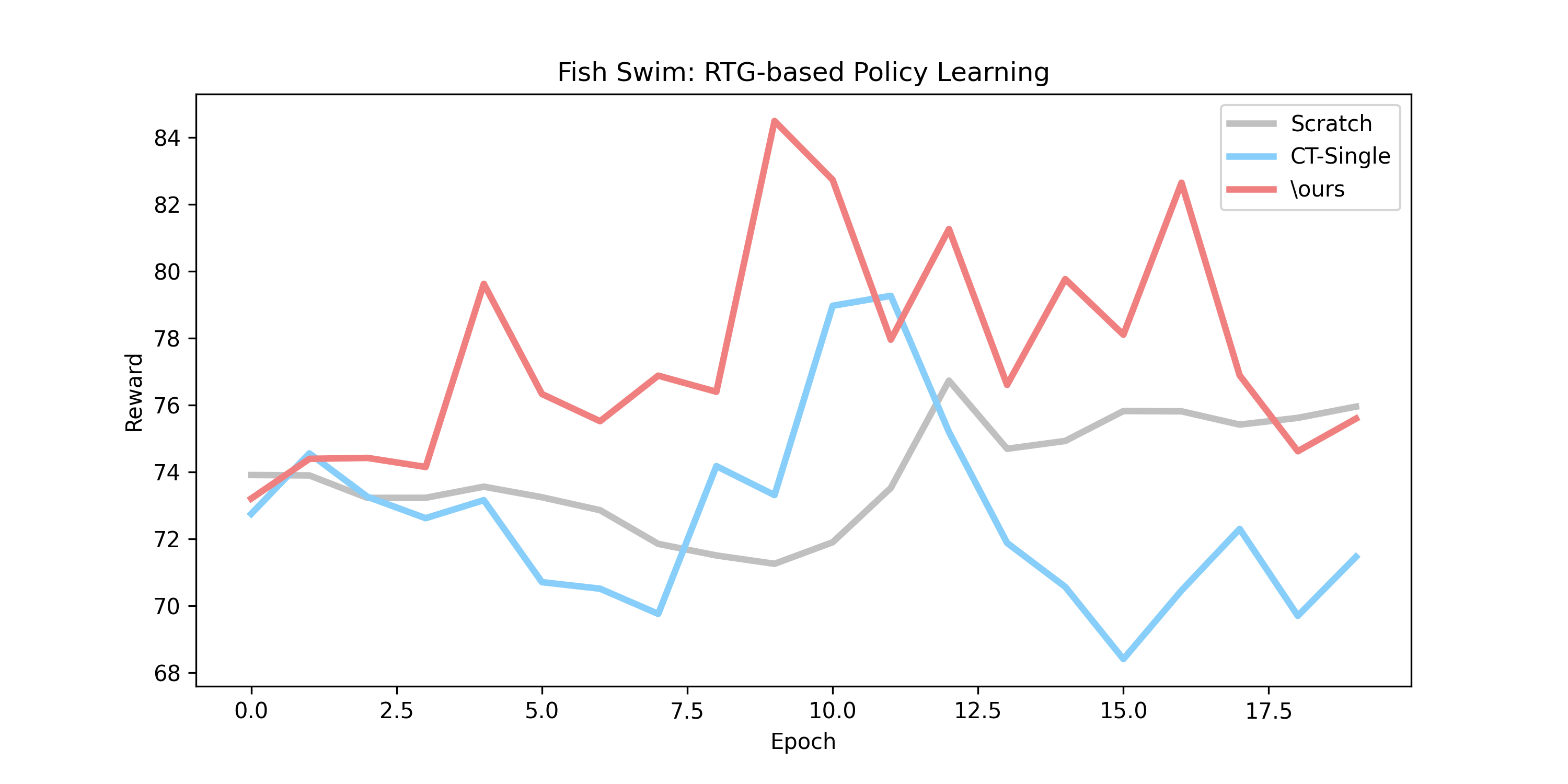}}
 \end{subfigure}
 \hfill
 \begin{subfigure}[t]{0.19\columnwidth}
  \resizebox{\textwidth}{!}{\input{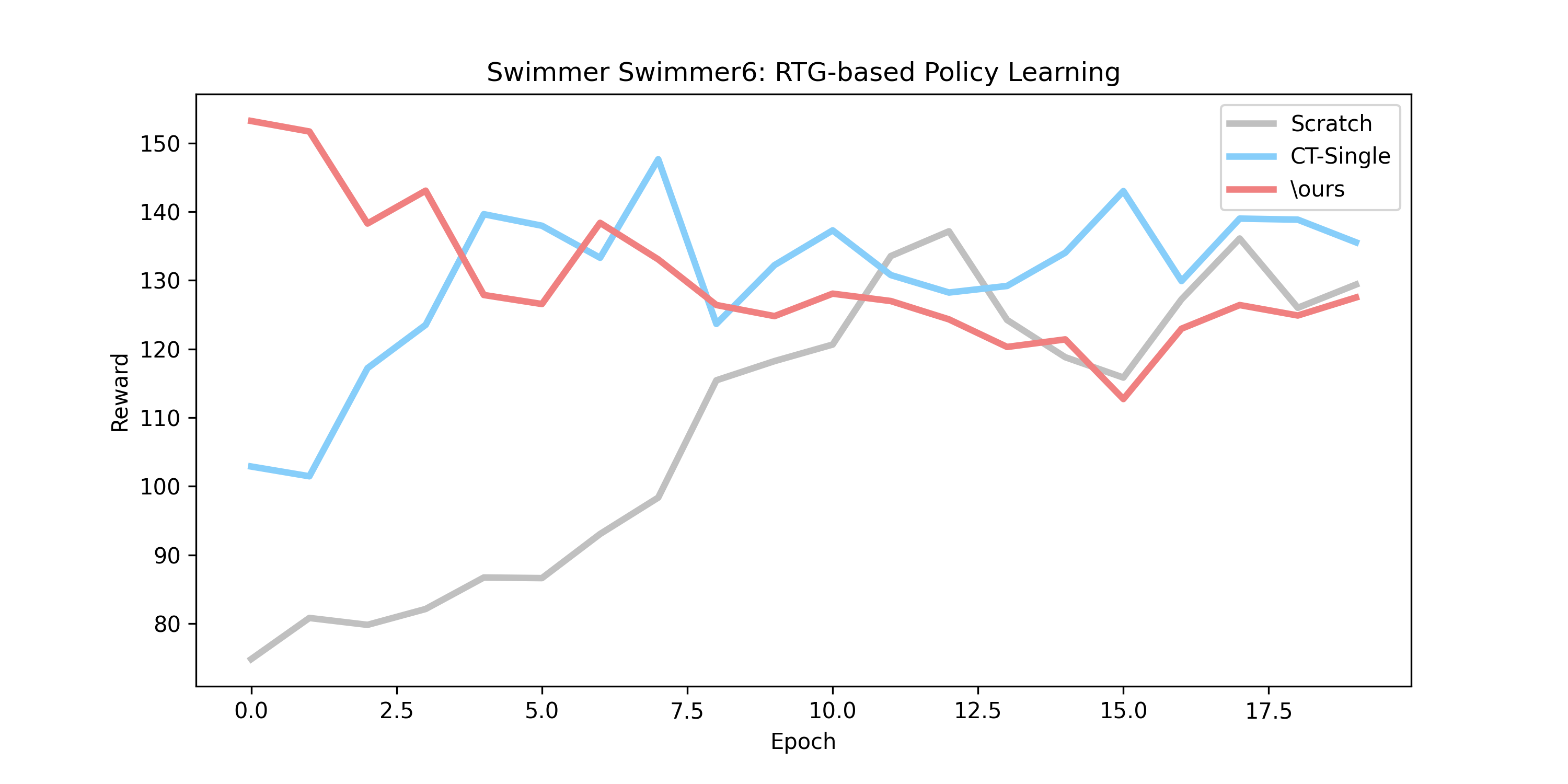}}
 \end{subfigure}
 \hfill
 \begin{subfigure}[t]{0.19\columnwidth}
  \resizebox{\textwidth}{!}{\input{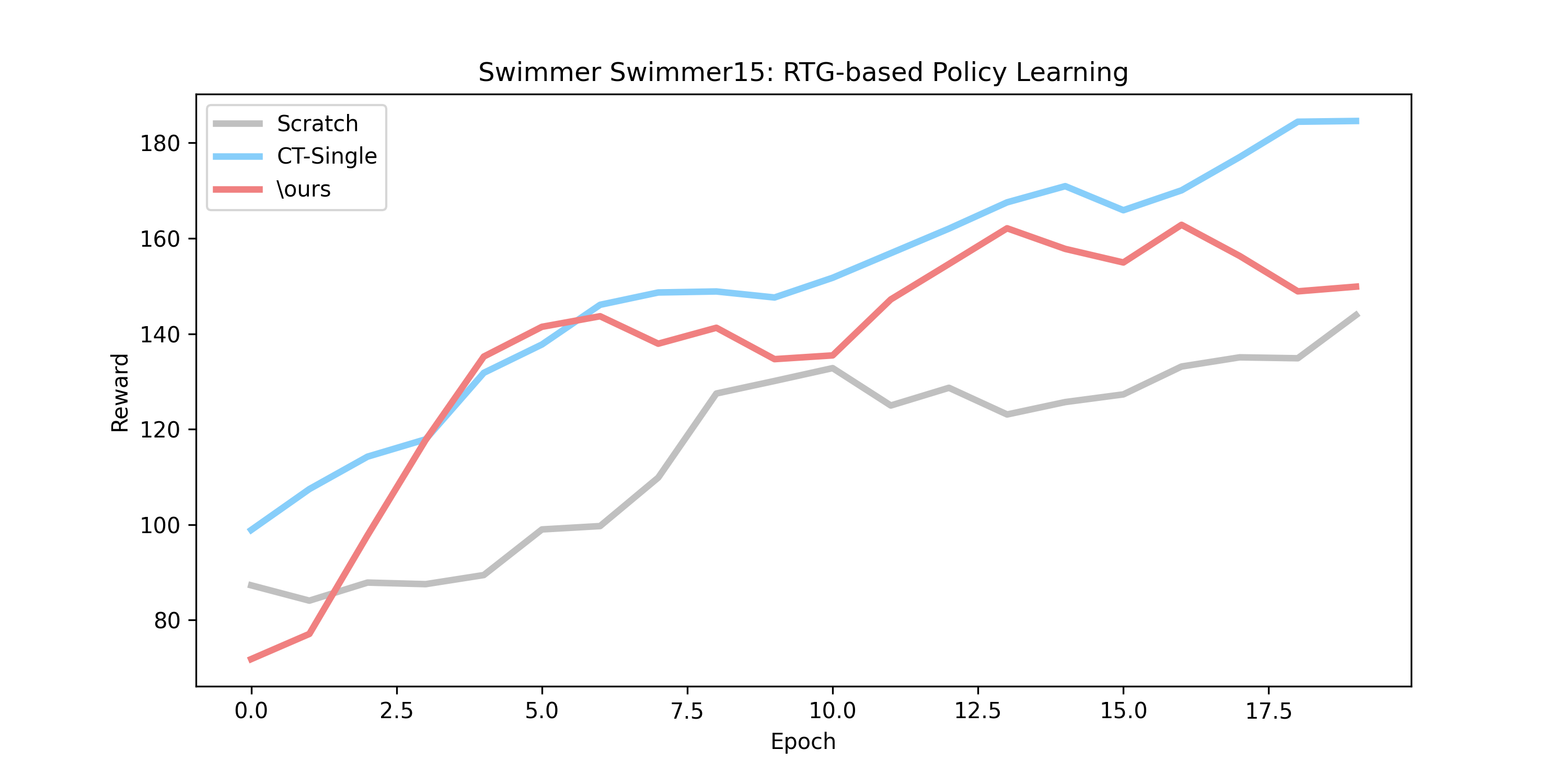}}
 \end{subfigure}

\rotatebox{90}{\scriptsize{\hspace{1cm}\textbf{BC}}}
 \begin{subfigure}[t]{0.19\columnwidth}
  \resizebox{1\textwidth}{!}{\input{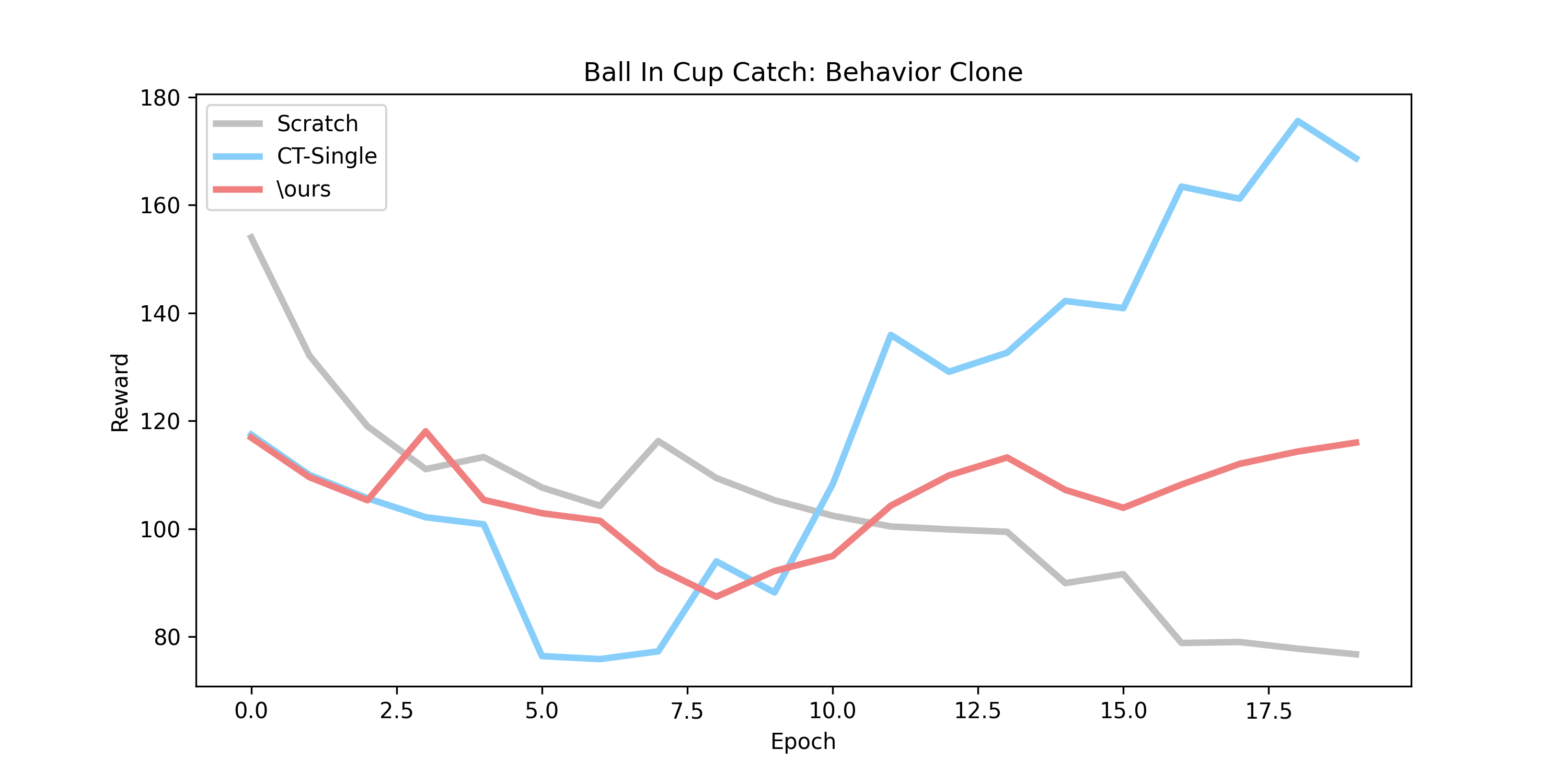}}
 \end{subfigure}
 \hfill
 \begin{subfigure}[t]{0.19\columnwidth}
  \resizebox{\textwidth}{!}{\input{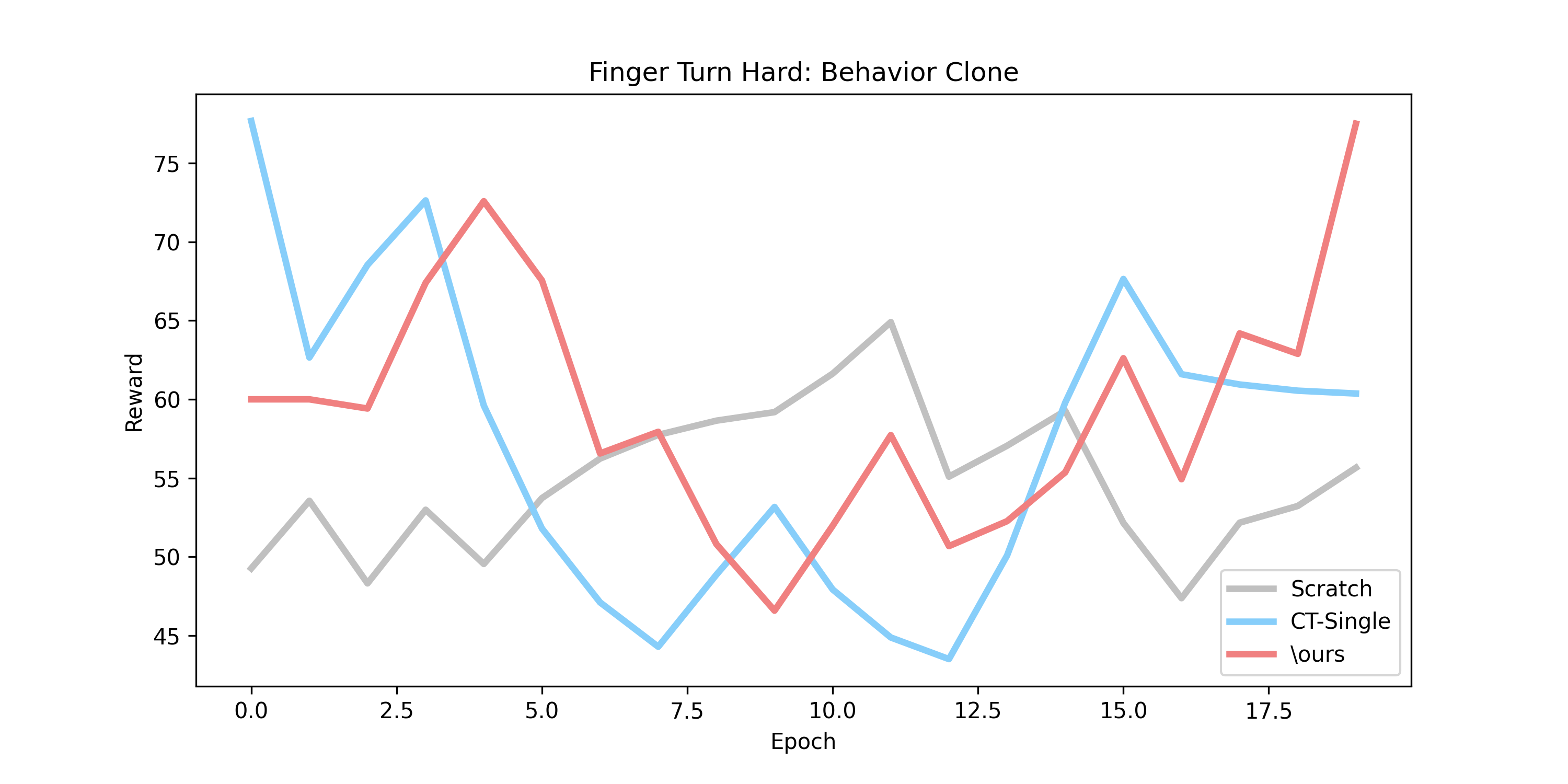}}
 \end{subfigure} 
 \hfill
 \begin{subfigure}[t]{0.19\columnwidth}
  \resizebox{\textwidth}{!}{\input{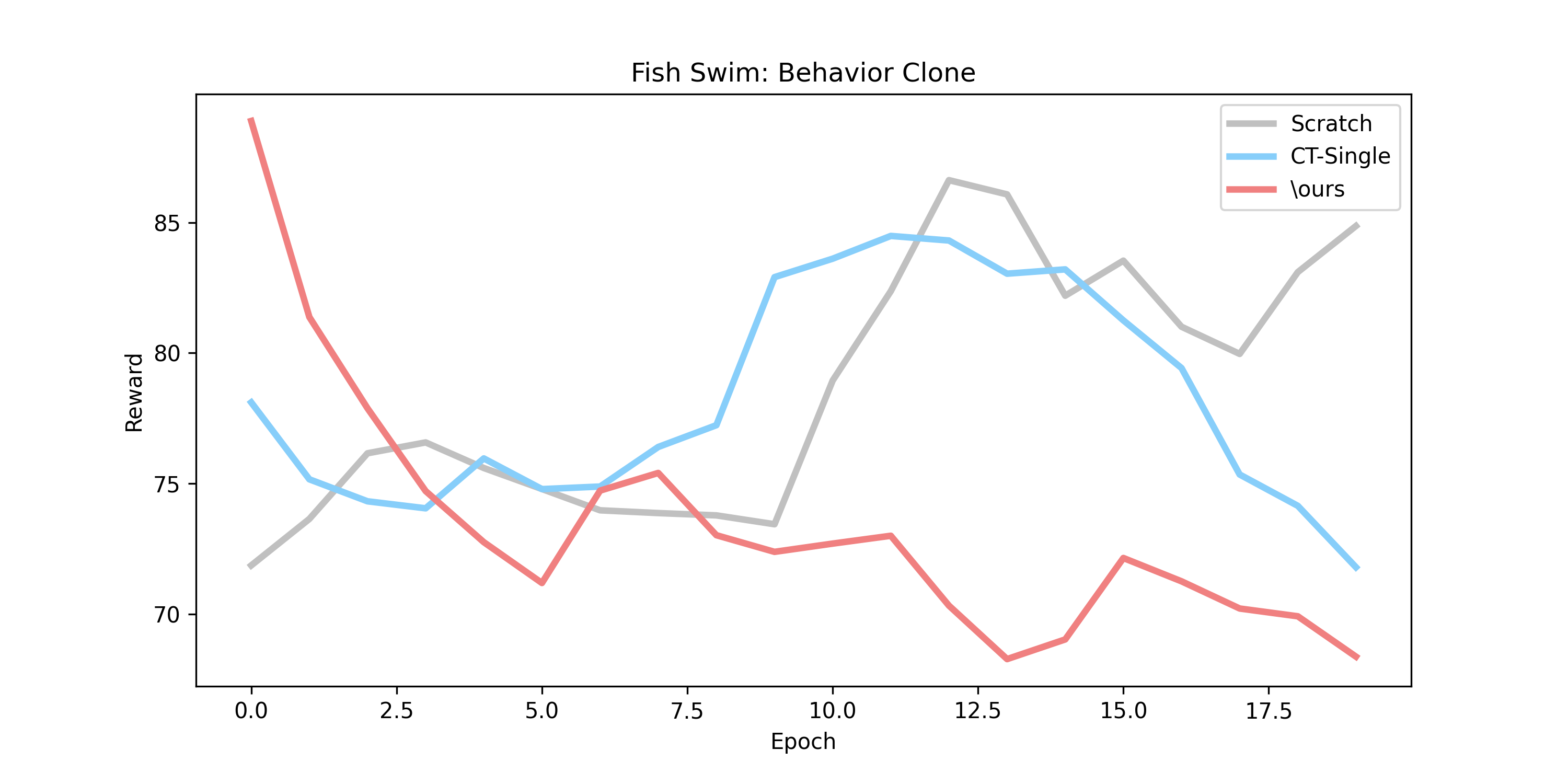}}
 \end{subfigure}
 \hfill
 \begin{subfigure}[t]{0.19\columnwidth}
  \resizebox{\textwidth}{!}{\input{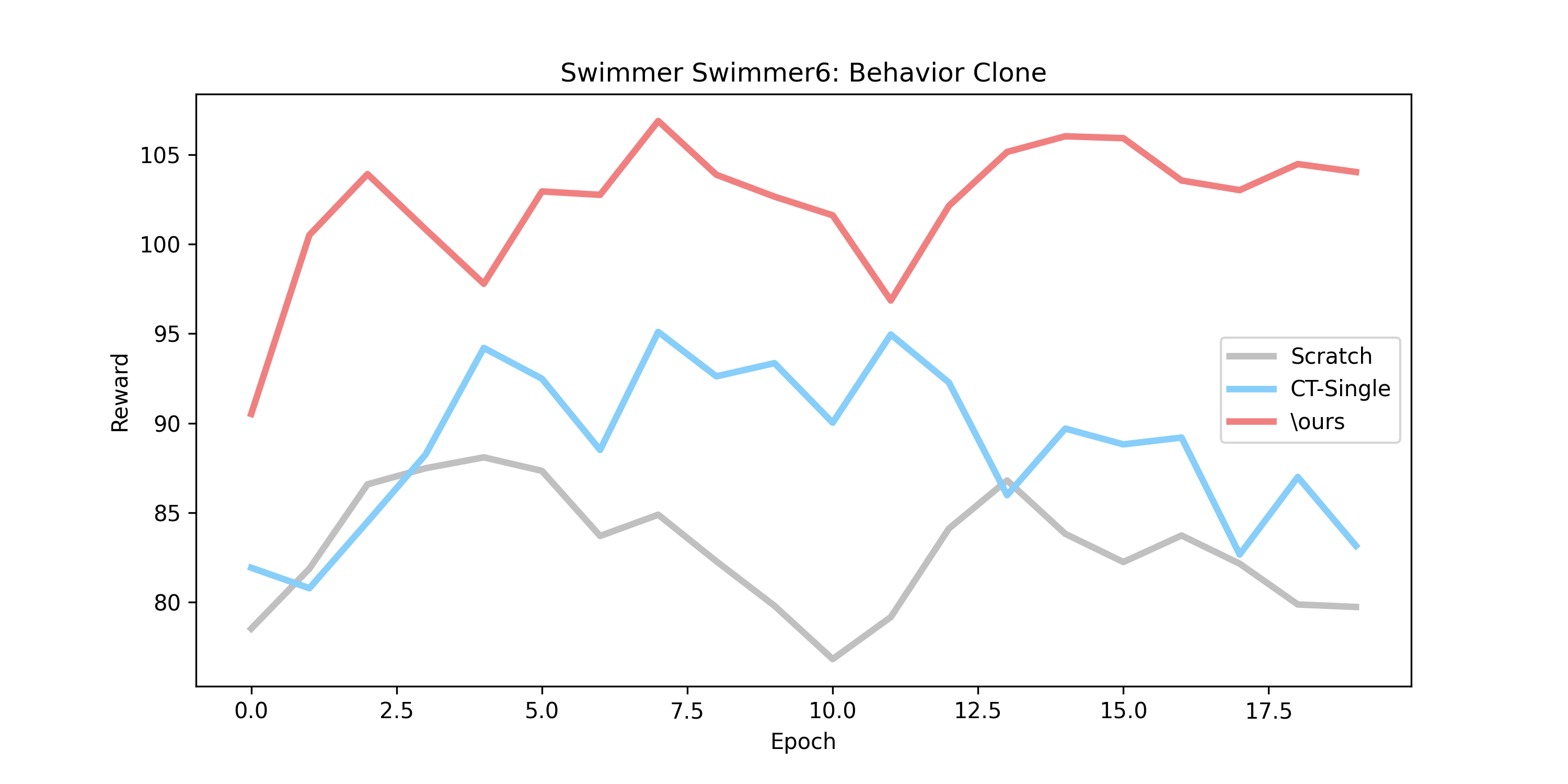}}
 \end{subfigure}
 \hfill
 \begin{subfigure}[t]{0.19\columnwidth}
  \resizebox{\textwidth}{!}{\input{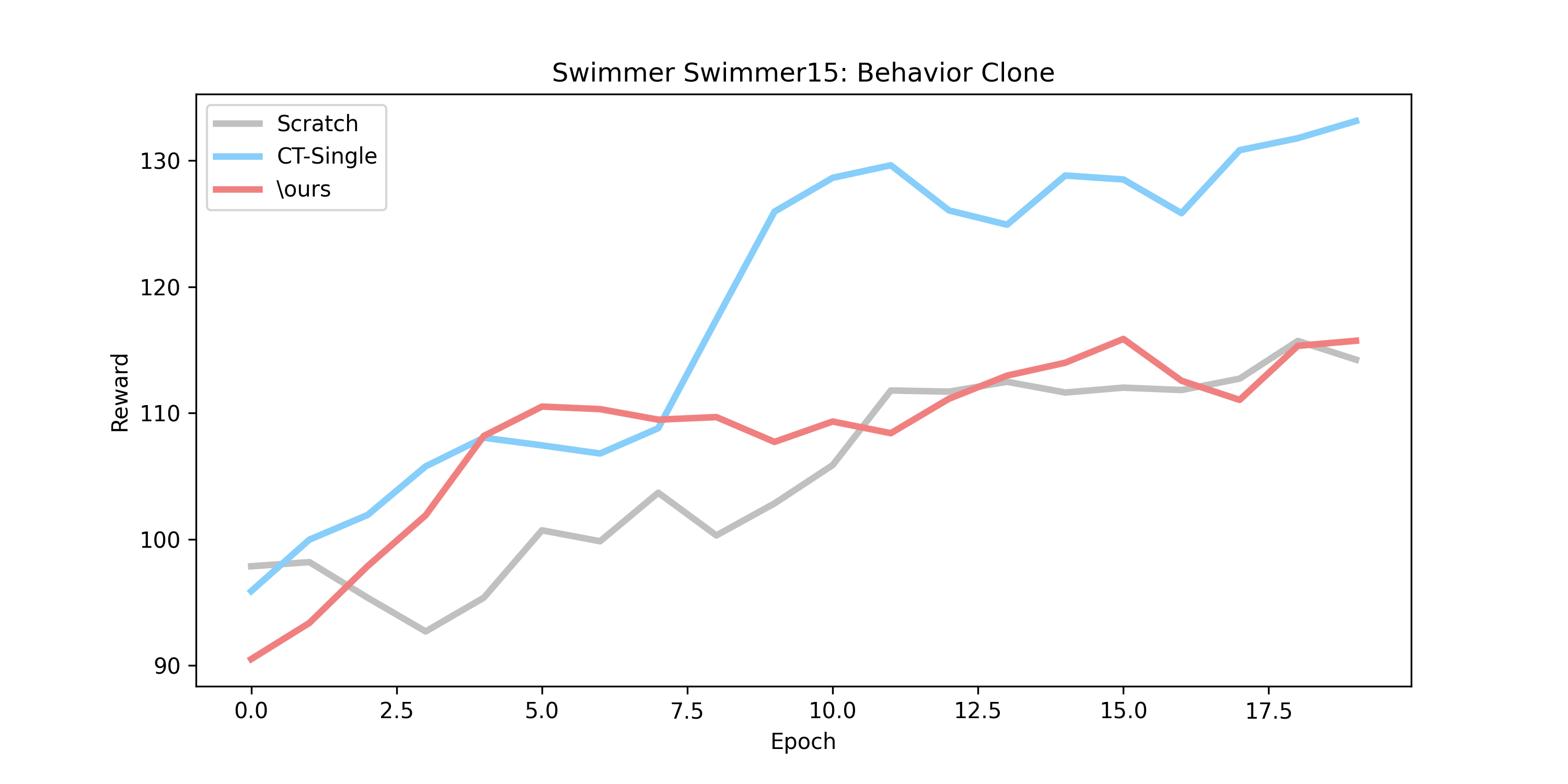}}
 \end{subfigure} 
 \vspace{-0.5em}
\caption{
Downstream learning rewards of \ours (\textcolor{red}{red}) in challenging tasks that have larger discrepancy with pretraining tasks, using the \texttt{Random} pretraining dataset. Results are from 1 random seed.
}
\label{fig:curves_add_rand}
\end{figure}

The results are shown in \cref{fig:curves_add_expl} and \cref{fig:curves_add_rand}, where we can see that the pretrained model can still work in most cases, even under such a large task discrepancy. Note that here \texttt{CT-Single} is pretrained with data from exactly the downstream task, where \ours has never seen a sample from the downstream tasks and is pretrained on significantly different domains. Therefore, it is unsurprising that \texttt{CT-Single} is generally better than \ours in this setting. However, it is interesting to see that \ours is comparable with or even better than \texttt{CT-Single} in some tasks, suggesting the strong generalizability of \ours. 

On the other hand, one can imagine that it is unavoidable that the performance of a pretrained model will decrease as the discrepancy between pretraining tasks and downstream tasks increases. Therefore, we stress the importance of using diverse multi-task data for pretraining in practice.

\subsection{Model Capacity Test}
\label{app:exp_capacity}

\begin{figure}[!htbp]
% \vspace{-1em}
 \centering
 \begin{subfigure}[t]{\columnwidth}
 \centering
  \resizebox{0.8\textwidth}{!}{\input{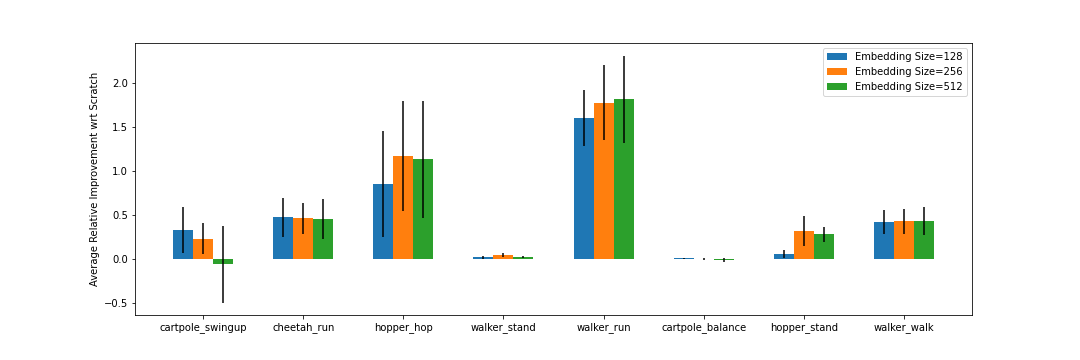}}
 \end{subfigure}

 \begin{subfigure}[t]{1\columnwidth}
 \centering
  \resizebox{0.8\textwidth}{!}{\input{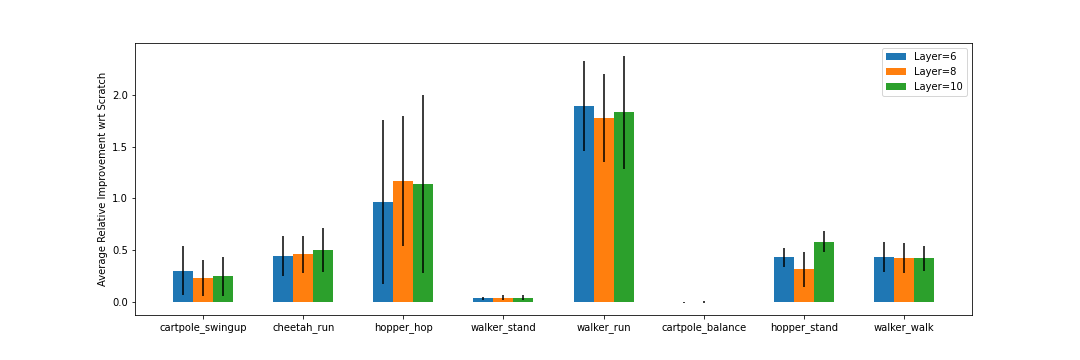}}
 \end{subfigure}
 \vspace{-1.5em}
\caption{Comparison of varying model capacities (embedding size and layer number) in different tasks in terms of relative improvement wrt training from scratch.
}
% \vspace{-1em}
\label{fig:capacity_task}
\end{figure}

\cref{fig:capacity_line} shows the results of varying model capacities averaged over tasks. To demonstrate the influence of model capacity on different tasks, we provide the per-task comparison in \cref{fig:capacity_task}.

\begin{wrapfigure}{r}{0.3\textwidth}
\vspace{-2em}
  \centering
  \resizebox{0.3\textwidth}{!}{\input{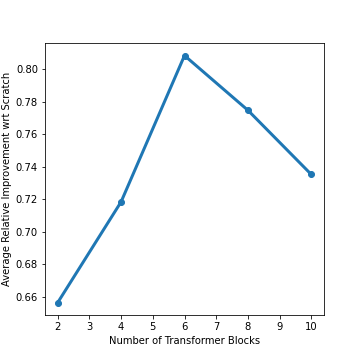}}
 \vspace{-2em}
  \caption{More model capacity tests. Results are from one random seed and are averaged over tasks.} 
  \label{fig:capacity_smaller}
 \vspace{-2em}
\end{wrapfigure}

\textbf{Further Decreasing Model Capacity.} In \cref{fig:capacity_line}, we can see that a 6-layer transformer backbone is slightly better than the 8-layer transformer. It may suggest that the current network we use is overly large for the tasks. To see whether it is the case, we further decrease the number of transformer layers to 4 and 2. The results are depicted in \cref{fig:capacity_smaller}, where we can see that a 4-layer or a 2-layer transformer is clearly worse than the 6-layer and 8-layer transformers. Therefore, the current architecture is not too heavy for the DMC tasks. The exact numbers in \cref{fig:capacity_smaller} are different from the corresponding ones in \cref{fig:capacity_line}, because \cref{fig:capacity_line} aggregates results from 3 random seeds but \cref{fig:capacity_smaller} is from 1 random seed.

\newpage

\subsection{Comparison with Pretrained ResNet Models}
\label{app:exp_resnet}

Our empirical evaluation is done for multiple transformer-based pretraining approaches. In parallel to them, there are some existing pretraining paradigms that use large-capacity models as ResNet instead of transformers as the backbones. Although it is hard to directly compare the performance of totally different model architectures, we still provide the results of ResNet pretraining, to better posit this work in literature and verify the significance of our results.

\textbf{Baselines and Implementation Details.} We use the following two state-of-the-art pretraining approaches. 
\begin{itemize}[noitemsep,leftmargin=*]
    \setlist{nolistsep}
    \item \texttt{CPC}~\citep{CPC:oord2018representation} is a self-supervised representation learning approach with contrastive predictive coding. It has been demonstrated success in many vision applications. When leveraging it in sequential decision making, state representations can be pretrained by decoupled from policy learning. 
    \item \texttt{ATC}~\citep{stooke2021decoupling} is another contrastive learning approach target on decision making tasks. It also decouples representation learning from policy learning. Different from CPC which performs InfoNCE~\citep{CPC:oord2018representation} loss in a predictive manner, ATC propose an Augmented Temporal Contrast to directly compute InfoNCE loss among temporally augmented clips.
\end{itemize}
For both \texttt{CPC} and \texttt{ATC}, we use 3D-ResNet18 as the encoder backbone. During pretraining, a 2-layer 2D-ConvNet is used as the prediction head for both CPC and ATC. For CPC, a single-layer ConvGRU is used as the aggregation network. Note that, during finetuning, both prediction heads and aggregation network are dropped. Only the pretrained encoder (3D-ResNet18) is used to produce the pretrained representations. During finetuning, we simply attach a single linear layer as the action prediction head on top of the pretrained encoder, which is under the same setting with ours.

For a fair comparison, we pretrain and finetune both \texttt{CPC} and \texttt{ATC} with a context length of 30 such that all comparing models are seeing the same time horizon. Follow the widely used training protocol in decision making tasks, we also leverage a frame stacking with stack size as 3 when training both of them. We keep the other hyperparameters the same with our default setting.

\begin{figure}[!htbp]
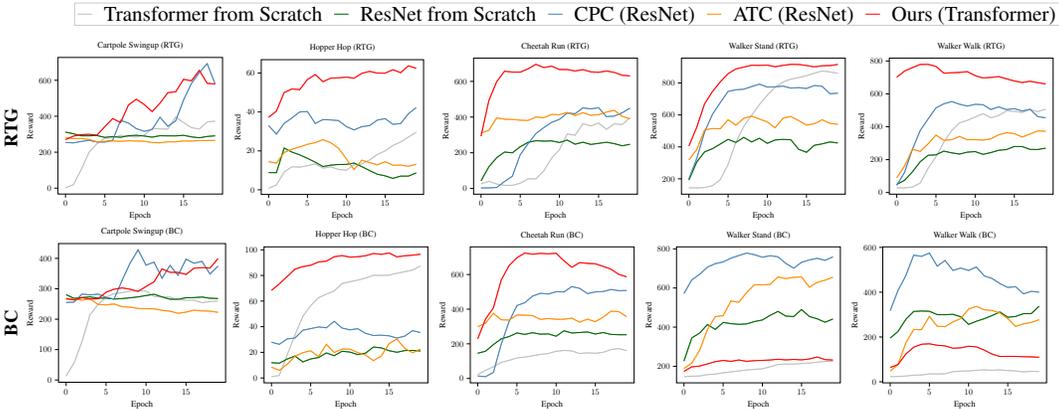

% \vspace{-1em}
 \centering

\rotatebox{90}{\scriptsize{\hspace{1cm}\textbf{RTG}}}
 \begin{subfigure}[t]{0.19\columnwidth}
  \resizebox{5.2\textwidth}{!}{\input{figures/resnet/rtg_expl_cartpole_swingup}}
 \end{subfigure}
 \hfill
 \begin{subfigure}[t]{0.19\columnwidth}
  \resizebox{0.97\textwidth}{!}{\input{figures/resnet/rtg_expl_hopper_hop}}
 \end{subfigure}
 \hfill
 \begin{subfigure}[t]{0.19\columnwidth}
  \resizebox{\textwidth}{!}{\input{figures/resnet/rtg_expl_cheetah_run}}
 \end{subfigure}
 \hfill
 \begin{subfigure}[t]{0.19\columnwidth}
  \resizebox{\textwidth}{!}{\input{figures/resnet/rtg_expl_walker_stand}}
 \end{subfigure} 
 \hfill
 \begin{subfigure}[t]{0.19\columnwidth}
  \resizebox{\textwidth}{!}{\input{figures/resnet/rtg_expl_walker_walk}}
 \end{subfigure} 
%  \vspace{0.5em}

\rotatebox{90}{\scriptsize{\hspace{1cm}\textbf{BC}}}
 \begin{subfigure}[t]{0.19\columnwidth}
  \resizebox{1.02\textwidth}{!}{\input{figures/resnet/bc_expl_cartpole_swingup}}
 \end{subfigure}
 \hfill
 \begin{subfigure}[t]{0.19\columnwidth}
  \resizebox{\textwidth}{!}{\input{figures/resnet/bc_expl_hopper_hop}}
 \end{subfigure}
 \hfill
 \begin{subfigure}[t]{0.19\columnwidth}
  \resizebox{\textwidth}{!}{\input{figures/resnet/bc_expl_cheetah_run}}
 \end{subfigure}
 \hfill
 \begin{subfigure}[t]{0.19\columnwidth}
  \resizebox{\textwidth}{!}{\input{figures/resnet/bc_expl_walker_stand}}
 \end{subfigure} 
 \hfill
 \begin{subfigure}[t]{0.19\columnwidth}
  \resizebox{\textwidth}{!}{\input{figures/resnet/bc_expl_walker_walk}}
 \end{subfigure} 
 \vspace{-0.5em}
\caption{Downstream learning rewards of \ours (\textcolor{red}{red}) compared with \texttt{CPC} (\textcolor{steelblue}{darkblue}) and \texttt{ATC} (\textcolor{darkorange}{darkorange}), using the \texttt{Exploratory} pretraining dataset. \texttt{CPC} and \texttt{ATC} are based on ResNet, whose performance of training from scratch is shown in
\textcolor{darkgreen}{darkgreen} in comparison to training transformer from scratch (\textcolor{silver}{grey}).
}
% \vspace{-1em}
\label{fig:curves_resnet_expl}
\end{figure}

\begin{figure}[!htbp]
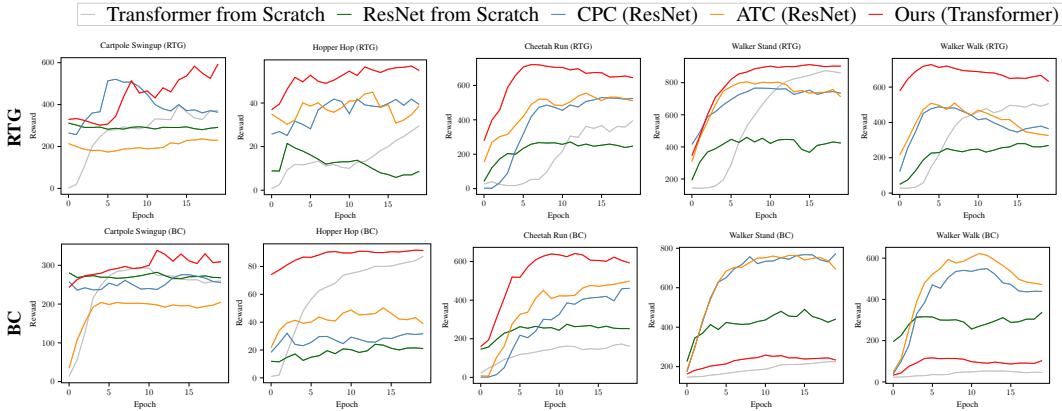

% \vspace{-1em}
 \centering

\rotatebox{90}{\scriptsize{\hspace{1cm}\textbf{RTG}}}
 \begin{subfigure}[t]{0.19\columnwidth}
  \resizebox{5.2\textwidth}{!}{\input{figures/resnet/rtg_rand_cartpole_swingup}}
 \end{subfigure}
 \hfill
 \begin{subfigure}[t]{0.19\columnwidth}
  \resizebox{0.97\textwidth}{!}{\input{figures/resnet/rtg_rand_hopper_hop}}
 \end{subfigure}
 \hfill
 \begin{subfigure}[t]{0.19\columnwidth}
  \resizebox{\textwidth}{!}{\input{figures/resnet/rtg_rand_cheetah_run}}
 \end{subfigure}
 \hfill
 \begin{subfigure}[t]{0.19\columnwidth}
  \resizebox{\textwidth}{!}{\input{figures/resnet/rtg_rand_walker_stand}}
 \end{subfigure} 
 \hfill
 \begin{subfigure}[t]{0.19\columnwidth}
  \resizebox{\textwidth}{!}{\input{figures/resnet/rtg_rand_walker_walk}}
 \end{subfigure} 
%  \vspace{0.5em}

\rotatebox{90}{\scriptsize{\hspace{1cm}\textbf{BC}}}
 \begin{subfigure}[t]{0.19\columnwidth}
  \resizebox{1.02\textwidth}{!}{\input{figures/resnet/bc_rand_cartpole_swingup}}
 \end{subfigure}
 \hfill
 \begin{subfigure}[t]{0.19\columnwidth}
  \resizebox{\textwidth}{!}{\input{figures/resnet/bc_rand_hopper_hop}}
 \end{subfigure}
 \hfill
 \begin{subfigure}[t]{0.19\columnwidth}
  \resizebox{\textwidth}{!}{\input{figures/resnet/bc_rand_cheetah_run}}
 \end{subfigure}
 \hfill
 \begin{subfigure}[t]{0.19\columnwidth}
  \resizebox{\textwidth}{!}{\input{figures/resnet/bc_rand_walker_stand}}
 \end{subfigure} 
 \hfill
 \begin{subfigure}[t]{0.19\columnwidth}
  \resizebox{\textwidth}{!}{\input{figures/resnet/bc_rand_walker_walk}}
 \end{subfigure} 
 \vspace{-0.5em}
\caption{Comparison with ResNet-based pretrained models using the \texttt{Random} pretraining dataset. 
}
% \vspace{-1em}
\label{fig:curves_resnet_rand}
\end{figure}

\textbf{Results.} We compare our \ours with \texttt{CPC} and \texttt{ATC} in RTG and BC downstream tasks, with \texttt{Exploratory} and \texttt{Random} dataset. The results are shown in \cref{fig:curves_resnet_expl} and \cref{fig:curves_resnet_rand}, respectively. 
Although training a transformer and training a ResNet model from scratch usually produces different rewards in the same task, the results show that pretrained models are able to their corresponding train-from-scratch baselines. 
% More importantly, we can see that \ours outperforms \texttt{CPC} and \texttt{ATC} in all downstream tasks, except for using BC in walker tasks where the transformer backbone fails to get a high score. 
When using RTG downstream learning, we can see that \ours outperforms \texttt{CPC} and \texttt{ATC} in all downstream tasks. When using BC as downstream learning method, the transformer backbone fails to get a high score and so does \ours pretrained models. But in remaining tasks, \ours is still significantly better than \texttt{CPC} and \texttt{ATC}.
This suggests the advantages of our pretraining framework \ours and the proposed model \ourmodfull.

\subsection{Pretrain \ourmod for Online Finetuning}
\label{app:exp_online}

Similar to most transformer-based decision making models, we consider policy learning in the form of IL and offline RL. 
For online RL, special care is needed due to the environment uncertainty and the need of exploration. A recent work by \citet{zheng2022online} proposes online decision transformer (ODT), which first pretrains the model with offline trajectories and then finetune the model in an online manner.  \citet{zheng2022online} demonstrate the effectiveness of ODT in a series of MuJoCo tasks with groundtruth state being observations. It is shown that the offline pretraining phase is crucial for online fintuning with transformer-based models. However, the pretraining phase of ODT is DT with reward supervision. With the self-supervised control-centric pretraining objective ($L_{\fwdmath}, L_{\invmath}$, $L_{\randinvmath}$) proposed in this work, we would like to ask the following questions. (1) Can we replace the supervised DT pretraining objective with our self-supervised pretraining objective that does not require reward supervision? (2) Can we improve ODT by combining our objectives with it?

\begin{figure}[!htbp]
\vspace{-1em}
 \centering
 \begin{subfigure}[t]{0.24\columnwidth}
  \resizebox{\textwidth}{!}{\input{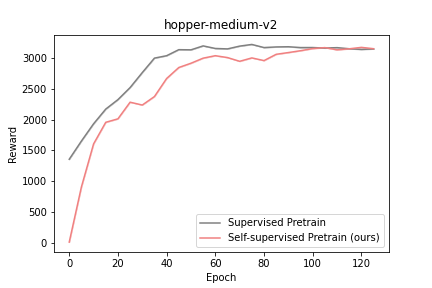}}
 \end{subfigure}
 \hfill
 \begin{subfigure}[t]{0.24\columnwidth}
  \resizebox{\textwidth}{!}{\input{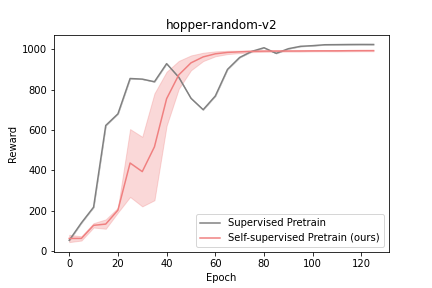}}
 \end{subfigure}
 \hfill
 \begin{subfigure}[t]{0.24\columnwidth}
  \resizebox{\textwidth}{!}{\input{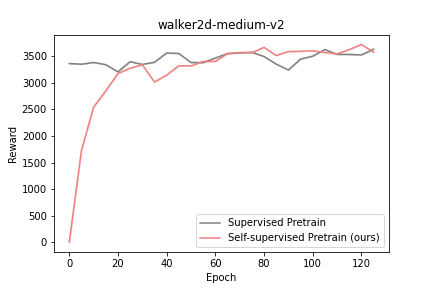}}
 \end{subfigure}
 \hfill
 \begin{subfigure}[t]{0.24\columnwidth}
  \resizebox{\textwidth}{!}{\input{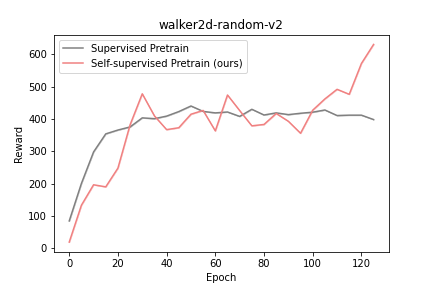}}
 \end{subfigure}
 \vspace{-0.5em}
\caption{Comparison of our self-supervised pretraining objective and the supervised pretraining objective of ODT in hopper and walker2d environments with medium and random pretraining dataset.
} 
\label{fig:odt_unsup}
\end{figure}

We follow the open-sourced implementation of ODT and evaluate our proposed objective using the default model and hyperparameter settings of ODT. We first replace the DT pretraining loss with our self-supervised losses. The results shown in \cref{fig:odt_unsup} suggests that even without any reward information, pretraining ODT with our self-supervised losses can achieve comparable performance with pretraining with reward supervisions. Although our pretrained model takes more steps to warm up due to the lack of supervised pretraining, it quickly converges to similar results as ODT in online finetuning.
Therefore, in practical applications where only reward-free pretraining trajectories are available, DT pretraining is infeasible while our pretraining can be used without sacrificing the finetuning performance.

\begin{figure}[!htbp]
\vspace{-1em}
 \centering
 \begin{subfigure}[t]{0.24\columnwidth}
  \resizebox{\textwidth}{!}{\input{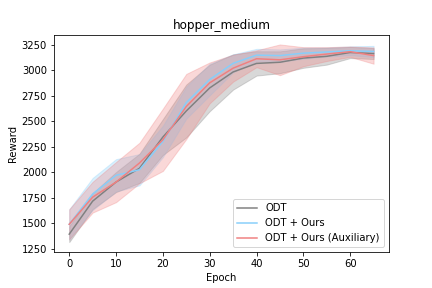}}
 \end{subfigure}
 \hfill
 \begin{subfigure}[t]{0.24\columnwidth}
  \resizebox{\textwidth}{!}{\input{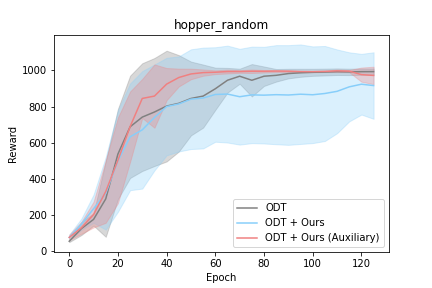}}
 \end{subfigure}
 \hfill
 \begin{subfigure}[t]{0.24\columnwidth}
  \resizebox{\textwidth}{!}{\input{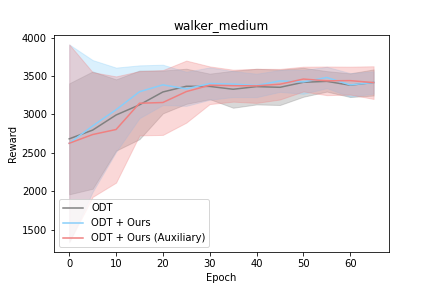}}
 \end{subfigure}
 \hfill
 \begin{subfigure}[t]{0.24\columnwidth}
  \resizebox{\textwidth}{!}{\input{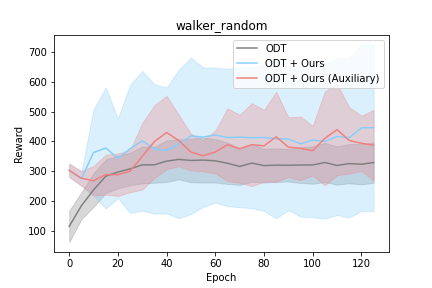}}
 \end{subfigure}
 \vspace{-0.5em}
\caption{Combining our pretraining objectives with ODT produces better results, especially when the pretraining dataset has low quality. Results are averaged over 5 random seeds.
} 
\label{fig:odt_aux}
\end{figure}

To answer the second question, we combine our pretraining objective with ODT. 
There are two potential ways of such combination: (1) only combine in pretraining, and (2) add our self-supervised losses during both pretraining and finetuning (the losses serve as auxiliary tasks in online finetuning). We evaluate both versions in our experiments, as shown in \cref{fig:odt_aux}. 
The results suggest that including our pretraining objective can improve the performance of ODT, especially when the data quality is low (\{task\}-random dataset is used). However, when our objective is only used during pretraining, it suffers from high variance in downstream learning. To address this issue, we find that adding our objective as auxiliary loss in finetuning can effectively improve the result and decrease the variance. 

It is also interesting to see that the online learning performance of ODT depends on the quality of pretraining dataset. When the random dataset is used, ODT converges to a suboptimal policy in finetuning, although it could have explored better solutions. Adding our self-supervised pretraining objective improves the performance, but it still cannot match the performance of models pretrained in a higher-quality dataset. We hypothesize that this is due to a very large distribution shift between the random dataset and the ideal dataset for online learning. That is, the model pretrained with random data overfits to the random behavior and fails to further explore and exploit.
How to make a transformer-based model adapt to an online environment with low-quality pretraining data is still an open problem that we aim to study in our future work.

\subsection{Additional Results of Ablation Study}
\label{app:exp_ablation}

\subsubsection{Ablation of Pretraining Objective}
\label{app:exp_ablation_obj}

\begin{figure}[!htbp]
    \centering
    \begin{subfigure}[t]{0.48\columnwidth}
        \resizebox{\textwidth}{!}{\input{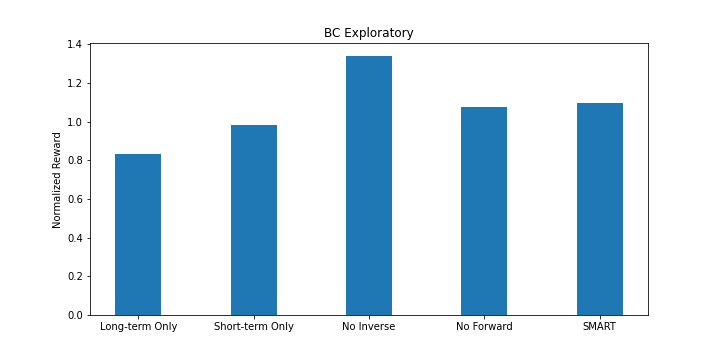}}
        % \caption{Ablation of objectives.}
    \end{subfigure}
    \hfill
    \begin{subfigure}[t]{0.48\columnwidth}
        \resizebox{\textwidth}{!}{\input{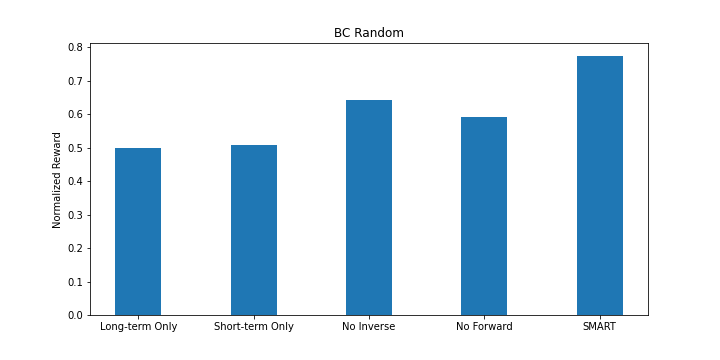}}
        % \caption{Ablation of long-term and short-term objectives.}
    \end{subfigure}
    \vspace{-3mm}
    \caption{Ablation study of our pretraining objective with BC (behavior cloning) being the downstream learning algorithm. Both long-term control information (\randinvshort) and short-term control information (Forward and Inverse) are important.}
    \label{fig:ablation_bc}
    \vspace{-3mm}
\end{figure}

In addition to \cref{fig:ablation} which uses RTG as the downstream learning objective, we also provide the ablation results of BC downstream learning in \cref{fig:ablation_bc}. Similar to the results of RTG, removing any term from the control-centric pretraining objective usually causes a performance drop, which verifies the effectiveness of our objective design. One exception is that removing the inverse prediction gives a better result for BC with exploratory pretraining data. Combined with results in other cases, we can find that inverse prediction is not as helpful as forward prediction and \randinv on average. But in most cases, especially when pretraining data is low-quality (\texttt{Random}), including the inverse prediction still improves the performance. 

\textbf{Implementation Details.}
For the ablation study, we pretrain the model with the same hyperparameter settings on the same datasets using different versions of (ablated) pretraining objectives. 
The results are averaged over 8 tasks: cartpole-swingup, cartpole-balance, hopper-hop, hopper-stand, cheetah-run, walker-stand walker-run, and walker-walk, spanning both seen and unseen tasks. 
Due to the large number of experiments (4 variants $\times$ 2 learning scenarios $\times$ 2 dataset selections $\times$ 8 downstream tasks = 128 experiments), the results in \cref{fig:ablation} and \cref{fig:ablation_bc} are from one random seed. Although randomness exists, we believe that the conclusion is meaningful as the results are averaged over multiple tasks and multiple learning scenarios.

\subsubsection{Variants of Pretraining Objective}
\label{app:exp_ablation_variant}

% This paper proposes \ours, a generic pretraining framework for control tasks. Our proposed \ourmodfull and the self-supervised control-centric objective are shown to be versatile, generalizable and resilient, and thus can be widely applied to model pretraining in many tasks. 
We also investigate some other possible variants of the pretraining objective. \\
\textbf{(1) Multi-step Inverse} (inspired by~\citet{lamb2022guaranteed}): masking all states/actions but $s_t$ and $s_{t+L}$ and predicting $a_t$.  \\
\textbf{(2) Max Fixed Mask}: using a fixed mask size $k=L$ and $k^\prime=L/2$ instead of gradually increasing $k$ and $k^\prime$.\\
\textbf{(3) + Masked State Prediction}: in addition to the original objective, this variant also predicts the masked state token in the third term. (The state token target is generated from the momentum encoder, as in forward prediction.)\\
\textbf{(4) + Contrastive Loss}: in addition to the original objective, this variant also employs an additional contrastive loss, which takes the state/action token and their representations for the same timestep as positive pairs, while regarding representations from other timesteps as negative samples  (inspired by \citet{banino2022coberl}). \\
The implementation and comparison of these variants follow the same setup with the ablation study in \cref{app:exp_ablation_obj}. Note that (3) and (4) are both adding new losses to the original method, which renders higher computational cost.

\begin{figure}[!htbp]
    \centering
    \begin{subfigure}[t]{0.48\columnwidth}
        \resizebox{\textwidth}{!}{\input{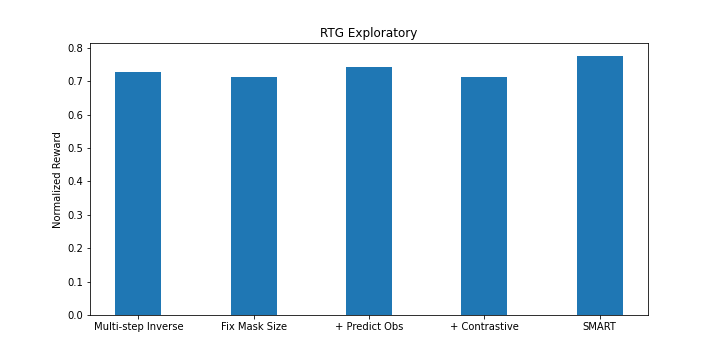}}
    \end{subfigure}
    \hfill
    \begin{subfigure}[t]{0.48\columnwidth}
        \resizebox{\textwidth}{!}{\input{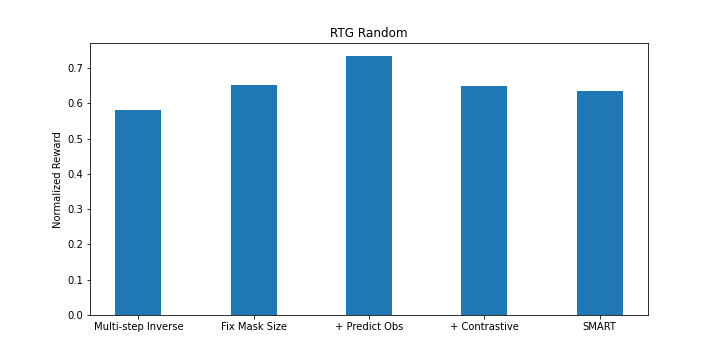}}
    \end{subfigure}
    \vspace{-3mm}
    \caption{Comparison between \ours and its variants. Adding masked state prediction and contrastive loss to \ours could increase the performance, but the improvement is not across all scenarios, with a cost of more computations.}
    \label{fig:variants}
    \vspace{-3mm}
\end{figure}

The results of comparison with these variants are shown in \cref{fig:variants}, with RTG being the downstream learning objective. The results show that Multi-step Inverse and Max Fixed Mask are in general worse than \ours, suggesting the effectiveness of our design. Interestingly, optimizing an extra state prediction loss or a contrastive loss gives better performance when the pretraining data is \texttt{Random}, while hurts the performance when the pretraining data is \texttt{Exploratory}. We hypothesize that this is because such two additional losses emphasize more on capturing visual information from pixel observations, which is more important for learning and transferring representation from the \texttt{Random} dataset. In other words, a better visual representation is more crucial under a larger distribution shift (on behavior policies) that happened when a \texttt{Random} dataset is used for pretraining.
On the other hand, when the pretraining data is of higher quality, adding these two components downgrades the original \ours. A potential reason is that with a smaller distribution shift, the control-centric objective can already capture sufficient information from data, while adding new losses may make the optimization harder and less stable.

In summary, although none of the tested variants can outperform \ours in all scenarios, some of them can render better results in certain cases by adding extra pretraining losses to the original objective, at a cost of more computations and possibly more challenging optimization. 
Therefore, whether to use these variants in practice depends on the specific use case and computation resources. A deeper and more thorough understanding of the relationships among these pretraining losses is desired for better application of pretraining methods, which would be our future work. 
We emphasize that multi-task pretraining for control tasks is a relatively new area. This paper takes one step further towards general large-scale pretraining models for sequential decision making. Exploring more possibilities of \ours would be an interesting and important future direction to further improve its performance in a variety of real-world tasks.

\end{document}